# Proceedings of the IJCAI 2017 Workshop on Learning in the Presence of Class Imbalance and Concept Drift (LPCICD'17)

# International Joint Conference on Artificial Intelligence

August 20th, 2017, Melbourne, Australia

**Organisers:**
Shuo Wang
Leandro L. Minku
Nitesh Chawla
Xin Yao

# Welcome to the IJCAI 2017 Workshop on Learning in the Presence of Class Imbalance and Concept Drift (LPCICD'17)

It is our pleasure to welcome you to the IJCAI 2017 Workshop on Learning in the Presence of Class Imbalance and Concept Drift (LPCICD'17), to be held in Melbourne, Australia, on August 20$^{th}$, 2017, at the International Joint Conference on Artificial Intelligence (IJCAI'17).

With the wide application of machine learning algorithms to the real world, class imbalance and concept drift have become crucial learning issues. Applications in various domains such as risk management, anomaly detection, fraud detection, software engineering, social media mining, and recommender systems are affected by both class imbalance and concept drift. Class imbalance happens when the data categories are not equally represented, i.e., at least one category is minority compared to other categories. It can cause learning bias towards the majority class and poor generalization. Concept drift is a change in the underlying distribution of the problem, and is a significant issue specially when learning from data streams. It requires learners to be adaptive to dynamic changes.

Class imbalance and concept drift can significantly hinder predictive performance, and the problem becomes particularly challenging when they occur simultaneously. This challenge arises from the fact that one problem can affect the treatment of the other. For example, drift detection algorithms based on the traditional classification error may be sensitive to the imbalanced degree and become less effective; and class imbalance techniques need to be adaptive to changing imbalance rates, otherwise the class receiving the preferential treatment may not be the correct minority class at the current moment. Therefore, the mutual effect of class imbalance and concept drift should be considered during algorithm design.

The aim of this workshop is to bring together researchers from the areas of class imbalance learning and concept drift in order to encourage discussions and new collaborations on solving the combined issue of class imbalance and concept drift. It provides a forum for international researchers and practitioners to share and discuss their original work on addressing new challenges and research issues in class imbalance learning, concept drift, and the combined issues of class imbalance and concept drift.

We received a total of 12 paper submissions. After a reviewing process, in which each paper has been reviewed by at least three members of the program committee, we accepted a total of 8 papers. The acceptance criteria were entirely based on the quality of the papers, without imposing any constraint on the number of papers to be accepted.

We are delighted to announce two outstanding keynotes. Dr. João Gama from University of Porto (Portugal) will give a keynote about "Monitoring the Learning Process". Prof. Dacheng Tao from the University of Sydney (Australia) will give a keynote about "Recent Progress in AI: From Perceiving, Learning and Reasoning to Behaving".

We would like to thank the keynote speakers, authors, program committee members and additional reviewers for their contributions to make this workshop a success.

We hope you will enjoy LPCICD'17. We certainly will!

The LPCICD'17 Organization Committee
Shuo Wang, Leandro Minku, Nitesh Chawla and Xin Yao

# LPCICD'17's Team

## Organizing Committee

Shuo Wang, *University of Birmingham, UK*
Leandro L. Minku, *University of Leicester, UK*
Nitesh Chawla, *University of Notre Dame, USA*
Xin Yao, *Southern University of Science and Technology, China*

## Webmaster

Mohammad Nabi Omidvar, *University of Birmingham, UK*

## Program Committee

Cesare Alippi, *Politecnico di Milano, Italy*
Alfred Bifet, *Télécom ParisTech, France*
Giacomo Boracchi, *Politecnico di Milano, Italy*
Gregory Ditzler, *The University of Arizona, USA*
João Gama, *University of Porto, Portugal*
Barbara Hammer, *Bielefeld University, Germany*
Haibo He, *University of Rhode Island, USA*
Nathalie Japkowicz, *American University, USA*
Xu-Ying Liu, *Southeast University, China*
Bilal Mirza, *Singapore Polytechnic, Singapore*
Adriano L. I. Oliveira, *Federal University of Pernambuco, Brazil*
Rita P. Ribeiro, *University of Porto, Portugal*
Manuel Roveri, *Politecnico di Milano, Italy*
Ke Tang, *University of Science and Technology of China, China*
Zhihua Zhou, *Nanjing University, China*

## Additional Reviewers

George G. Cabral, *Federal Rural University of Pernambuco, Brazil*
Rodolfo C. Cavalcante, *Federal University of Pernambuco, Brazil*
Tao Chen, *University of Birmingham, UK*
Xiaofen Lu, *University of Science and Technology of China, China*
Liyan Song, *Southern University of Science and Technology, China*
Yu Sun, *University of Science and Technology of China, China*

# Table of Contents

# Keynote 1

## Monitoring the Learning Process
*Dr. João Gama, University of Porto, Portugal*

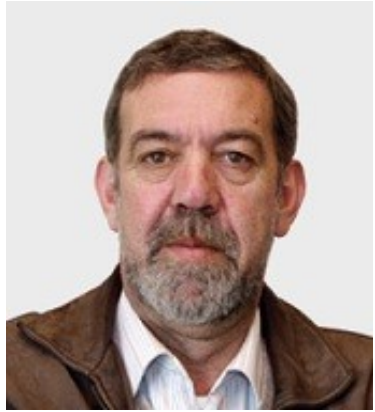

## Abstract

Data Mining is faced with new challenges. In emerging applications (like financial data, traffic TCP/IP, sensor networks, etc) data continuously flow eventually at high speed. The processes generating data evolve over time, and the concepts we are learning change. In this talk we present a one-pass classification algorithm able to detect and react to changes. We present a framework that identifies contexts using drift detection, characterises contexts using meta-learning, and selects the most appropriate base model for the incoming data using unlabelled examples.

Evolving data requires that learning algorithms must be able to monitor the learning process and the ability of predictive self-diagnosis. A significant and useful characteristic is diagnostics – not only after failure has occurred, but also predictive (before failure). These aspects require monitoring the evolution of the learning process, taking into account the available resources, and the ability of reasoning and learning about it.

## Biography

Dr. João Gama is an Associate Professor at the University of Porto, Portugal. He is also a senior researcher and member of the board of directors of the Laboratory of Artificial Intelligence and Decision Support (LIAAD), a group belonging to INESC Porto. João Gama serves as the member of the Editorial Board of Machine Learning Journal, Data Mining and Knowledge Discovery, Intelligent Data Analysis and New Generation Computing. He served as Co-chair of ECML 2005, DS09, ADMA09 and a series of Workshops on KDDS and Knowledge Discovery from Sensor Data with ACM SIGKDD. He was also the chair for the conference of Intelligent Data Analysis 2011. He has given 7 keynotes and 2 plenary talks. His main research interest is in knowledge discovery from data streams and evolving data. He is the author of a recent book on Knowledge Discovery from Data Streams. He has extensive publications in the area of data stream learning.

# Keynote 2

## Recent Progress in AI:
## From Perceiving, Learning and Reasoning to Behaving

*Prof. Dacheng Tao, University of Sydney, Australia*

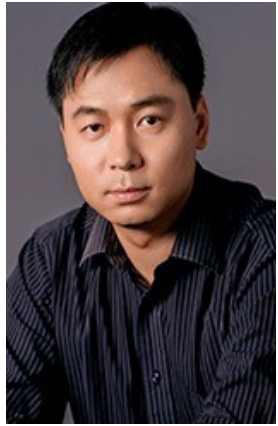

## Abstract

Since the concept of Turing machine has been first proposed in 1936, the capability of machines to perform intelligent tasks went on growing exponentially. Artificial Intelligence (AI), as an essential accelerator, pursues the target of making machines as intelligent as human beings. It has already reformed how we live, work, learning, discover and communicate. In this talk, I will review our recent progress on AI by introducing some representative advancements from algorithms to applications, and illustrate the stairs for its realization from perceiving to learning, reasoning and behaving. To push AI from the narrow to the general, many challenges lie ahead. I will bring some examples out into the open, and shed lights on our future target. Today, we teach machines how to be intelligent as ourselves. Tomorrow, they will be our partners to get into our daily life.

## Biography

Dacheng Tao is Professor of Computer Science and ARC Future Fellow in the School of Information Technologies and the Faculty of Engineering and Information Technologies at The University of Sydney. He was Professor of Computer Science and Director of the Centre for Artificial Intelligence in the University of Technology Sydney. He mainly applies statistics and mathematics to Artificial Intelligence and Data Science. His research interests spread across computer vision, data science, image processing, machine learning, and video surveillance. His research results have expounded in one monograph and 500+ publications at top journals and conferences, such as IEEE T-PAMI, T-NNLS, T-IP, JMLR, IJCV, IJCAI, AAAI, NIPS, ICML, CVPR, ICCV, ECCV, ICDM; and ACM SIGKDD, with several best paper awards, such as the best theory/algorithm paper runner up award in IEEE ICDM'07, the best student paper award in IEEE ICDM'13, and the 2014 ICDM 10-year highest-impact paper award. He received the 2015 Australian Scopus-Eureka Prize, the 2015 ACS Gold Disruptor Award and the 2015 UTS Vice-Chancellor's Medal for Exceptional Research. He is a Fellow of the IEEE, OSA, IAPR and SPIE.

# Linear Classifier Design for Heteroscedastic LDA under Class Imbalance

Kojo Sarfo Gyamfi, James Brusey, Andrew Hunt, Elena Gaura
Faculty of Engineering and Computing, Coventry University, Coventry, United Kingdom

## Abstract

Linear Discriminant Analysis (LDA) yields the optimal Bayes classifier for binary classification for normally distributed classes with equal covariance. To improve the performance of LDA, heteroscedastic LDA (HLDA) that removes the equal covariance assumption has been developed. In this paper, we show that the existing approaches either have no principled computational procedure for optimal parameter selection, or underperform in terms of the accuracy of classification and the area under the receiver operating characteristics curve (AUC) under class imbalance. We then derive a Bayes optimal linear classifier for heteroscedastic LDA that is robust against class imbalance and is obtained via an efficient gradient descent optimisation procedure. Our experimental work on one artificial dataset shows that our proposed algorithm achieves the minimum misclassification rate as compared to existing HLDA approaches if the errors in both the minority and majority classes are of equal importance. In the scenario where the errors in the minority class may be of more importance, further experiments on five real-world datasets show the superiority of our algorithm in terms of the AUC as compared to the original LDA procedure, existing HLDA algorithms, and the linear support vector machine (SVM).

## 1 Introduction

Statistical classification is a fundamental task in many machine learning applications. It can take many forms, such as classifying an incoming email as either spam or non-spam, or categorising a person as standing or sitting (in the field of human activity recognition). More generally, classification involves assigning an object **x** to belong to one of $K$ distinct classes. Arguably, one of the most well-known algorithms for performing such tasks is the linear support vector machine (SVM).

However, for applications that tend to have datasets whose distributions are close to normal, a low-complexity alternative to the SVM for binary classification is Linear Discriminant Analysis (LDA). At its core, LDA makes assumptions on the data, namely that the data in each class has a multivariate normal distribution, and that the covariance matrices of these distributions are equal among the classes. Another assumption LDA makes is that the distributions of the data in the individual classes are non-overlapping [Izenman, 2009]. This permits the construction of a linear boundary to discriminate between the classes. When these assumptions are met, LDA yields the optimal Bayes linear classifier [Izenman, 2009; Hamsici and Martinez, 2008]. While these assumptions are not often encountered in practice, the robustness provided by it being a linear classifier has encouraged the use of LDA for many applications [Mika *et al.*, 1999]. Moreover, the fact that many physical data tend to have distributions that are close to normal [Lyon, 2014] make the performance of LDA satisfactory for a lot of applications [Guo *et al.*, 2007; Yu and Yang, 2001; Sharma and Paliwal, 2008].

However, experimental results, in work by, for example, [Mika *et al.*, 1999; Hastie and Tibshirani, 1996; Marks and Dunn, 1974; Zhao *et al.*, 2009], have shown that if one accounts for the violation of the assumptions in the original procedure, the performance of LDA can be improved. Kernel Fisher Discriminant (KFD) [Mika *et al.*, 1999], for instance, has been developed for the case where the patterns are overlapping, where a linear boundary will be inappropriate. The KFD applies the kernel trick [Barber, 2012] to LDA in order to learn non-linear decision boundaries. Other work has focused on the violation of the normality assumption by modelling a non-normal distribution as a mixture of Gaussians [Hastie and Tibshirani, 1996]. Other non-parametric approaches to LDA include using a local neighbourhood structures to overcome the normality assumption [Cai *et al.*, 2007].

Accounting for the difference in covariance matrices has led to several heteroscedastic extensions of LDA [Duin and Loog, 2004; Decell *et al.*, 1981; Malina, 1981; Loog and Duin, 2002; McLachlan, 2004; Decell Jr and Marani, 1976; Decell and Mayekar, 1977]. However, most of the above references have been concerned with linear dimensionality reduction, which involves finding a linear transformation that transforms the feature vector **x** into one of reduced dimensionality, while maximising the class discriminatory information. Our focus here, however, is not on dimensionality reduction, but on the design of an optimum linear classifier or discriminant for binary classification under heteroscedasticity.

We note that the natural extension of LDA to the heteroscedastic setting is the Quadratic Discriminant Analysis (QDA). However, the decision boundary is quadratic in QDA, and it may not be as robust as a linear classifier for a lot of applications, as far as generalisation is concerned. This has led to several linear approximations of the quadratic boundary in QDA such as the works by [Marks and Dunn, 1974], [Anderson and Bahadur, 1962], [Peterson and Mattson, 1966] and [Fukunaga, 2013]. Unfortunately, the approaches taken in [Marks and Dunn, 1974] and [Anderson and Bahadur, 1962] present no principled computational procedure for optimum parameter selection; we refer to this as random heteroscedastic linear discriminant (RHLD). [Peterson and Mattson, 1966] and [Fukunaga, 2013], on the other hand, present an iterative procedure for obtaining the optimum linear classifier under heteroscedasticity; we refer to this procedure in this paper as the heteroscedastic linear discriminant (HLD). The HLD is however only locally optimal, as we show in Section 3. Under class imbalance [Xue and Titterington, 2008], where there are far more objects in one class than the other, if the population means are not well-separated, the HLD achieves a relatively poorer accuracy than the original LDA procedure, which has been shown to be robust against class imbalance [Xue and Titterington, 2008].

We propose in this paper an approach for obtaining the Bayes optimal linear discriminant function for heteroscedastic LDA, which involves minimising the probability of misclassification under the normality assumption. We derive the first and second-order optimality conditions, using these to obtain an efficient gradient descent optimisation procedure, as opposed to the systematic trial and error approach in [Marks and Dunn, 1974] and [Anderson and Bahadur, 1962]. Our algorithm, unlike that in [Peterson and Mattson, 1966] and [Fukunaga, 2013], is robust against unbalanced data, and is discussed in Sections 4 and 5.

We evaluate the proposed classifier experimentally on one artificial dataset and 5 real world datasets from the University of California, Irvine (UCI) machine learning repository. We compare our algorithm to the original LDA procedure, HLD, RHLD, QDA and linear SVM [Hsu and Lin, 2002]. The results of these experiments are presented in Section 6.

## 2 Background and Related Work

Consider a training dataset $\mathcal{D}$ made up of $N$ feature vectors each of dimensionality $d$, i.e. $\mathcal{D} = \{\mathbf{x}^{(1)}, ..., \mathbf{x}^{(N)}\}$. We consider binary classification where there are only two distinct classes $\mathcal{C}_1$ and $\mathcal{C}_2$ in $\mathcal{D}$ (See [Hsu and Lin, 2002] for multi-class classification). In line with LDA, we assume that the data in each class has a normal distribution with a mean of $\bar{\mathbf{x}}_1$ and covariance of $\mathbf{\Sigma}_1$ for $\mathcal{C}_1$, and mean of $\bar{\mathbf{x}}_2$ and covariance of $\mathbf{\Sigma}_2$ for $\mathcal{C}_2$.

We seek to find a vector of weights $\mathbf{w} \in \mathbb{R}^d$ and a threshold $w_0 \in \mathbb{R}$ such that for a given feature vector $\mathbf{x}$ from a test set, the predicted class $\mathcal{C}^*(\mathbf{x})$:

$$\mathcal{C}^*(\mathbf{x}) = \begin{cases} \mathcal{C}_1 & \text{if} \quad y = \mathbf{w}^T \mathbf{x} \geq w_0 \\ \mathcal{C}_2 & \text{if} \quad y = \mathbf{w}^T \mathbf{x} < w_0 \end{cases} \quad (1)$$

Since $\mathbf{x}$ belongs to either class $\mathcal{C}_1$ or $\mathcal{C}_2$, $y$ is normally distributed with a mean of $\mu_1$ and a variance of $\sigma_1^2$ in class $\mathcal{C}_1$, and normally distributed with a mean of $\mu_2$ and a variance of $\sigma_2^2$ in class $\mathcal{C}_2$ given as:

$$\mu_1 = \mathbf{w}^T \bar{\mathbf{x}}_1 \quad \mu_2 = \mathbf{w}^T \bar{\mathbf{x}}_2 \quad \sigma_1^2 = \mathbf{w}^T \mathbf{\Sigma}_1 \mathbf{w} \quad \sigma_2^2 = \mathbf{w}^T \mathbf{\Sigma}_2 \mathbf{w} \quad (2)$$

The Bayes optimal linear classifier may be obtained minimising the probability of misclassification [Fukunaga, 2013] as given by:

$$p_e = \pi_1 p(y < w_0 | \mathcal{C}_1) + \pi_2 p(y \geq w_0 | \mathcal{C}_2) \quad (3)$$

where $\pi_1$ and $\pi_2$ are the prior probabilities of the data in classes $\mathcal{C}_1$ and $\mathcal{C}_2$ respectively.

The individual misclassification probabilities can be expressed as:

$$p(y < w_0 | \mathcal{C}_1) = \int_{-\infty}^{w_0} \frac{1}{\sqrt{2\pi}\sigma_1} \exp\left[-\frac{(\zeta - \mu_1)^2}{2\sigma_1^2}\right] d\zeta = 1 - Q\left(\frac{w_0 - \mu_1}{\sigma_1}\right) \quad (4)$$

and

$$p(y \geq w_0 | \mathcal{C}_2) = \int_{w_0}^{\infty} \frac{1}{\sqrt{2\pi}\sigma_2} \exp\left[-\frac{(\zeta - \mu_2)^2}{2\sigma_2^2}\right] d\zeta = Q\left(\frac{w_0 - \mu_2}{\sigma_2}\right) \quad (5)$$

where $Q(\cdot)$ is the Q-function. Therefore, the probability of misclassification, which is to be minimised, may be rewritten as:

$$p_e = \pi_1 \big[1 - Q(z_1)\big] + \pi_2 \big[Q(z_2)\big] \quad (6)$$

where

$$z_1 = \frac{w_0 - \mu_1}{\sigma_1} \quad \text{and} \quad z_2 = \frac{w_0 - \mu_2}{\sigma_2} \quad (7)$$

If the homoscedasticity assumption holds, i.e. if $\mathbf{\Sigma}_1 = \mathbf{\Sigma}_2 = \mathbf{\Sigma}_x$, then it can be shown [Izenman, 2009] that LDA minimises $p_e$ (6) with the following weight vector $\mathbf{w}$ and threshold $w_0$:

$$\mathbf{w} = \mathbf{\Sigma}_x^{-1}(\bar{\mathbf{x}}_1 - \bar{\mathbf{x}}_2) \quad (8)$$

and

$$w_0 = \ln \frac{\pi_2}{\pi_1} + \frac{1}{2}(\bar{\mathbf{x}}_1^T \mathbf{\Sigma}_x^{-1} \bar{\mathbf{x}}_1 - \bar{\mathbf{x}}_2^T \mathbf{\Sigma}_x^{-1} \bar{\mathbf{x}}_2) \quad (9)$$

where $\mathbf{\Sigma}_x = n_1 \mathbf{\Sigma}_1 + n_2 \mathbf{\Sigma}_2$. Here, $n_1$ and $n_2$ are the cardinalities of classes $\mathcal{C}_1$ and $\mathcal{C}_2$ respectively.

Unfortunately, there is no closed form solution to obtain a linear classifier that minimises $p_e$ when the homoscedasticity asssumption does not hold [Anderson and Bahadur, 1962]. Instead, the following solutions are proposed by [Marks and Dunn, 1974]:

$$\mathbf{w} = \big[s_1 \mathbf{\Sigma}_1 + s_2 \mathbf{\Sigma}_2\big]^{-1} (\bar{\mathbf{x}}_1 - \bar{\mathbf{x}}_2)$$
$$w_0 = \mu_1 - s_1 \sigma_1^2 = \mu_2 + s_2 \sigma_2^2 \quad (10)$$

where $s_1$ and $s_2$ scalars. However, the optimal values of $s_1$ and $s_2$ are unknown and have to be chosen by systematic trial and error.

Rather than adjust two different parameters, one may control only one parameter by multiplying $\mathbf{w}$ and $w_0$ proportionally by the scalar $s_1 + s_2$ so that one obtains $\mathbf{w}$ in the form:

$$\mathbf{w} = \big[s \mathbf{\Sigma}_2 + (1-s) \mathbf{\Sigma}_1\big]^{-1} (\bar{\mathbf{x}}_1 - \bar{\mathbf{x}}_2) \quad (11)$$

Note that if $s_1 + s_2$ is positive, then the discrimination criterion given by (1) is not changed. We show in Section 4 that this is the case. Still, the optimal value of $s$ has to be obtained by systematic trial and error as is done in [Anderson and Bahadur, 1962].

In order to obtain this optimal $s$ and thus find the linear discriminant that minimises (6), the following optimisation procedure is described in [Fukunaga, 2013] as the "theoretical approach":

1. Change $s$ from 0 to 1 with small step increments $\Delta s$.
2. Evaluate **w** as given by:
$$\mathbf{w} = \left[ s\mathbf{\Sigma}_1 + (1-s)\mathbf{\Sigma}_2 \right]^{-1} (\bar{\mathbf{x}}_1 - \bar{\mathbf{x}}_2) \qquad (12)$$
3. Evaluate $w_0$ as given by:
$$w_0 = \frac{s\mu_2\sigma_1^2 + (1-s)\mu_1\sigma_2^2}{s\sigma_1^2 + (1-s)\sigma_2^2} \qquad (13)$$
4. Compute the probability of misclassification $p_e$.
5. Choose **w** and $w_0$ that minimise $p_e$.

However, the solution obtained thus is only locally optimal, and may not be the global optimal solution, as the parameter $s$ is restricted to the range $[0, 1]$. As we show in the next section, $s$ can actually be unbounded, for instance, when there is a class imbalance and the means are not well-separated.

## 3 The Class Imbalance Problem

The class imbalance problem arises when the number of objects in one class far exceeds the cardinality of the other classes. Such datasets are often encountered in anomaly detection applications like medical diagnosis, falls detection [Ojetola *et al.*, 2015] in remote health monitoring or machine health monitoring, where a "fault" state is not as probable as the "normal" state of the system. As the prior probabilities are often estimated from the cardinality of each class, data imbalance in binary classification results in the case where $\pi_1 \gg \pi_2$ or $\pi_1 \ll \pi_2$. If the means are not well separated, class imbalance has the effect of shifting the threshold $w_0$ well beyond the boundaries of the projected class means $\mu_1$ and $\mu_2$. This then tends to skew the classification accuracy in favour of the majority class.

We refer the reader to [Xue and Titterington, 2008] for a more detailed study of the class imbalance problem.

## 4 Optimality Conditions

First, we obtain the necessary first-order equations for the minimisation of $p_e$, i.e.

$$\nabla p_e(\tilde{\mathbf{w}}) = \left[ \frac{\partial p_e}{\partial \mathbf{w}^T}, \frac{\partial p_e}{\partial w_0} \right]^T = \mathbf{0} \qquad (14)$$

where $\tilde{\mathbf{w}} = [\mathbf{w}^T, w_0]^T$. From (6), it can be shown that:

$$\frac{\partial p_e}{\partial \mathbf{w}} = \frac{1}{\sqrt{2\pi}} \left[ -\pi_1 e^{-z_1^2/2} \left( \frac{\sigma_1 \bar{\mathbf{x}}_1 + z_1 \mathbf{\Sigma}_1 \mathbf{w}}{\sigma_1^2} \right) + \pi_2 e^{-z_2^2/2} \left( \frac{\sigma_2 \bar{\mathbf{x}}_2 + z_2 \mathbf{\Sigma}_2 \mathbf{w}}{\sigma_2^2} \right) \right] \qquad (15)$$

It can similarly be shown from (6) that,

$$\frac{\partial p_e}{\partial w_0} = \frac{\pi_1}{\sqrt{2\pi}} \left( \frac{1}{\sigma_1} e^{-z_1^2/2} \right) - \frac{\pi_2}{\sqrt{2\pi}} \left( \frac{1}{\sigma_2} e^{-z_2^2/2} \right) \qquad (16)$$

Now, equating the gradient $\nabla p_e(\mathbf{w}, w_0)$ to zero, the following set of equations are obtained:

$$\left( \frac{\pi_2 z_2}{\sigma_2^2} e^{-z_2^2/2} \mathbf{\Sigma}_2 - \frac{\pi_1 z_1}{\sigma_1^2} e^{-z_1^2/2} \mathbf{\Sigma}_1 \right) \mathbf{w} = \left( \frac{\pi_1}{\sigma_1} e^{-z_1^2/2} \right) \bar{\mathbf{x}}_1 - \left( \frac{\pi_2}{\sigma_2} e^{-z_2^2/2} \right) \bar{\mathbf{x}}_2 \qquad (17)$$

$$\frac{\pi_1}{\sigma_1} e^{-z_1^2/2} = \frac{\pi_2}{\sigma_2} e^{-z_2^2/2} \qquad (18)$$

Substituting (18) into (17) yields:

$$\left( \frac{z_2}{\sigma_2} \mathbf{\Sigma}_2 - \frac{z_1}{\sigma_1} \mathbf{\Sigma}_1 \right) \mathbf{w} = (\bar{\mathbf{x}}_1 - \bar{\mathbf{x}}_2) \qquad (19)$$

Then the vector **w** can be given by:

$$\mathbf{w} = \left( \frac{z_2}{\sigma_2} \mathbf{\Sigma}_2 - \frac{z_1}{\sigma_1} \mathbf{\Sigma}_1 \right)^{-1} (\bar{\mathbf{x}}_1 - \bar{\mathbf{x}}_2) \qquad (20)$$

It will be noted however that **w** as given by (20) is still in terms of $w_0$ through $z_1$ and $z_2$. Therefore, an explicit representation of $w_0$ in terms of **w** is needed from (18) to substitute in $z_1$ and $z_2$ in (20).

Now, solving for $w_0$ from (18) results in the following quadratic:

$$\frac{z_2^2}{2} - \frac{z_1^2}{2} - \ln\left( \frac{\tau \sigma_1}{\sigma_2} \right) = 0 \qquad (21)$$

which can be simplified to:

$$\left( \frac{w_0 - \mu_2}{\sigma_2} \right)^2 - \left( \frac{w_0 - \mu_1}{\sigma_1} \right)^2 - 2\ln\frac{\tau \sigma_1}{\sigma_2} = 0, \qquad (22)$$

where $\tau = \pi_2/\pi_1$. If $\tau$ is defined and not equal to zero, and $\sigma_1^2 \neq \sigma_2^2$, (22) can be shown to have the following solutions:

$$w_0 =$$
$$\frac{\mu_2 \sigma_1^2 - \mu_1 \sigma_2^2 \pm \sigma_1 \sigma_2 \sqrt{(\mu_1 - \mu_2)^2 + 2(\sigma_1^2 - \sigma_2^2) \ln\left( \frac{\tau \sigma_1}{\sigma_2} \right)}}{\sigma_1^2 - \sigma_2^2} \qquad (23)$$

Having two solutions of $w_0$, it is necessary that we find which one to substitute into (20).

We now consider a second-order necessary condition for the minimisation of $p_e$, namely,

$$\frac{\partial^2 p_e}{\partial w_0^2} \geq 0 \qquad (24)$$

From (16), it can be shown that:

$$\frac{\partial^2 p_e}{\partial w_0^2} = \frac{\pi_1}{\sqrt{2\pi}} \left( -\frac{z_1}{\sigma_1^2} e^{-z_1^2/2} \right) + \frac{\pi_2}{\sqrt{2\pi}} \left( \frac{z_2}{\sigma_2^2} e^{-z_2^2/2} \right) \qquad (25)$$

If we plug in (18) (from which we get the stationary point of (16)), we end up with the following inequality:

$$\frac{z_2}{\sigma_2} \geq \frac{z_1}{\sigma_1} \qquad (26)$$

Thus, if $\mathbf{w}$ is to be a local minima of $p_e$, it is necessary that (26) is satisfied.

In the following, we show that the two solutions of $w_0$ in (23) cannot both satisfy (26), that is, only one of the solutions corresponds to a local minimum. We then make the choice for that $w_0$ which corresponds to a local minimum in order to substitute into (20).

**Theorem 1.** *Let $w_0^+$ and $w_0^-$ be the two distinct solutions of (23), then $w_0^+$ and $w_0^-$ cannot both satisfy (26) given that $\sigma_1 \neq \sigma_2$.*

*Proof.* Let

$$\beta = \sqrt{(\mu_1 - \mu_2)^2 + 2(\sigma_1^2 - \sigma_2^2)\ln\left(\frac{\tau\sigma_1}{\sigma_2}\right)} \quad (27)$$

and let

$$w_0^+ = \frac{\mu_2\sigma_1^2 - \mu_1\sigma_2^2 + \sigma_1\sigma_2\beta}{\sigma_1^2 - \sigma_2^2} \quad (28)$$

Then

$$\frac{z_2}{\sigma_2} = \frac{(\mu_2 - \mu_1)\sigma_2 + \beta\sigma_1}{\sigma_2(\sigma_1^2 - \sigma_2^2)}, \quad \frac{z_1}{\sigma_1} = \frac{(\mu_2 - \mu_1)\sigma_1 + \beta\sigma_2}{\sigma_1(\sigma_1^2 - \sigma_2^2)} \quad (29)$$

Suppose that $w_0^+$ satisfies (26), then

$$\frac{(\mu_2 - \mu_1)\sigma_2 + \beta\sigma_1}{\sigma_2(\sigma_1^2 - \sigma_2^2)} \geq \frac{(\mu_2 - \mu_1)\sigma_1 + \beta\sigma_2}{\sigma_1(\sigma_1^2 - \sigma_2^2)} \quad (30)$$

i.e.,

$$\frac{\beta\sigma_1^2}{\sigma_1^2 - \sigma_2^2} \geq \frac{\beta\sigma_2^2}{\sigma_1^2 - \sigma_2^2} \quad (31)$$

This implies that $\sigma_1^2/(\sigma_1^2 - \sigma_2^2) > \sigma_2^2/(\sigma_1^2 - \sigma_2^2)$ since $\beta$ is a positive scalar.

Consider now $w_0^-$ given as:

$$w_0^- = \frac{\mu_2\sigma_1^2 - \mu_1\sigma_2^2 - \sigma_1\sigma_2\beta}{\sigma_1^2 - \sigma_2^2} \quad (32)$$

Then

$$\frac{z_2}{\sigma_2} = \frac{(\mu_2 - \mu_1)\sigma_2 - \beta\sigma_1}{\sigma_2(\sigma_1^2 - \sigma_2^2)}, \quad \frac{z_1}{\sigma_1} = \frac{(\mu_2 - \mu_1)\sigma_1 - \beta\sigma_2}{\sigma_1(\sigma_1^2 - \sigma_2^2)} \quad (33)$$

In order for (26) to be satisfied, it can be shown, similar to (31), that

$$\frac{-\beta\sigma_1^2}{\sigma_1^2 - \sigma_2^2} \geq \frac{-\beta\sigma_2^2}{\sigma_1^2 - \sigma_2^2} \quad (34)$$

which can be simplified to give $1 \leq 0$. Since this conclusion is false, only $w_0^+$ satisfies (26). $\square$

Having obtained $w_0$ that satisfies (26), and substituting into (20), we find that $\mathbf{w}$ still has no closed-form solution from (20). However, if we let $s_2 = z_2/\sigma_2$ and $s_1 = z_1/\sigma_1$, then we get the optimal $\mathbf{w}$ as given by (20) in the form of (10). Note that by multiplying $\mathbf{w}$ and $w_0$ by the same positive scalar $c$, the discrimination criterion as given by (1) is not changed.

Suppose $c = (\sigma_1 z_2 - \sigma_2 z_1)/\sigma_1\sigma_2$, then $\mathbf{w}$ can also be written in the form of (11) where

$$s = \frac{\sigma_1 z_2}{\sigma_1 z_2 - \sigma_2 z_1}. \quad (35)$$

$c$ is positive due to the inequality of (26).

We make note of two important points at this stage:

1. $s$ is unbounded. As long as the inequality of (26) is satisfied, $s$ can range from $-\infty$ to $\infty$. In the special case where $\sigma_1 z_2 > 0$ and $\sigma_2 z_1 < 0$, then $s$ is constrained in the interval $(0, 1)$. However, if the threshold $w_0$ falls outside the two projected means $\mu_1$ and $\mu_2$, as happens often when the population means are not well separated under class imbalance, then $z_2$ and $z_1$ have the same sign. In such a case $s$ falls outside the interval $(0, 1)$. For instance, if $z_2 > 0$ and $z_1 > 0$, then $s > 1$, and if $z_2 < 0$ and $z_1 < 0$, then $s < 0$.

   It will be observed then that since $s$ is constrained to $(0, 1)$ in the HLD procedure of [Fukunaga, 2013], the $\mathbf{w}$ obtained is only locally optimal, because the optimal $s$ may lie outside this interval. Moreover, if $s \in (0, 1)$, then the threshold $w_0$ as given by (13) is a convex combination of $\mu_1$ and $\mu_2$, and thus always falls within the interval $(\mu_2, \mu_1)$.

2. $s$ is not a free parameter. As we see from (35), $s$ is a function of $\mathbf{w}$ and $w_0$. Therefore, rather than arbitrary choosing $s$ from systematic trial and error, one may obtain improving values of $s$ by improving upon an initial choice of $\mathbf{w}$ and $w_0$. We describe this procedure in the next section.

## 5 Optimisation

As there is no closed form solution to $\mathbf{w}$ as given by (20), an iterative procedure is needed for the optimisation of $\mathbf{w}$. However, as $s$ has been shown to be unbounded, it cannot be varied between 0 and 1 with small step increments. The alternative of varying $s$ from $-\infty$ to $\infty$ with small steps $\Delta s$ is a computationally demanding task. Moreover, randomly trying different values of $s$ is a rather unguided procedure. For this reason, we take a gradient descent approach to minimising $\mathbf{w}$ and $w_0$ starting from some initial choice of $\mathbf{w}$ and $w_0$. i.e.

For $i = 0$ to some maximum number of iterations $I$:

$$\mathbf{w}^{i+1} = \mathbf{w}^i - \alpha\frac{\partial p_e}{\partial \mathbf{w}^i} \quad (36)$$

$$w_0^{i+1} = w_0^i - \alpha\frac{\partial p_e}{\partial w_0^i} \quad (37)$$

Every step along this gradient descent path therefore corresponds to an improved value of $s$ due to the fact that $s$ is a function of $\mathbf{w}$ and $w_0$.

In order to improve upon the rate of convergence of the gradient descent procedure, the optimal value of the threshold $w_0$ that satisfies (26) may be calculated for every step in $\mathbf{w}$ from (23), rather than using (37).

Still, the choice of the initial solution to be used in (36) leads one to consider the convexity or otherwise of the objective function. If the objective function is convex, then we are guaranteed that for every choice of the initial solution, we converge on the same final solution. However, $p_e$ is known to be non-convex and is characterised by multiple local minima [Anderson and Bahadur, 1962]. For this reason, the gradient descent algorithm is performed for several initial choices of $\mathbf{w}$ for $R$ runs. As we know the general form of the optimal solution of $\mathbf{w}$ as given by (11), we may choose the initial solutions in accordance. Each run of the gradient descent procedure then converges on a local minimum of $p_e$ giving us an ensemble of $R$ classifiers. One can then find the classifier among this ensemble that minimises the probability of misclassification $p_e$.

The proposed algorithm is explained in much detail in Algorithm 1.

---

**Algorithm 1** Proposed

1: Input: $\mathcal{D}_1$ and $\mathcal{D}_2$
2: Evaluate $\bar{\mathbf{x}}_1, \bar{\mathbf{x}}_2, \mathbf{\Sigma}_1, \mathbf{\Sigma}_2$
3: **for** $r \leftarrow 1$ to $R$ **do**
4:     Randomly choose $s$.
5:     Initialise $\mathbf{w}$ as $\mathbf{w} = \left[s\mathbf{\Sigma}_2 + (1-s)\mathbf{\Sigma}_1\right]^{-1}(\bar{\mathbf{x}}_1 - \bar{\mathbf{x}}_2)$
6:     **while** Stopping criteria are not satisfied **do**
7:         Evaluate $\mu_1, \mu_2, \sigma_1, \sigma_2$.
8:         Calculate $w_0$ from (23) ensuring that it satisfies the inequality (26).
9:         Evaluate $z_1, z_2$
10:        Evaluate the probability of misclassification $p_e$.
11:        Calculate $\frac{\partial p_e}{\partial \mathbf{w}}$ using (15).
12:        Update $\mathbf{w}$, i.e. $\mathbf{w}^{t+1} = \mathbf{w}^t - \alpha \frac{\partial p_e}{\partial \mathbf{w}}$
13:     **end while**
14: **end for**
15: Choose $\mathbf{w}$ and $w_0$ having the minimum $p_e$.

---

It would be noted that this approach does in no way constrain the optimal value of $s$ in the interval $[0, 1]$, and is therefore suited for class imbalance.

After the algorithm has converged to give the optimal weight $\mathbf{w}^*$ and threshold $w_0^*$, one may calculate the optimal value of $s$, $s*$ as:

$$s^* = \frac{\sigma_1^* z_2^*}{\sigma_1^* z_2^* - \sigma_2^* z_1^*}. \qquad (38)$$

where

$$\sigma_1^* = \mathbf{w}^{*T}\mathbf{\Sigma}_1\mathbf{w}^*, \quad \sigma_2^* = \mathbf{w}^{*T}\mathbf{\Sigma}_2\mathbf{w}^*$$
$$z_1^* = \frac{w_0^* - \mathbf{w}^{*T}\bar{\mathbf{x}}_1^*}{\sigma_1^*}, \quad z_2^* = \frac{w_0^* - \mathbf{w}^{*T}\bar{\mathbf{x}}_2^*}{\sigma_2^*} \qquad (39)$$

Thus, $\mathbf{w}^*$ may equivalently be calculated as:

$$\mathbf{w}^* = \left[s^*\mathbf{\Sigma}_1 + (1-s^*)\mathbf{\Sigma}_2\right]^{-1}(\bar{\mathbf{x}}_1 - \bar{\mathbf{x}}_2) \qquad (40)$$

In the event of a class imbalance, the error rate might be skewed toward the majority class. For this reason, the threshold $w_0$ may be varied to improve the correct classification of the minority class at the expense of the overall classification accuracy. In this manner, the false positive rate (FPR) may be plotted against the true positive rate (TPR) to obtain the receiver operating characteristics (ROC). (We have assumed that the positive class is the minority class). Note however that the weight vector $\mathbf{w}$ is a function of the threshold $w_0$ itself, as can be seen from (20) or (40). Thus, the optimal weight vector, $\mathbf{w}^*$ is obtained for the optimal threshold $w_0^*$. Therefore varying the threshold to obtain the FPR and TPR necessarily causes $\mathbf{w}$ to be suboptimal. To correct this, $\mathbf{w}$ out to be optimised for every choice of $w_0$. For this reason, we now express $\mathbf{w}$ as a function of $w_0$ thus:

$$\mathbf{w}(w_0) = \left[s'\mathbf{\Sigma}_1 + (1-s')\mathbf{\Sigma}_2\right]^{-1}(\bar{\mathbf{x}}_1 - \bar{\mathbf{x}}_2) \qquad (41)$$

where

$$s' = \frac{\sigma_1^* z_2'}{\sigma_1^* z_2' - \sigma_2^* z_1'}. \qquad (42)$$

and

$$z_1' = \frac{w_0 - \mathbf{w}^{*T}\bar{\mathbf{x}}_1^*}{\sigma_1^*}, \quad z_2' = \frac{w_0 - \mathbf{w}^{*T}\bar{\mathbf{x}}_2^*}{\sigma_2^*} \qquad (43)$$

## 6 Experimental Validation

We conduct two sets of experiments. In our first, we validate the proposed algorithm on an artificial dataset having two classes and satisfying the normality assumption, but with different covariance matrices. The two classes have the following Gaussian parameters:

$$\bar{\mathbf{x}}_2 = [3.86, 3.10, 0.84, 0.84, 1.64, 1.08, 0.26, 0.01]^T,$$
$$\mathbf{\Sigma}_2 = \text{diag}(8.41, 12.06, 0.12, 0.22, 1.49, 1.77, 0.35, 2.73)$$
$$\bar{\mathbf{x}}_1 = \bar{\mathbf{x}}_2 - 0.3, \quad \mathbf{\Sigma}_1 = \mathbf{I} \qquad (44)$$

The above parameters are those of Fukunaga's two-class heteroscedastic data with the exception of $\bar{\mathbf{x}}_1$ which is equal to $\mathbf{0}$ in [Fukunaga, 2013]. In our experiments, we have made $\bar{\mathbf{x}}_1 = \bar{\mathbf{x}}_2 - 0.3$ so that the means are not well-separated. We randomly generate 100 points in Class $\mathcal{C}_1$ and $100f$ points in Class $\mathcal{C}_2$, where $f \in \{1, 2, 4, 8, 16, 32\}$ to simulate varying degrees of unbalanced data. We then perform 10 trials of 10-fold cross validation, and show results for the average misclassification rate on the test dataset.

In the proposed algorithm, we set $R = 1$ using Fisher's solution (8) as our initial solution, and we run the gradient descent procedure for $I = 1000$ iterations, terminating prematurely when the norm of (17) is less than $\epsilon = 10^{-6}$. We compare the misclassification rate of our algorithm with LDA [Izenman, 2009], RHLD [Anderson and Bahadur, 1962] (with 1000 trials) and HLDA [Fukunaga, 2013] (with step rate $\Delta s = 0.001$). The parameters for our proposed algorithm, RHLD and HLD were optimised via cross-validation. The results are shown in Figure 1.

In our second experiment, we have used five real-world datasets for which the fraction of the minority class range between $0.77\%$ and $42.03\%$. The characteristics of the datasets are shown in Table 1.

For the Yeast dataset, the positive minority examples belong to class ME2, while the rest are negative examples. Also, for the Abalone dataset, the positive minority examples belong to class 19, and the rest are negative examples.

We proceed by performing $10-$fold cross-validation. Since the classification accuracy may be skewed toward the majority class, we have provided the area under the ROC curve, AUC, as our evaluation metric. We have presented the average AUC and training time over all 10 folds for each of the 5 datasets for our algorithm compared with LDA, QDA, RHLD and HLD using the same parameters as in the first experiment. Additionally, we compare these LDA approaches with the linear SVM [Hsu and Lin, 2002]. These results are shown in Tables. 2-6.

Table 1: Characteristics of Datasets

| Dataset | $N$ | Minority (%) | Majority (%) |
|---|---|---|---|
| Liver | 345 | 42.03 | 57.97 |
| Diabetes | 768 | 34.90 | 65.10 |
| WpBC | 194 | 23.71 | 76.29 |
| Yeast-ME2 | 1484 | 3.44 | 96.56 |
| Abalone-19 | 4177 | 0.77 | 99.23 |

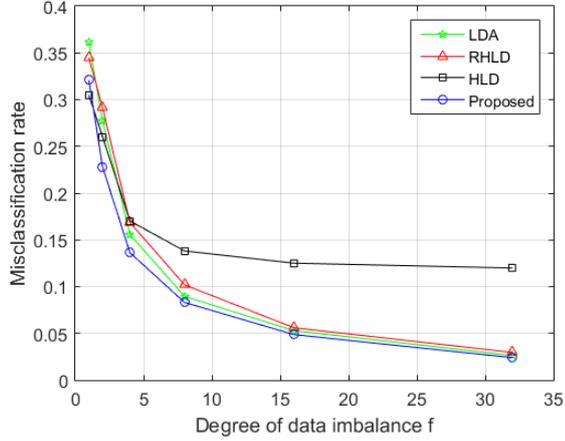

Figure 1: Effect of varying the degree of class imbalance on misclassification rate

Table 2: Liver disorders dataset: Average AUC and training times

| Algorithm | AUC | Time (s) |
|---|---|---|
| QDA | 0.7033 | 0.01 |
| HLD | 0.7041 | 0.15 |
| RHLD | 0.7041 | 0.13 |
| LDA | 0.7051 | 0.01 |
| SVM | 0.7205 | 2.77 |
| Proposed | **0.7376** | 0.60 |

Table 3: Pima Indians Diabetes: Average AUC and training times

| Algorithm | AUC | Time (s) |
|---|---|---|
| QDA | 0.8110 | 0.01 |
| HLD | 0.8317 | 0.15 |
| RHLD | 0.8316 | 0.14 |
| LDA | 0.8311 | 0.01 |
| SVM | 0.8289 | 10.94 |
| Proposed | **0.8443** | 0.61 |

Table 4: Wisconsin Breast Cancer (Prognosis) dataset: Average AUC and training times

| Algorithm | AUC | Time (s) |
|---|---|---|
| QDA | 0.6274 | 0.01 |
| HLD | 0.8257 | 0.46 |
| RHLD | 0.8257 | 0.44 |
| LDA | 0.8248 | 0.04 |
| SVM | 0.7029 | 3.23 |
| Proposed | **0.9034** | 1.31 |

Table 5: Yeast dataset: Average AUC and training times

| Algorithm | AUC | Time (s) |
|---|---|---|
| QDA | 0.8472 | 0.01 |
| HLD | 0.8790 | 0.16 |
| RHLD | 0.8785 | 0.14 |
| LDA | 0.8799 | 0.01 |
| SVM | 0.7825 | 0.03 |
| Proposed | **0.9133** | 0.42 |

Table 6: Abalone dataset: Average AUC and training times

| Algorithm | AUC | Time (s) |
|---|---|---|
| QDA | 0.7697 | 0.01 |
| HLD | 0.8608 | 0.15 |
| RHLD | 0.7586 | 0.14 |
| LDA | 0.8610 | 0.01 |
| SVM | 0.5853 | 0.08 |
| Proposed | **0.9011** | 0.42 |

## 7 Discussion of Results

The results in Fig. 1 show that for lower values of the parameter $f$, i.e. for more balanced heteroscedastic data, our proposed algorithm outperforms LDA in terms of the misclassification rate. This is due to the fact that by assuming equal covariance, LDA does not minimise the probability of misclassification for heteroscedastic data. As the degree of class imbalance $f$ increases, however, the performance of our proposed algorithm and that of LDA converge. This is because in the limit of $f$, the majority class $C_2$ becomes far more probable than $C_1$. Therefore, the decision rule depends less on the covariance matrices, but dictated by the threshold $w_0$. Since the threshold obtained by LDA as given by (9) is unbounded and depends on the ratio of the prior probabilities, LDA is able to track the optimal $w_0$ in the limit of $f$ and yields a satisfactory performance in terms of the misclassification rate. This result confirms the conclusions in [Xue and Titterington, 2008] that unbalanced data have no negative effect on LDA in terms of the error rate.

For the HLD [Fukunaga, 2013], a reverse relationship is seen. For more balanced data, the HLD compares favourably with our proposed algorithm as the parameter $s$ likely lies in the interval $(0,1)$. However, as $f$ increases, the performance of HLD becomes comparatively poorer due to the fact that by restricting $s$ to the interval $(0,1)$, the threshold $w_0$ found by (13) is bounded between the projected class means. Therefore, the HLD, unlike our proposed algorithm and LDA, is unable to track the optimal $w_0$ in the limit of $f$, when one class becomes far more probable than the other.

With regards to the RHLD [Anderson and Bahadur, 1962], the relationship between the misclassification rate and the degree of data imbalance $f$ is not straightforward. As $s$ has been shown to be unbounded, obtaining the optimal $s$ from the interval $(-\infty, \infty)$ by systematic trial and error, as the RHLD does, can be tedious. Therefore, a finite interval $[a, b]$ is used. If the range of this interval is large, the tedium in obtaining the optimal $s$ still exists. However, if the range is too small, the optimal $s$ may not be found, as it may exist outside the specified interval $[a, b]$. An optimum choice of the interval $[a, b]$ for all values of $f$ results in the performance of the RHLD as shown in Fig. 1.

For the real-world datasets, the proposed algorithm is shown to be consistently superior in terms of the AUC to all the other algorithms, including the linear SVM. This is due to the fact that, unlike the original LDA or existing HLDA approaches, the optimal $s$ in our proposed algorithm is not constant, but is a function of the threshold $w_0$. Therefore, the weight vector $\mathbf{w}$ is optimised for every choice of $w_0$, as $w_0$ is varied to obtain the ROC. The degree of data imbalance. In contrast, the optimal value of $s$ is a constant given by $s = N_2/N$ for LDA, where $N_2$ is the cardinality of class 2; for RHLD and HLD, $s$ is detached from $\mathbf{w}$ and $w_0$ and made a free parameter which is optimised with respect to the probability of error $p_e$.

## 8 Conclusion

In this paper, we have shown that existing heteroscedastic linear discriminants are either suboptimal under class imbalance or have no principled optimisation procedure. We have thus presented a method for obtaining the Bayes optimal linear classifier for heteroscedastic LDA. Our approach, unlike those in the literature, has been shown to be robust against class imbalance and is obtained by an efficient gradient descent optimisation procedure. Experimental results based on one artificial and five real-world datasets show that the proposed algorithm achieves superior performance in terms of the misclassification rate and the AUC as compared to the SVM, LDA, QDA and other existing heteroscedastic LDA approaches. Moreover, the short training time of our algorithm makes it very well-suited for large-data applications.

Our future work is focused on going beyond Gaussian families of probability distribution to obtain the Bayes error for more general distributions. Alternatively, work is on-going in obtaining a kernel transformation that implicitly maps classes of known non-normal distribution into a feature space where the classes are nearly-normally distributed.

# Under-Sampling the Minority Class to Improve the Performance of Over-Sampling Algorithms in Imbalanced Data Sets

**Romero F. A. B. de Morais and Germano C. Vasconcelos**
Center for Informatics
Federal University of Pernambuco
{rfabm, gcv}@cin.ufpe.br

## Abstract

Imbalanced distribution of class instances occur in many real-world problems. Since many classifiers usually expect balanced class distribution, imbalanced data sets may represent a hindrance to their learning process. Over-sampling algorithms are a common solution to balance class distribution, through random replication or synthesis of new examples in the minority class. However, current over-sampling algorithms usually use all available examples in the minority class to synthesise new examples, which may include noisy data. In this work, we propose $k$-INOS, a new algorithm to make over-sampling algorithms more robust to noisy examples in the minority class. $k$-INOS is an adaptation of the concept of neighbourhood of influence and works as a wrapper around any over-sampling algorithm. Experimentation conducted with 50 benchmark data sets, 7 over-sampling techniques and 5 classifiers, showed, particularly for weak classifiers, the proposed algorithm significantly increased performance for most over-sampling algorithms according to different metrics.

## 1 Introduction

Imbalanced distribution of class instances is a situation found in many real-world applications such as diagnosis of diseases [Krawczyk *et al.*, 2016], software fault identification [Wang and Yao, 2013], churn prediction [Burez and Van den Poel, 2009], fraud detection [Wei *et al.*, 2013] and many others [Haixiang *et al.*, 2016]. As some common classifiers expect distribution of examples among data classes to be balanced, they tend to poorly learn and recognise the minority class. However, proper modelling and recognition of all classes is necessary and, even more important, in many cases, with respect to the minority class. For instance, authorising a fraudulent credit card transaction or diagnosing a cancerous patient as healthy is likely to incur into serious consequences and high costs for both the person involved and stakeholders.

Sampling algorithms are a common approach to deal with the uneven distribution of classes in data sets. Their popularity can be attributed to their role in enhancing classification performance and usual independence from the chosen classifier [López *et al.*, 2013]. In general, sampling algorithms either reduce the size of the majority class (under-sampling) or increase the size of the minority class (over-sampling). The opposite, over-sampling the majority class or under-sampling the minority class, is rarely seen [Japkowicz, 2001]. The common explanation is performing sampling in either opposite directions would further imbalance the data set. Moreover, as few examples are available in the minority class, under-sampling it would allegedly discard important information.

SMOTE [Chawla *et al.*, 2002] is a prominent over-sampling algorithm in the literature. It synthesises new examples in the line segment joining neighbouring examples of the minority class. A known problem with SMOTE is its sensitivity to noise in the minority class. Noise is meant here as isolated cases, minority examples mostly surrounded by majority examples, or potentially inconsistent data points present in the data set. To address the drawbacks of SMOTE many adaptations were developed. Borderline-SMOTE (BDL-SMOTE) [Han *et al.*, 2005] modifies SMOTE to synthesise new examples only between the minority examples close to the decision border. ADASYN [He *et al.*, 2008] synthesises more examples near the minority class examples considered "hard to learn". For a given minority class example, learning hardness is measured according to the number of neighbours from the majority class. The greater the number of neighbours from the majority class the harder to learn. Safe-Level-SMOTE (SL-SMOTE) [Bunkhumpornpat *et al.*, 2009] defines a safeness level for each example in the minority class and synthesises new examples according to those safe levels. Safeness of an example is then defined as the proportion of neighbours belonging to the minority class. Recently, [Sáez *et al.*, 2015] created SMOTE-IPF, which adds a post-processing step to SMOTE to remove noisy examples in the over-sampled data, using an ensemble-based filtering approach.

Here we propose $k$-INOS ($k$-Influential Neighbourhood for Over-Sampling), an algorithm to make general over-sampling algorithms more robust to noise in the minority class. The algorithm adapts the concept of "neighbourhood of influence" [Ha and Lee, 2016] to remove examples from the minority class, before applying an over-sampling algorithm to a data set. Afterwards, the removed examples are put back into the over-sampled data set. Hence, the proposed algo-rithm under-samples the minority class before over-sampling it, and can be interpreted as a wrapper to over-sampling algorithms.

Experimentation results showed the algorithm was able to increase Accuracy and $F_1$-Measure performances for most classifiers and over-sampling algorithms. In fact, weak classifiers used experienced the most significant improvements. Positive results were also observed for AUROC metric and strong classifiers, although with only part of them being statistically significant in this case.

## 2 Related Work

As far as we are concerned, under-sampling the minority class before over-sampling the minority class has never been carried out. Thus, it is included here a review of works that perform pre- or post-processing steps to remove noisy examples in imbalanced data sets, devised to improve the performance of over-sampling algorithms.

In [Batista *et al.*, 2004], a comprehensive experimentation with 10 sampling algorithms was performed. Amongst the algorithms, two of them were modifications of SMOTE [Chawla *et al.*, 2002] to include a data cleaning step. The first method (SMOTE + ENN) uses the Edited Nearest Neighbours (ENN) algorithm [Wilson, 1972] to remove noisy examples present in both classes, after over-sampling the data with SMOTE. ENN works by removing examples misclassified by the $k$-nearest neighbours. The second method (SMOTE + Tomek Links) works similarly to SMOTE + ENN, but performs cleaning by removing examples (from both classes) belonging to "Tomek Links" (see [Tomek, 1976]). Tomek Links are formed by pairs of examples from different classes that are nearest neighbours to each other. For experimentation, the work employed C4.5 decision tree as base classifier, 10 imbalanced data sets, and AUROC as the performance metric. Results showed both methods, SMOTE + ENN and SMOTE + Tomek Links, outperformed SMOTE.

SMOTE-IPF was proposed by [Sáez *et al.*, 2015] to remove noisy examples from both classes, after over-sampling the imbalanced data set with SMOTE. To remove noisy examples, it employs an ensemble-based method called Iterative-Partioning Filter (IPF). Experimentation employed the C4.5 decision tree as base classifier, 9 imbalanced data sets, and AUROC as the performance metric. In addition, SMOTE-IPF was also tested on artificial imbalanced data sets and on versions with added noise. Results showed SMOTE-IPF outperformed SMOTE, as well as SMOTE + ENN and SMOTE + Tomek Links. Further experiments reported with different classifiers, showed SMOTE-IPF had more significant improvements when combined with weak classifiers.

Although efforts have been made to remove noise in imbalanced data sets, from both majority and minority classes, the focus is always after the over-sampling process. The method described here breaks this pattern and removes minority examples before the over-sampling process.

## 3 $k$-INOS

Before introducing the proposed algorithm, named here $k$-INOS, we review the definition of "neighbourhood of influence" as defined by Ha and Lee [2016].

The $k$-influential neighbourhood ($k$-IN) of an example (data point) $x$ can be defined in terms of the $k$-nearest neighbours ($k$-NN) and the reverse $k$-nearest neighbours ($k$-RNN) of $x$. In the following definitions, consider a data set $D$ and an example $x \in D$:

**Definition 1.** ($k$-nearest neighbours of $x$): The set containing the $k$ points from $D\setminus\{x\}$ closest to $x$ is called the set of $k$-nearest neighbours of $x$ and is represented by $k$-NN$(x)$.

**Definition 2.** (reverse $k$-nearest neighbours of $x$): The set containing the points from $D\setminus\{x\}$ with example $x$ as one of their $k$-NNs is called the set of reverse $k$-nearest neighbours and is represented by $k$-RNN$(x)$.

**Definition 3.** ($k$-influential neighbourhood of $x$):

$$k\text{-IN}(x) = \{x\} \cup k\text{-NN}(x) \cup k\text{-RNN}(x)$$

The definition of $k$-IN$(x)$ can be interpreted as the region in feature space where example $x$ has direct influence, through its $k$-NNs, and indirect influence, through its $k$-RNNs. As a result, for an example $x$, the bigger the size of the associated $k$-IN the denser the region where the example is located in feature space.

$k$-INOS uses an adaptation of the concept of neighbourhood of influence by removing examples from the minority class with small neighbourhoods of influence.

- First, compute the $k$-IN for each example in the minority class and remove from $k$-IN all examples that belong to the majority class, in order to avoid association of large neighbourhoods of influence to noisy examples from the minority class;

- Second, remove from the data set, the minority examples with a $k$-IN smaller than a given threshold;

- Third, apply an over-sampling algorithm to the whole filtered data set;

- Finally, return the removed examples in the second step to the over-sampled data set, in order to avoid losing information from the minority class.

A more detailed description of $k$-INOS is shown in Algorithm 1. Two parameters control the behaviour of $k$-INOS: $k$ and $\tau$. The number of neighbours, $k$, controls how far the influence of an example should be measured in feature space. Large values are likely to associate large and similar regions to all examples, whereas small values may fail to associate examples in an existing appropriate region. The threshold, $\tau$, is responsible for eliminating examples from the minority class with small regions of influence. Large values may remove most of the points in the minority class, which is not desirable, while small values are more conservative and more likely to remove only noisy examples. For instance, $\tau = 2$ removes only the minority examples isolated from the other minority examples in feature space.

**Example 1.** Figure 1 contrasts the behaviour of SMOTE [Chawla *et al.*, 2002] and the proposed algorithm in an artificial imbalanced data set. Noisy examples were added to the minority class in order to show how both algorithms handle noise. SMOTE fails to handle noise and synthesises new examples along the line segments, by joining the noisy and the genuine examples. On the other hand, $k$-INOS successfully filters the noisy examples before applying SMOTE, thus avoiding generation of new data near the noisy examples.

---

Algorithm 1: $k$-INOS

Objective

Perform pre- and post-processing steps before and after application of an over-sampling algorithm. The pre-processing step filters noisy examples from the minority class while the post-processing step returns the filtered examples to the over-sampled data set.

Input

- $D$: An imbalanced data set.
- $k$: Number of neighbours to compute the $k$-IN.
- $\tau$: Threshold value used to filter examples from the minority class. Examples from the minority class in which the size of $k$-IN is $< \tau$ are removed.
- $\phi$: An over-sampling algorithm.

Output

- $D*$: Balanced version of input data set $D$.

Algorithm

1. **$k$-IN**: For each example in the minority class compute its $k$-IN as in **Definition 3**. Remove from $k$-IN all examples that belong to the majority class.
2. **Pre-Process**: Filter from $D$ all examples from the minority class for which the size of $k$-IN is $< \tau$.
3. **Over-Sample**: Perform over-sampling by passing the filtered data set to $\phi$.
4. **Post-Process**: Return the filtered examples to the final over-sampled data set $D*$.

---

## 4 Experimental Setup

The experimentation framework devised to evaluate $k$-INOS included 50 imbalanced data sets, 7 over-sampling algorithms, 5 classifiers, and 5 performance metrics. For each combination of data set, over-sampling algorithm and classifier, a 5×2-fold cross-validation was employed to compute mean values for the 5 metrics used. A Wilcoxon signed-ranks test [Wilcoxon, 1945] was used to analyse the difference in performance between the over-sampling algorithm with and without $k$-INOS, as recommended by Demšar [2006]. Significance level of 5% and one-sided hypothesis test were employed.

### 4.1 Data Sets

The 50 data sets used mainly come from the UCI Machine Learning Repository [Lichman, 2013], the Keel-dataset Repository [Alcalá et al., 2010], and the Promise Repository [Menzies et al., 2015]. Apart from the data sets from the Promise Repository, all sets were cleaned up with removal of (in order) constant features, repeated examples, and inconsistent examples (same values for the features but different classes). The data sets from the Promise Repository were already cleaned up by Shepperd et al. [2013].

Whenever a new sampling algorithm is published it rarely employs the same data sets used in previous works [He et al., 2008; Bunkhumpornpat et al., 2009; Zhang and Li, 2014]. Moreover, in several cases, claims about algorithms are made based on a few data sets, which is clearly undesirable. Therefore, data sets were selected here following a few criteria. Data sets were required to have at least 40 examples in the minority class, no missing values, at most 60 features (all of them numerical), and at least an imbalance ratio (IR) of 1.5. The IR is defined as the ratio between number of examples in the majority class and number of examples in the minority class. The list of selected data sets is shown in Table 1.

### 4.2 Over-Sampling Algorithms

In order to assess the performance gains of $k$-INOS on over-sampling algorithms, 7 over-sampling algorithms were considered: Random Over-Sampling (ROS), SMOTE [Chawla et al., 2002], Borderline-SMOTE-{1, 2} (BDL-SMOTE-{1, 2}) [Han et al., 2005], ADASYN [He et al., 2008], Safe-Level-SMOTE (SL-SMOTE) [Bunkhumpornpat et al., 2009], and RWO-Sampling (RWO) [Zhang and Li, 2014]. All algorithms were set to balance class distribution and parameters for each algorithm were set as follows:

- SMOTE: $k$ = 5.
- BDL-SMOTE-{1, 2}: $m$ = 11, $k$ = 5.
- ADASYN: $d_{th}$ = 0.75, $k$ = 5.
- SL-SMOTE: $k$ = 5.

With the only exception of parameter $m$ from Borderline-SMOTE-{1, 2}, where the suggested value is not precisely defined, the particular parameter values employed were the default values suggested by the authors. Note $d_{th}$ parameter in ADASYN is irrelevant in this context case, as the value 1/IR in our selected data sets is always $< 0.75$.

As for $k$-INOS, parameter values where set to $k$ = 11 and $\tau$ = 3, after initial experimentation. Although $k$-INOS and most over-sampling algorithms have a parameter named $k$, the algorithms use the parameter in different ways. While $k$-INOS uses $k$ to compute the $k$-IN, the other over-sampling algorithms use $k$ to compute the $k$-NN set.

### 4.3 Classifiers

A diverse set of 5 classifiers, usually referred to as weak (C4.5 Decision Tree (DT), $k$-Nearest Neighbours ($k$-NN)) and strong (Gradient Boosting Machine (GBM), Random Forest (RF), Support Vector Machine (SVM) with Gaussian Kernel) were considered for performance comparison. The implementations below were employed:

- DT: R interface to *Weka's* J48 [Hornik et al., 2009].
- GBM: R interface to the *eXtreme Gradient Boosting* system [Chen and Guestrin, 2016].

Table 1: Data sets employed. All data sets were cleaned up by removing (in order) constant features, repeated examples, and inconsistent examples. Numbers shown were computed after cleaning. Entries are sorted by increasing values of IR.

| Name | Source | # Examples | # Features | # Majority | # Minority | IR |
|---|---|---|---|---|---|---|
| Spambase | [Lichman, 2013] | 4204 | 57 | 2528 | 1676 | 1.5084 |
| Ionosphere | [Lichman, 2013] | 350 | 33 | 225 | 125 | 1.8000 |
| glass1 | [Alcalá et al., 2010] | 213 | 9 | 137 | 76 | 1.8026 |
| fourclass | [Cieslak et al., 2012] | 862 | 2 | 555 | 307 | 1.8078 |
| MC2" | [Menzies et al., 2015] | 125 | 39 | 81 | 44 | 1.8409 |
| ecoli-0_vs_1 | [Alcalá et al., 2010] | 220 | 6 | 143 | 77 | 1.8571 |
| Pima Indians Diabetes | [Lichman, 2013] | 768 | 8 | 500 | 268 | 1.8657 |
| MAGIC Gamma Telescope | [Lichman, 2013] | 18905 | 10 | 12332 | 6573 | 1.8762 |
| SVMguide1 | [Cieslak et al., 2012] | 3025 | 4 | 2000 | 1025 | 1.9512 |
| QSAR biodegradation | [Lichman, 2013] | 1052 | 41 | 698 | 354 | 1.9718 |
| glass0 | [Alcalá et al., 2010] | 213 | 9 | 144 | 69 | 2.0870 |
| Vertebral Column (2C) | [Lichman, 2013] | 310 | 6 | 210 | 100 | 2.1000 |
| yeast1 | [Alcalá et al., 2010] | 1453 | 8 | 1028 | 425 | 2.4188 |
| phoneme | [Cieslak et al., 2012] | 5395 | 5 | 3818 | 1577 | 2.4211 |
| PC5" | [Menzies et al., 2015] | 1711 | 38 | 1240 | 471 | 2.6327 |
| Haberman's Survival | [Lichman, 2013] | 277 | 3 | 204 | 73 | 2.7945 |
| vehicle2 | [Alcalá et al., 2010] | 846 | 18 | 628 | 218 | 2.8807 |
| Blood Transfusion Service Center | [Lichman, 2013] | 471 | 4 | 353 | 118 | 2.9915 |
| Parkinsons | [Lichman, 2013] | 195 | 22 | 147 | 48 | 3.0625 |
| glass-0-1-2-3_vs_4-5-6 | [Alcalá et al., 2010] | 213 | 9 | 162 | 51 | 3.1765 |
| vehicle0 | [Alcalá et al., 2010] | 846 | 18 | 647 | 199 | 3.2513 |
| ecoli1 | [Alcalá et al., 2010] | 336 | 7 | 259 | 77 | 3.3636 |
| JM1" | [Menzies et al., 2015] | 7782 | 21 | 6110 | 1672 | 3.6543 |
| ecoli2 | [Alcalá et al., 2010] | 336 | 7 | 284 | 52 | 5.4615 |
| segment0 | [Alcalá et al., 2010] | 2084 | 18 | 1788 | 296 | 6.0405 |
| PC4" | [Menzies et al., 2015] | 1287 | 37 | 1110 | 177 | 6.2712 |
| PC3" | [Menzies et al., 2015] | 1077 | 37 | 943 | 134 | 7.0373 |
| estate | [Cieslak et al., 2012] | 5211 | 12 | 4591 | 620 | 7.4048 |
| yeast3 | [Alcalá et al., 2010] | 1453 | 8 | 1291 | 162 | 7.9691 |
| yeast-2_vs_4 | [Alcalá et al., 2010] | 489 | 8 | 438 | 51 | 8.5882 |
| pendigits | [Cieslak et al., 2012] | 10992 | 16 | 9850 | 1142 | 8.6252 |
| yeast-0-2-5-6_vs_3-7-8-9 | [Alcalá et al., 2010] | 977 | 8 | 879 | 98 | 8.9694 |
| yeast-0-2-5-7-9_vs_3-6-8 | [Alcalá et al., 2010] | 977 | 8 | 879 | 98 | 8.9694 |
| yeast-0-3-5-9_vs_7-8 | [Alcalá et al., 2010] | 504 | 8 | 454 | 50 | 9.0800 |
| satimage | [Cieslak et al., 2012] | 6430 | 36 | 5805 | 625 | 9.2880 |
| yeast-0-5-6-7-9_vs_4 | [Alcalá et al., 2010] | 527 | 8 | 476 | 51 | 9.3333 |
| page-blocks0 | [Alcalá et al., 2010] | 5383 | 10 | 4879 | 504 | 9.6806 |
| vowel0 | [Alcalá et al., 2010] | 988 | 13 | 898 | 90 | 9.9778 |
| PC1" | [Menzies et al., 2015] | 705 | 37 | 644 | 61 | 10.5574 |
| covtype | [Cieslak et al., 2012] | 38500 | 10 | 35753 | 2747 | 13.0153 |
| oil | [Cieslak et al., 2012] | 937 | 48 | 896 | 41 | 21.8537 |
| letter | [Cieslak et al., 2012] | 18668 | 16 | 17912 | 756 | 23.6931 |
| winequality-red-4 | [Alcalá et al., 2010] | 1359 | 11 | 1306 | 53 | 24.6415 |
| compustat | [Cieslak et al., 2012] | 13655 | 20 | 13135 | 520 | 25.2596 |
| yeast4 | [Alcalá et al., 2010] | 1453 | 8 | 1402 | 51 | 27.4902 |
| ism | [Cieslak et al., 2012] | 7845 | 6 | 7592 | 253 | 30.0079 |
| mammography | [Woods et al., 1993] | 7847 | 6 | 7594 | 253 | 30.0158 |
| yeast5 | [Alcalá et al., 2010] | 1453 | 8 | 1410 | 43 | 32.7907 |
| MC1" | [Menzies et al., 2015] | 1988 | 38 | 1942 | 46 | 42.2174 |
| shuttle-2_vs_5 | [Alcalá et al., 2010] | 3316 | 9 | 3267 | 49 | 66.6735 |

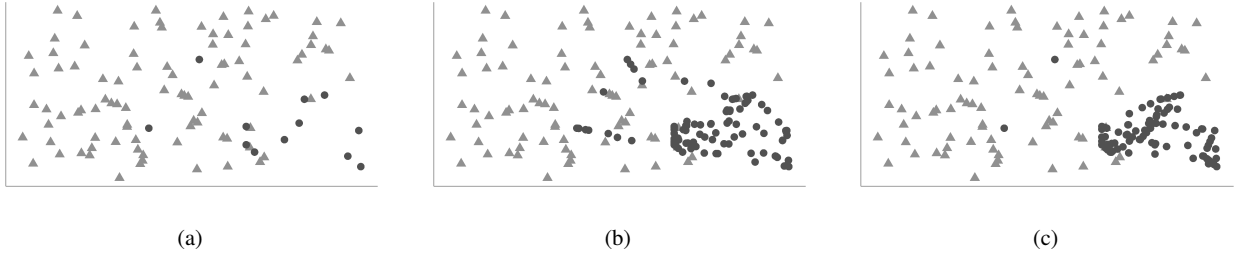

Figure 1: The effect of $k$-INOS to avoid multiplication of noise. (a) An imbalanced data set containing noisy examples in the minority class. (b) The behaviour of SMOTE with $k$ = 5 and set to balance the classes. (c) The behaviour of $k$-INOS with $k$ = 7, $\tau$ = 2, and SMOTE as the over-sampling algorithm (same parameters as in (b)).

- $k$-NN: R's *kknn* package [Schliep and Hechenbichler, 2016].
- RF: R's *randomForest* package [Liaw and Wiener, 2002].
- SVM: R's *kernlab* package [Karatzoglou *et al.*, 2004].

All classifiers had parameters set to their recommended values. The only exceptions were $k$-NN, which had $k$ = 3, and used Euclidean distance, and GBM, which had the number of boosting iterations set to 50, after some experimentation.

### 4.4 Performance Metrics

There is still no consensus on performance metrics to assess classifier performance on imbalanced data sets [He and Garcia, 2009; Haixiang *et al.*, 2016]. Single metrics alone can sometimes lead to biased or misled interpretations. For example, if all input instances are classified as belonging to the majority class – an inappropriate decision strategy in real world scenarios – measuring Accuracy would lead to high rates depending on the IR of the data set. For instance, if a data set has an IR of 5, simply assigning all examples to the majority class would produce an accuracy of approximately 83%. Since any metric alone can fail to capture important information about performance, for consistency of analysis, a set of metrics composed of Accuracy, $F_1$-Measure, AUROC, Recall and Precision was considered in the experimentation.

## 5 Results and Discussion

Results of the Wilcoxon signed-ranks tests, for each metric, measured with respect to the 50 data sets, are displayed in Table 2. Each row corresponds to a classifier and each column corresponds to an over-sampling algorithm. There are five symbols in the table: ▽, ▼, △, ▲, and "-". Symbol △ means the proposed algorithm increased the performance of the over-sampling algorithm in more data sets than decreased, whereas a ▽ means the opposite. Symbol "-" represents a tie in the number of increases and decreases. Black triangles mean the increase (or decrease) was statistically significant.

**Accuracy**: as seen in Table 2, results measured with respect to accuracy showed a remarkable improvement in the vast majority of classifier-oversampling algorithm combinations, after application of $k$-INOS. In 31 out of 35 situations, accuracy was improved, with 27 of them with statistical significance according to the Wilcoxon test. In only 4 out of 35 there was a decrease in performance but none of them was significant. To complete the analysis it is seen accuracy was not changed in 2 out of 35 situations.

**$F_1$-Measure**: with respect to $F_1$-Measure, most classifiers and over-sampling algorithms showed performance enhancement. The 3-NN classifier had increase in six out of seven over-sampling algorithms, with five of them being significant. Only the ROS algorithm had performance reduction but it was not significant. The DT and GBM classifiers also showed performance increase for most over-sampling techniques. DT had an increase in five out of seven algorithms, with three of them being significant. A tie occurred for BDL-SMOTE-2, with the same number of increases and decreases, and a not significant decrease occurred for ADASYN. GBM, similarly to 3-NN, had an improvement in six out of seven over-sampling algorithms, however with only two of them being significant. RWO was the only algorithm to have a performance decrease but it was not significant. For the SVM classifier, improvements were moderate. There were four increases, two decreases, and one tie in performance, but none of them were significant. RF classifier was the only one to mainly attain performance losses. The only increase was observed for RWO and it was significant.

The majority of over-sampling algorithms experienced performance improvements regardless of the classifier. The RWO, SMOTE, and SL-SMOTE algorithms reached improvements in four out of five classifiers, with three of them significant for RWO; and two of them significant for SMOTE and SL-SMOTE. BDL-SMOTE-1 and BDL-SMOTE-2 had improvements in three out of five classifiers with one of them significant. ROS and ADASYN only attained two improvements with one of them significant.

**Precision and Recall**: $F_1$-Measure is the harmonic mean between recall and precision. Hence, it is interesting to analyse recall and precision rates individually. Clearly, $k$-INOS sacrificed recall for boosting precision. In almost all cases, recall performance was decreased, in many of them significantly. The only increases happened for combinations of classifier-oversampling algorithms DT-RWO, GBM-ADASYN, 3-NN-ADASYN, and RF-RWO. For the former two cases, increases were significant. On the other hand, precision performance significantly increased in almost all situations (34 out of 35).

Table 2: Results of the Wilcoxon signed-ranks test. Black arrows indicate statistically significant increase/decrease in performance whereas transparent arrows indicate increase/decrease but not statistically significant. "-" represents ties.

|  | ROS | SMOTE | BDL-SMOTE-1 | BDL-SMOTE-2 | ADASYN | SL-SMOTE | RWO |
|---|---|---|---|---|---|---|---|
| **Accuracy** | | | | | | | |
| DT | ▲ | ▲ | ▲ | ▲ | ▲ | ▲ | ▲ |
| GBM | ▲ | ▲ | ▲ | ▲ | △ | ▲ | ▽ |
| 3-NN | - | ▲ | ▲ | ▲ | ▲ | ▲ | ▲ |
| RF | △ | ▲ | ▲ | ▲ | △ | ▽ | △ |
| SVM | ▲ | ▲ | ▲ | ▲ | - | ▲ | ▲ |
| **$F_1$-Measure** | | | | | | | |
| DT | ▲ | ▲ | △ | - | ▽ | △ | ▲ |
| GBM | △ | △ | △ | △ | ▲ | ▲ | ▽ |
| 3-NN | ▽ | ▲ | ▲ | ▲ | △ | ▲ | ▲ |
| RF | ▼ | ▽ | ▽ | - | ▽ | - | ▲ |
| SVM | ▽ | △ | - | △ | ▽ | △ | △ |
| **AUROC** | | | | | | | |
| DT | ▼ | ▽ | ▽ | - | △ | △ | △ |
| GBM | △ | △ | △ | △ | ▽ | △ | ▽ |
| 3-NN | ▽ | △ | △ | ▲ | △ | △ | ▲ |
| RF | △ | ▽ | ▽ | △ | ▼ | ▽ | ▽ |
| SVM | ▼ | ▽ | ▼ | ▼ | ▼ | ▼ | ▼ |
| **Recall** | | | | | | | |
| DT | ▼ | ▼ | ▼ | ▼ | ▽ | ▽ | ▲ |
| GBM | ▽ | ▼ | ▼ | ▼ | ▲ | ▽ | ▽ |
| 3-NN | ▽ | ▼ | ▽ | - | △ | ▽ | ▽ |
| RF | ▼ | ▼ | ▼ | ▼ | ▽ | ▼ | △ |
| SVM | ▼ | ▼ | ▼ | ▼ | ▼ | ▼ | ▼ |
| **Precision** | | | | | | | |
| DT | ▲ | ▲ | ▲ | ▲ | ▲ | ▲ | ▲ |
| GBM | ▲ | ▲ | ▲ | ▲ | △ | ▲ | △ |
| 3-NN | ▽ | ▲ | ▲ | ▲ | ▲ | ▲ | ▲ |
| RF | ▲ | ▲ | ▲ | ▲ | △ | △ | ▲ |
| SVM | ▲ | ▲ | ▲ | ▲ | ▲ | ▲ | ▲ |

The only decrease occurred for combination 3-NN-ROS, and was not significant. Although there was a general decrease in recall, the gains in $F_1$-Measure performance showed the increase in precision outweighed the degradation observed in recall. Indeed, as the harmonic mean tends towards the minimum of its arguments, to observe an increase in $F_1$-Measure while having a decrease in recall means the growth in precision far outweighed the reduction in recall. For example, in the case of 3-NN-SMOTE, the average increase in precision was 2.2 times the decrease in recall.

**AUROC**: according to AUROC, 3-NN experienced the most positive results. It attained performance improvements in six out of seven over-sampling algorithms, with significant increases for BDL-SMOTE-2 and RWO. The only decrease occurred for ROS and was not significant. The GBM classifier had performance increase in five out of seven oversampling algorithms, however, none of them were significant. The DT classifier showed three increases, three decreases and one tie. Only the decrease for ROS was significant. RF and SVM classifiers had performance losses in most cases. While RF had two non-significant increases for ROS and BDL-SMOTE-2, SVM had six significant decreases.

**Weak and Strong Classifiers**: performance improvements were consistent for $F_1$-Measure. DT, GBM, and 3-NN classifiers had increases in five, six, and six out of seven oversampling algorithms, with three, two, and five of them being significant, respectively. SVM, on the other hand, had many improvements but none of them significant, and RF had one significant improvement. It seems the proposed algorithm is more effective when combined with weak classifiers, in this case, DT and 3-NN. Since SVM and RF are considered stronger classifiers and more robust to noise, they were probably capable of dealing with noise in the minority class. GBM is also a strong classifier, however, yet produced consistent gains as a result of $k$-INOS. This is consistent with the results of [Sáez *et al.*, 2015] with SMOTE-IPF, where more significant performances were reached with weak classifiers.

Figure 2 represents all data sets, according to results for different combinations of classifier-oversampling algorithms, before and after application of $k$-INOS. "Black" points indicate performance increase (+), "grey" means performance is degraded (-) and "light grey" represents ties (=), after application of $k$-INOS. In case (a), 3-NN is analysed together with SMOTE and, for most data sets, there is performance improvement. In case (b), DT and RWO are employed and still there are many more situations of enhancements. In case (c), results with the GBM classifier are similar to that seen in cases (a) and (b), whereas in case (d) the number of performance decreases is more evident with the RF classifier.

# 6 Conclusions

In this work, a new algorithm based on "neighbourhood of influence" has shown to increase the effectiveness of oversampling algorithms in many situations, by making them more robust to noise. The proposed method filters noisy

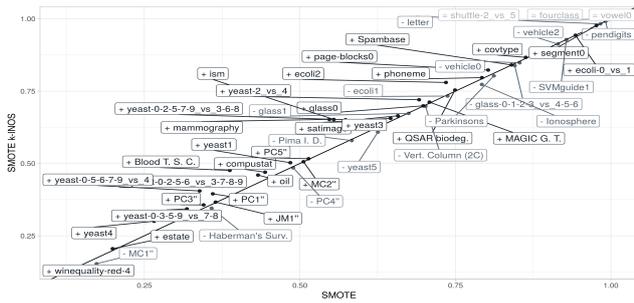
(a) 3-NN classifier and SMOTE.

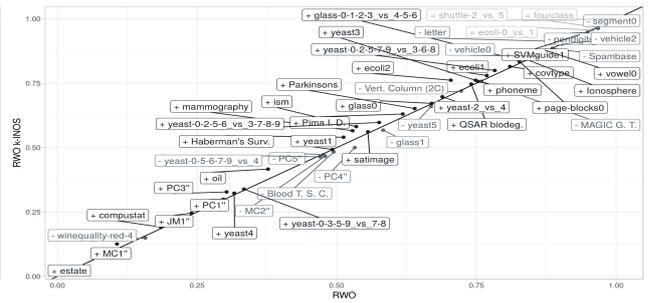
(b) DT classifier and RWO.

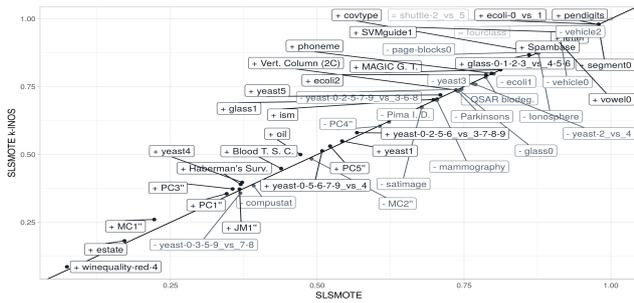
(c) GBM classifier and SL-SMOTE.

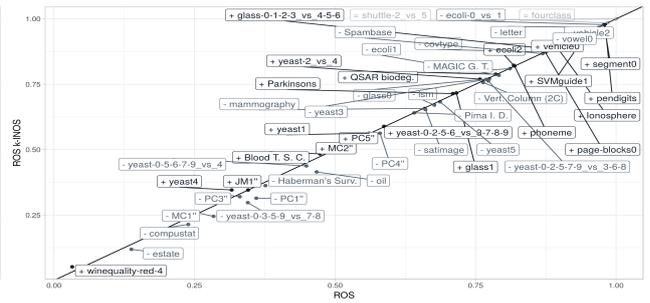
(d) RF classifier and ROS.

Figure 2: Performance comparison for $F_1$-Measure between over-sampling algorithms with and without application of $k$-INOS. Points above the line (black) represent improvement, below the line (grey) mean the opposite, and on the line represent ties (light grey). To aid visualisation, names are prefixed with a "+", "-", and "=" meaning increase, decrease, and tie respectively.

examples present in the minority class, applies an over-sampling algorithm, and returns the filtered examples to the over-sampled data set. The experimentation results showed recognition of the minority class, particularly for weak classifiers, significantly increased. Attractive results were also observed for the AUROC metric, however with only part of them being statistically significant. A recommendation emphasised in this work is the use of guidelines to select imbalanced data sets for experimentation and comparison of new algorithms. The community should pay more attention to this issue and work together towards standardisation. Regarding future works, there are many paths to be explored. It is important to further analyse the method so as to understand in which situations improvements are expected to occur. New over-sampling algorithms could also be devised using the concept of neighbourhood of influence. Currently, most algorithms simply rely on the nearest neighbours. The pre- and post-processing steps could be independently manipulated to refine the proposed algorithm. For instance, the post-processing step could be modified to not return the filtered examples to the over-sampled data set. The distribution of the sizes of the $k$-INs from the minority class could be used as a complexity measure for the imbalanced data set. We informally observed imbalanced data sets that were hard to achieve high $F_1$-Measures had many examples in the minority class with small neighbourhoods of influence. As reviewed in [López et al., 2013], this is likely to be related to the problem of small disjuncts and lack of density in the minority class.

## Acknowledgements

This work was partially supported by CNPq, grant 132229/2016-1.

# Dominance-based Rough Set Approach to Learn in the Presence of Class Imbalance

**Sarra BOUZAYANE**[1,3] **and Inès SAAD**[1,2]

[1]University of Picardie Jules Verne, MIS Laboratory, Amiens 80039, France
[2]Amiens Business School, Amiens 80039, France
[3]Higher Institute of Computer Science and Multimedia, MIRACL Laboratory, Sfax 3021,Tunisia
{sarra.bouzayane, ines.saad}@u-picardie.fr

**Abstract**

The DRSA (Dominance-based Rough Set Approach) is a multicriteria decision-making approach used for the supervised learning issues. The DRSA involves the human dimension for the "learning sample" construction; a thing that overcomes the problem of imbalanced data that is one of the important problems when training in data mining. Hence, the training sample in DRSA is based on the quality of the examples it contains rather than on their quantity. In this paper, we propose a method based on the DRSA for a supervised learning in a case of imbalanced data with multiple classes. This method consists of two phases: the first aims at inferring a preference model while the second consists of classifying each object in one of the predefined decision classes based on the previously inferred preference model. The first phase is made of three steps: the first is to identify assignment examples of objects, the second is to construct a coherent criteria family for the objects characterization and the third is to infer a preference model resulting in a set of decision rules. The method is applied in the context of MOOCs (Massive Open Online Courses).

## 1 Introduction

In the last decade, the amount of data collected and stored into databases has dramatically increased due to the software and the hardware advancements that led to the automation of the data collection. Consequently, traditional data analysis techniques have become unsuitable for the processing of such volumes of data and new techniques have emerged [Delen *et al.*, 2005]. In this work, we focus on the machine learning techniques that were mainly used for the prediction or the classification issues. Prediction is the reasoning from the current events to the future properties and events they will cause [Shanahan, 1989]. Recently, prediction has been given a special importance in the temporal reasoning literature and has been involved into several areas of the real world such as those of medicine [Ture *et al.*, 2005], psychology [Yaroslavsky *et al.*, 2013], finance [Shen and Tzeng, 2014], biology [Thusberg *et al.*, 2011], etc.

However, despite their effectiveness, the efficiency of the machine learning techniques degrades when faced to the problem of imbalanced data. This problem is the result of an unequal distribution between the classes used in the learning phase. In the literature, several techniques have been proposed to annihilate the effects of imbalanced data problem on the quality of classifiers and predictive models, such as the sampling methods [Krishna Veni and Sobha Rani, 2011], the cost-sensitive learning methods [Elkan, 2001; Ting, 2002] and Kernel-based methods [Vapnik, 1995].

Thus, in order not to use additional techniques when learning in the presence of class imbalance, we introduce a new approach called DRSA (Dominance-based Rough Set Approach) [Greco *et al.*, 2001] where the learning is no longer based on the sample size but rather on the quality of the examples herein. The approach DRSA requires the intervention of human decision makers for the construction of the representative sample of objects and the classification of each object in one of the predefined decision classes. According to [Elkan, May 28 2013], for predictive analytics to be successful, the training data must be representative of the test data.

In this paper, we propose a method based-DRSA used for a prediction purpose. It consists of two phases: the first aims at inferring a preference model while the second consists of classifying each object in one of the predefined decision classes based on the previously inferred preference model. The first phase is made of three steps: the first is to identify assignment examples of objects, the second is to construct a coherent criteria family for the objects characterization and the third is to infer a preference model resulting in a set of decision rules. The method is applied in the context of MOOCs (Massive Open Online Courses).

The paper is organized as follows. Section 2 introduces the problem of learning in the presence of class imbalance. Section 3 presents the method based-DRSA for a prediction objective. Section 4 explains the application of this method in the context of MOOC. Section 5 discusses the results. Section 6 concludes the paper and advances some prospects.

## 2 Problem of learning in the presence of class imbalance

There are several studies based on class imbalance problem and sampling methods in the literature. The imbalanced learning problem is concerned with the performance of learning algorithms in the presence of underrepresented data and severe class distribution skews. This problem can appear in several real world cases: for example to make prediction about the cancer disease using a randomly selected population, we obtain a majority class of unaffected people. Also, for fraud prediction in banks, the correct instances are much more than the impacted ones.

In this section, we explain the problem of imbalanced data in the domain of MOOCs (Massive Open Online Courses). The MOOC is an open course that is offered online to a massive number of learners who are enrolled for an educative purpose. The MOOC is based on the mutual assistance between the learners on the forum since the pedagogical team is small and therefore unable to accompany all the learners. However, the MOOC still suffer from an excessive dropout rate that reaches 90%. Thus, several prediction models are proposed in order to motivate the learners not to drop the course. Generally, three prediction objectives have been addressed: predicting the At-risk learners, predicting the discussions in which the tutor should intervene and predicting the resources that could help a learner. However, the two first objectives are strongly faced with the imbalanced data. On the one hand, since the dropout rate is 90%, we obtain a majority class of learners who are at risk to abandon the MOOC. On the other hand, since the tutor does not intervene much on the forum, we obtain much more of discussions that are not intervened.

Amnueypornsakul et al. [2014] proposed a predictive model for learner attrition using features that are related to the quiz attempt and those that capture interaction with the MOOC environment. The learner's activities are given to the model as actions sequence chained chronologically. According to the length of their action sequence, three classes of learners are defined: active learners who are the learners formally enrolled with an action sequence longer that two, inactive learners who are learners enrolled with an action sequence less than two and drop learners who are the learners formally not enrolled. For experiments, two modes are considered: Mod1 where inactive users are considered as dropped and Mod2 where inactive users are considered as active. Also, two cases are considered: specific case where the inactive users are excluded and the general case. The Support Vector Machine technique is used. This model was faced to an imbalanced data problem because the number of learners who still formally connected is more than double the size of those who dropped; which significantly affected the model performance. The authors used the oversampling techniques to overcome this problem but this was unsuccessful. Chaplot et al. [2015] proposed a model based on the Neural Network to predict a students' attrition in MOOCs. Other than the classical attributes, such as the number of clicks made by the learner and the weekly number of the forum pages viewed, authors integrated a sentiment score attribute. This is calculated using a lexicon-based sentiment analysis of the forum posts. Authors proved that the analysis of the students sentiment is an important indicator of their dropout intention. This model is highly imbalanced towards negative (will not dropout) class. To counter this problem, authors set the boundary for classification to the ratio of drop out data points to total number of data points in the training set. This means that if the value of output neuron is greater than this ratio, then the student is predicted to drop out in the next week, and vice-versa otherwise.

Chandrasekaran et al. [2015] proposed a model in order to predict exactly to which discussion the tutor must intervene with respect to his cognitive capacity (bandwidth). A supervised binary classifier based on regularized logistic regression is used. The classification process has incorporated a new attribute that is of the type of forum to which the discussion belongs (ie lecture, exam, homework, errata). The model training is based on previous threads, considering the intervened discussions as positive examples. However, in this case the problem of imbalanced data is significant since for a typical MOOC forum the intervened discussions are much less than the none intervened discussions (Majority class). To overcome this problem higher weights are granted for the minority class instances. These weights are learned by optimizing the maximum F-measure according to the training / validation set. 80% of the learning discussions are used for training / validation (to determine the class weight parameter) and the remaining 20% for testing. This problem remains sensitive to instructor's intervention preferences. The forum type attribute and the data weights improved performance to 45.54% (compared to 35.29 for [Chaturvedi et al., 2014]). Ren et al. [2016] developed a multiple linear regression model for predicting the performance of learners in the evaluation activities proposed by the MOOCs. The prediction is based on the number of clicks made by the learner. Two categories of learners were considered: those who completed all activities and those who partially finished them. Experiments carried out on three MOOCs showed that weekly model training inputs perform better than those of the last week alone. The authors used the accuracy and the F1-score to assess the prediction model efficiency since they are known to be a suitable metric for imbalanced datasets problem that strongly impacts the results.

Several studies have denounced the existence of an imbalanced data problem that impacts the quality of the prediction model by proposing -or not- a solution. However, many other works do not cite this problem despite the high probability that it causes the unsatisfactory obtained results. In what follows, we present a prediction method based on the DRSA that overcomes the imbalanced data problem.

## 3 Prediction method based on the approach DRSA

The approach DRSA [Greco et al., 2001] allows to compare objects through a dominance relation and takes into account the preferences of a decision maker to extract a Preference Model resulting in a set of decision rules. According to DRSA, a data table is a 4-tuple S= $\langle K, F, V, f \rangle$, where: K is a finite set of reference objects; F is a finite set of criteria; V= $\cup_{a \in F} V_a$ is the set of possible values of criteria; and $f$ denotes an information function $f : K \times F \longrightarrow V$ such that $f(x, a) \in V_a$, $\forall x \in K$, $\forall a \in F$. The criteria family F is often divided into a subset $C \neq \varnothing$ of condition attributes and a subset $D \neq \varnothing$ of decision attributes such that $C \cup D = F$ and

$C \cap D = \varnothing$. In this case, S is called a decision table. The decision attribute set D={d} is a singleton. The unique decision attribute d partitions K into a finite number of decision classes Cl= $\{Cl_n, n \in \{1,...,N\}\}$, such that each $x \in K$ belongs to one and only one class. Furthermore, we suppose that the classes are preference-ordered, i.e., for all r, s $\in \{1,...,N\}$ such that r>s, objects from Clr are preferred to objects from Cls.

In what follows, we present the method we propose to predict the decision class to which belongs each object in the information system. This method consists of two phases: the first concerns the inference of a preference model. The second targets the classification of the "Potential objects" using the previously inferred preference model. In a case of dynamic information system where information about objects comes periodically in portions, this method can be implemented periodically by incrementing the set of "Objects of reference". Let P= $\{P_1,... P_i,... P_t\}$ be the $t$ periods that should be considered throughout the training process.

## 3.1 Phase 1: Construction of a preference model of the period $P_i$

This phase comprises three steps. The first one concerns the construction of a representative training examples of objects for each of the predefined decision classes. The training set is called "Objects of reference". The second is the construction of a coherent family of criteria for the objects evaluation. The third targets the inference of the preference model.

### Step 1.1: Construction of a set of "Objects of reference"

It consists of defining a training sample including an adequate number of representative examples for each decision class. In order to comply with the terminology used in the DRSA, we call the training examples "Objects of reference". Since the human dimension is very involved in the validation process of the DRSA approach, the size of the set of "Objects of reference" must be considered. In fact, from a psychological viewpoint [Miller, 1956], a human decision maker is characterized by a channel capacity representing the upper limit to which he can match his responses to the stimuli we give him. So, to meet the channel capacity of the decision maker, we do not focus on the number of objects in the training sample but rather on their quality. Otherwise, a large training sample can degrade the quality of the decisions made by the decision maker, a thing which eventually affects the efficiency of the preference model. Thus, our goal is to build a set of "Objects of reference" both of a high quality and of a reasonable quantity in harmony with the decision maker's channel capacity in order to ensure an efficient set of decision rules.

In a case of dynamic information system, this training sample cannot be stable over many periods. Thus, upon each receipt of a new information portion of the period $P_i$, the decision maker has to define a new set $K'_i$ of "Objects of reference" that has to be directly appended as an additional set to the sets of "Objects of reference", $K_{i-1}$, of the previous periods. The set of "Objects of reference" of the period $W_i$ is thus $K_i = K'_i + K_{i-1}, \forall i \in \{1..t-1\}$. The size of the set of "Objects of reference" can change from a period to another (i.e. $|K'_i| \neq |K'_{i-1}|$). Yet, it should respect the channel capacity of the decision maker and represent each of the predefined decision classes as well. However, such an incremental add of the "Objects of reference" permits to give a training sample of high quality and more reasonable size which improves the prediction quality.

### Step 1.2: Construction of a coherent family of criteria

The criteria family would permit to characterize all objects in the information system. A criterion is a tool designed to evaluate and compare potential objects from a well-defined point of view. It is convenient to start from one or more indicators to define a set of criteria. Also, compared to an attribute or indicator, a criterion must allow the measuring of the decision maker's preference according to a personal viewpoint [Roy and Mousseau, 1996]. Two approaches are proposed to build a family of criteria: the top-down approach and the bottom-up approach. The top-down approach consists of defining one or more strategic objectives according to different points of view. The bottom-up approach is to build a family of criteria based on a list of indicators that can influence the opinion of decision-makers. The construction of the list of indicators must be determined either on the basis of a bibliographic analysis or on the basis of information gathered with the domain experts.

In this step some direct meetings have to be conducted with the decision maker in order to elicit his ordered preferential information for each criterion. For example, for an attribute $g_i$, we can define the increasing ordered scales: 1: Weak; 2: Average; 3: Strong.

### Step 1.3: Inference of a preference model

This step is made of two sub-steps: (i) the construction of the decision table, and (ii) the inference of the preference model. The decision table $S_i$ built during the period $P_i$ is a matrix whose rows form the set of the "$m$" "Objects of reference" identified in step 1.1 and contained in $K'_i$; and whose columns represent the "$p$" evaluation criteria constructed in step 1.2 and contained in $F_i$ (cf. Table1).

The content of this matrix is the evaluation function $f(O_{j,i}, g_k)$ of each object $O_{j,i} \in K'_i$ on each criterion $g_k \in F_i$ such that $i \in \{1..t\}$, $j \in \{1..m\}$ and $k \in \{1..p\}$. Variables $t$, $m$ and $p$ are respectively the number of considered periods, the size $|K'_i|$ of the "Objects of reference" set defined in the $i^{th}$ period and the size $|F_i|$ of the criteria family. Analogously, variables $O_{j,i}$ and $g_k$ are respectively the $j^{th}$ "Object of reference" in the set $K'_i$ and the $k^{th}$ criterion in the set $F_i$. We remind that $K'_i$ and $F_i$ are respectively the set of "Objects of reference" selected in the $i^{th}$ period and the family of criteria built. The last column in the decision table concerns the decision of the decision maker about the assignment of each "Object of reference" in one of the predefined decision classes (cf. Table 1).

Table 1: Example of a decision table

| $O_j$ | $g_1$ | ... | $g_k$ | ... | $g_p$ | $D_i$ |
|---|---|---|---|---|---|---|
| $O_1$ | $f(O_1, g_1)$ | ... | $f(O_1, g_k)$ | ... | $f(O_1, g_p)$ | Cli |
| $O_2$ | $f(O_2, g_1)$ | ... | $f(O_2, g_k)$ | ... | $f(O_2, g_p)$ | Cli |
| ... | ... | ... | ... | ... | ... | Cli |
| $O_k$ | $f(O_k, g_1)$ | ... | $f(O_k, g_k)$ | ... | $f(O_k, g_p)$ | Cli |
| ... | ... | ... | ... | ... | ... | Cli |
| $O_n$ | $f(O_n, g_1)$ | ... | $f(O_n, g_k)$ | ... | $f(O_n, g_p)$ | Cli |

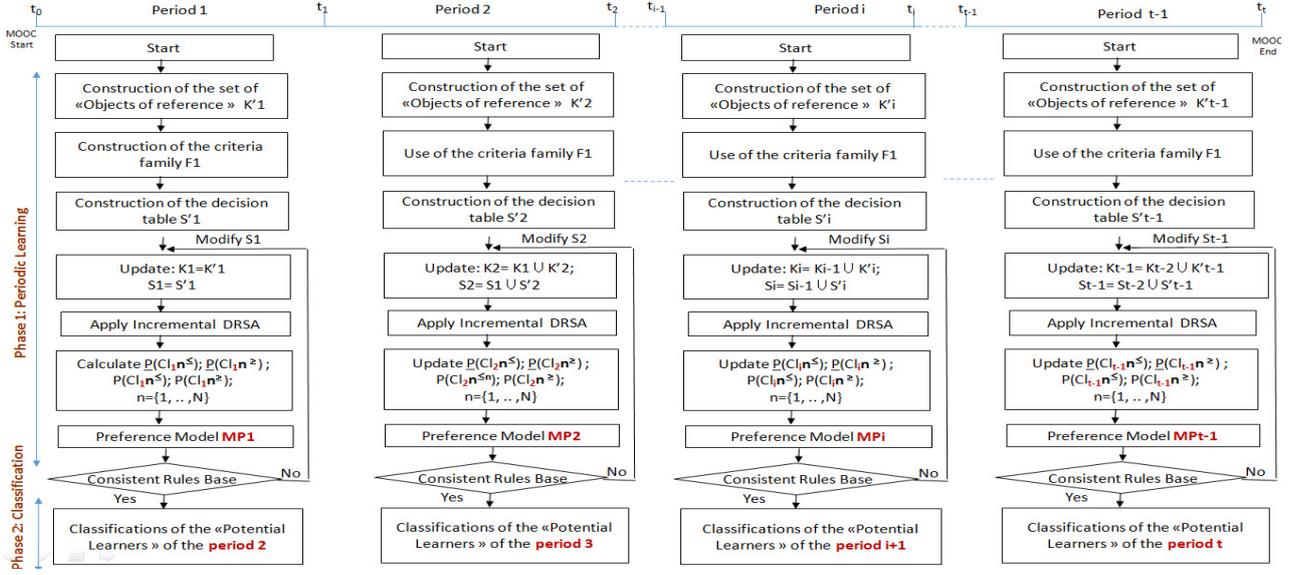

Figure 1: Periodic method for the prediction of "Leader learners" over weeks

The assignment decisions of each "Object of reference" should be based on his/her assessment values on the set of all criteria. The approach DRSA applies a dominance relation on the decision table in order to calculate the lower and the upper approximations (noted $\overline{P}(Cl_in^\geq)$, $\underline{P}(Cl_in^\geq)$, $\overline{P}(Cl_in^\leq)$ and $\underline{P}(Cl_in^\leq)$) of the upward and the downward unions of the decision classes [Greco et al., 2001]. The approximations will be given in input to the algorithm DOMLEM proposed by the approach DRSA. This algorithm outputs a preference model resulting in a set of decision rules that permits to classify each object in one of the decision classes.

In a case of periodic prediction, we proposed an algorithm to update the lower and the upper approximations when incrementing the set of "Objects of reference" in order to make the set of decision rules consistent. The algorithm is detailed in [Bouzayne and Saad, 2017]. The output of this phase is a preference model of the period $P_i$ that will be given as input to the phase 2.

### 3.2 Phase 2: Classification of the "Potential objects" in the period $P_{i+1}$

The second phase exploits the inferred decision rules of $P_i$ in order to predict in which decision class each "Potential object" is likely to belong in the period $P_{i+1}$. We mean by "Potential objects" those who are likely to be classified in one of the three decision classes.

This method runs periodically for a prediction purpose : the first phase runs at the end of each period $P_i$ while the second phase runs at the beginning of each period $P_{i+1}$. The two phases are chained in a way that each phase inputs the output of the previous one (cf. Figure 1).

## 4 Method application within a MOOC context

MOOCs (Massive Open Online Courses) are free online courses that are open, for a predefined duration, to anyone anywhere and at anytime as long as he has an internet connection. They are animated by a small pedagogical team and accessible by a massive number of learners with heterogeneous profiles who enrol to be trained on a given subject.

Since 2008, when the first MOOC has been coined by Downes and Siemens [Downes, 2013], the number of MOOCs and of their platform providers has rapidly increased around the world, especially in 2015 and 2016 where the total number of MOOCs reached respectively 4,200 and 4550 compared to 309 in 2013 [1]. However, despite their increasing popularity and proliferation, MOOCs suffer from a big limitation that is of the high dropout rate that usually reaches 90% [Yuan and Powell, 2013].

This excessive dropout rate has encouraged researchers and experts to think about methods for early predicting learners who are at risk to dropout a MOOC in order to help them not give it up. Several prediction models were proposed in literature based on many learning machine techniques. However, in the context of MOOC these techniques are strongly facing the issue of learning from a set of imbalanced data [He and Garcia, 2009]. Indeed, the excessive rate of abandon makes it obligatory to have a number of instances belonging to the majority class of "At-risk learners" which is much larger than the instances belonging to the other classes.

Hence, in what follows, we present how we applied the prediction method based on the approach DRSA (See Section 3) in the context of MOOCs without being faced with the

---

[1] http://www.onlinecoursereport.com/state-of-the-mooc-2016-a-year-of-massive-landscape-change-formassive-open-online-courses/

imbalanced data issue. The prediction process is based on three preference ordered decision classes:

- Cl1. The decision class of the "At-risk Learners" corresponding to learners who are likely to dropout the course in the next week of the MOOC.
- Cl2. The decision class of the "Struggling Learners" corresponding to learners who have some difficulties but still active on the MOOC environment and don't have the intention to leave it at least in the next week of the MOOC.
- Cl3. The decision class of the "Leader Learners" corresponding to learners who are able to lead a team of learners by providing them with an accurate and an immediate response.

We note also that a MOOC advances by week. Indeed, every week, the course materials are updated, new activities are proposed and a new forum space is created. Therefore, the behaviour of the learners and their evaluations on the criteria change from one week to another. So, at the end of each week a portion of the learners' information is received that must be taken into account for learning and prediction. To take into account this evolutionary character of the MOOC, the method is executed weekly. The objective is to predict the decision class to which a learner is likely to belong during the next week of the MOOC, based on his data (assessment values on all of the criteria) of the current week.

Obviously, since the dropout rate reaches 90%, the instances in the majority class of the "At-risk Learners" are always much more than instances in the other two other decision classes (Cl2 and Cl3). The imbalanced classes make the conventional machine learning techniques, such as the artificial neural networks or the simple multivariate statistical clustering procedures, face to the imbalance data issue [Amnueypornsakul et al., 2014; Chaplot et al., 2015]. However, this problem does not appear using the approach DRSA which is based on the quality of the learning sample rather than its size. The method was run as follows:

**step 1.1:** Given the massive number of learners involved in a MOOC, it is difficult to analyze and evaluate all of them. Hence, it is necessary to define a training sample, called "Learners of reference", including an adequate number of representative examples for each decision class; the decision class Cl1 of the "At-risk learners", the decision class Cl2 of the "Struggling learners" and the decision class Cl3 of the "Leader learners". The size of the "Learners of reference" set has to respect the channel capacity of the pedagogical team. At the end of each week $W_i$ such that $i \in \{1..t\}$ and $t$ is the number of weeks a MOOC holds, some meetings are held with the pedagogical team to select a sample $K_i'$ of $m$ representative examples of learners according to its preferences and based on its expertise. Thus, the size of the set of "Learners of reference" at the week $W_i$ is $|K_i| = m*i$.

**step 1.2:** Second, to construct a coherent family of criteria we applied the bottom-up approach. So, we started by identifying a list of indicators that would permit to characterize a learner within a MOOC. These indicators may be static or dynamic and would give sign about the learner's skills, profile, motivation, etc. Among the indicators mentioned in literature, we quote the previous MOOC experience that a learner acquired and his study level [Morris et al., 2015], the MOOC language mastery and the motivation to participate in the MOOC [Barak, 2015], the cultural background [Suriel and Atwater, 2012], the level of technical skills and the lack of time [Fini, 2009]. Besides the static data, experts also consider the dynamic data which are traced according to the learner's activity on the learning environment. In the e-learning field, Wolff et al. [2014] distinguished three types of activities that permit to predict the dropout of a learner. These activities are the access to a resource or to a course material; the publishing of a message on the forum and the access to the evaluation space. To validate a final family of criteria, some meetings have to be held with the pedagogical team of the concerned MOOC. The pedagogical team has to select $p$ criteria and to apply a preference order on each of them. For example, for the attribute "Study level", four increasing ordered scales are defined: 1: Scholar student; 2: High school student; 3: PhD Student; 4: Doctor. This step is detailed in [Bouzayane and Saad, 2017].

**Step 1.3:** Third, we constructed the Decision table $S_i$ and determined with the pedagogical team the decision vector that classifies each learner in $K_i'$ in one of the three decision classes, Cl1, Cl2 or Cl3. Then we apply the incremental algorithm [Bouzayane and Saad, 2017] to update the approximations in DRSA under the addition of a set of "Learners of reference" of the week $W_i$ to the learning sample of the week $W_{i-1}$. Finally, we run the algorithm DOMLEM proposed by the approach DRSA [Greco et al., 2001] in order to infer a preference model related to the week $W_i$.

**Phase 2:** At the beginning of each week $W_i$ such that $i \in \{2..t\}$, the preference model is applied on the set of all learners in the MOOC in order to identify the decision class to which they can belong.

# 5 Results and discussion

This section presents the results given when applying the method in a French MOOC offered by a Business School in France and broadcasted on a French platform. For reasons of anonymity, we were discreet about its name. The MOOC started with 2565 learners and lasted "$t= 5$" weeks. It required a weekly availability between one and three hours and did not require any prior knowledge. The first, the second and the fourth weeks were closed with a quiz while the third and the fifth were ended with a peer-to-peer assessment. Data was saved in a CSV (Comma-Separated Values) file. However, only data about 1535 learners were used in these experiments. Learners who have been neglected are those who have not completed the registration form. The dropout rate reaches 90% at the fifth week of the MOOC.

All algorithms in this paper are coded by Java and run on a personal computer with Windows 7, Intel (R) $Core^{TM}$ i3-3110M CPU @ 2.4 GHz and 4.0 GB memory.

An extract from the decision table built at the end of the first week is given in Table 2.

An extract from the decision rules inferred at the end of each week during the MOOC broadcast is given in Table 3.

Table 2: An excerpt from the decision table built at the end of the first week

| $L_{j,1}$ | $g_1$ | $g_2$ | $g_3$ | $g_4$ | $g_5$ | $g_6$ | $g_7$ | $g_8$ | $g_9$ | $g_{10}$ | $g_{11}$ | $D_1$ |
|---|---|---|---|---|---|---|---|---|---|---|---|---|
| $L_{1,1}$ | 2 | 2 | 3 | 3 | 1 | 2 | 2 | 2 | 1 | 1 | 1 | Cl1 |
| $L_{2,1}$ | 2 | 1 | 3 | 2 | 1 | 2 | 4 | 2 | 1 | 2 | 1 | Cl1 |
| $L_{3,1}$ | 2 | 1 | 3 | 1 | 1 | 0 | 5 | 2 | 1 | 5 | 4 | Cl2 |
| $L_{4,1}$ | 2 | 3 | 2 | 3 | 1 | 1 | 5 | 4 | 1 | 5 | 4 | Cl3 |

Table 3: Extract of decision rules inferred throughout the MOOC broadcast

| Week | Examples of decision rules extracted over weeks | Force |
|---|---|---|
| Week 1 | Rule 1.1: **If** $g_{4,1} \leq 1$ and $g_{12,1} \leq 1$ **Then** $L_2 \in Cl1^{\leq}$ | 28.57% |
|  | Rule 1.2: **If** $g_{2,1} \geq 3$ and $g_{7,1} \geq 5$ **Then** $L_2 \in Cl2^{\geq}$ | 26% |
| Week 2 | Rule 2.1: **If** $g_{10,2} \leq 3$ and $g_{11,2} \leq 3$ **Then** $L_3 \in Cl2^{\leq}$ | 52% |
|  | Rule 2.2: **If** $g_{5,2} \leq 0$ and $g_{9,2} \leq 1$ and $g_{11,2} \leq 3$ **Then** $L_3 \in Cl1^{\leq}$ | 40% |
| Week 3 | Rule 3.1: **If** $g_{9,3} \geq 4$ **Then** $L_4 \in Cl3^{\geq}$ | 16% |
|  | Rule 3.2: **If** $g_{7,3} \leq 3$ and $g_{10,3} \leq 1$ and $g_{12,3} \leq 2$ **Then** $L_4 \in Cl1^{\leq}$ | 40% |
| Week 4 | Rule 4.1: **If** $g_{9,4} \leq 1$ **Then** $L_5 \in Cl2^{\leq}$ | 27.27% |
|  | Rule 4.2: **If** $g_{11,4} \leq 1$ **Then** $L_5 \in Cl1^{\leq}$ | 45.45% |

Experiments are based on two versions of the approach DRSA:

- The pessimistic / optimistic approach: the decision rules inferred by applying the DOMLEM algorithm consider the "upward" as well as the "downward" unions of the decision classes. Thus, a decision rule can classify an object in different unions such as the "at-most Cl1", the "at-most Cl2", the "at-least Cl2" and the "at-least Cl3". An object classified in the union "at-most Cl2" can belong either to the class Cl1 or to the class Cl2. Similarly, an object classified in the union "at-least Cl2" may belong either to the class Cl2 or to the class Cl3. However, an object has to be classified in one and only one decision class. In this case, the pessimistic approach assigns the objects of the union "at-most Cl2" to the class Cl1 and those of the union "at-least Cl2" to the class Cl2. Likewise, the optimistic approach assigns them respectively to the decision classes Cl2 and Cl3.

- The incremental/ non-incremental approach: the training phase of the model can be non-incremental, so based only on the training sample ("Learners of reference") defined in the current week of the MOOC so as to make predictions concerning the following week during this same MOOC (so $K_i = K'_i$). Likewise, this sample can be an incremental one if we add to the learning sample of the current week samples of all the previous weeks (so $K_i = K'_i + K_{i-1}$).

Based on these approaches, four cases were considered in order to test the proposed method: the non-incremental pessimistic approach, the incremental pessimistic approach, the non-incremental optimistic approach and the incremental optimistic approach. The prediction model efficiency is assessed using the F-measure evaluation throughout the five weeks of the MOOC broadcast. The results are summarized in Figure 2.

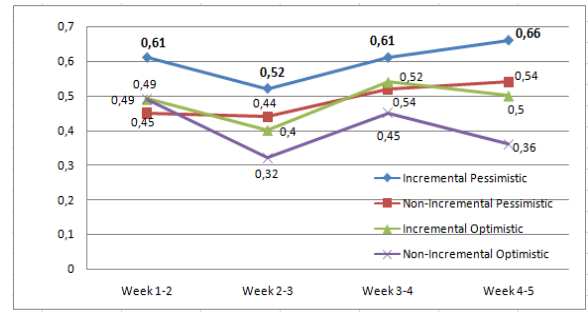

Figure 2: Comparison of the global F-measure calculated for the four approaches during the MOOC broadcast

We note that the pessimistic incremental approach gives the highest prediction efficiency. In effect, as we mentioned, the DSRA approach focuses rather on the quality of the training sample and not on its size. Thus, to meet the cognitive capacity of the decision maker, we generally select training samples of a limited size. However, the sample size is also important to diversify the examples to consider when predicting. Thus, the incremental approach has addressed this limitation relative to the learning sample size and therefore led to better results compared to the non-incremental one. This method has also permit to overcome the imbalanced data. Therefore, if we consider certain external factors of the MOOCs dropout such as the lack of time and the absence of a concrete commitment, the "struggling learners" are more likely to abandon the MOOC than to be "Leader learners". Hence, we can say that the pessimistic approach is more efficient compared to the optimistic one.

In Figure 3 we focus on the pessimistic incremental approach and we present the precision, the recall and the F-measure measures relative to each decision class Cl1, Cl2 and Cl3 for each week during the broadcast of the concerned MOOC.

- The F-measure corresponding to the decision class Cl1 of the "At-risk learners" increases over time. Thus, the efficiency of the class Cl1 prediction increases from a week to another. In effect, a MOOC is known by the presence of what we call "lurkers". These are the participants who register just to discover the MOOC concept and who leave it at the first evaluation. And even though they remain active during the first week of the MOOC, they have a prior intention to abandon it. This type of learners degrades the quality of the prediction model which is based on the profile and the behaviour of the learner and not on his intention. Consequently, the more the number of lurkers decreases, the higher the prediction quality is improved.

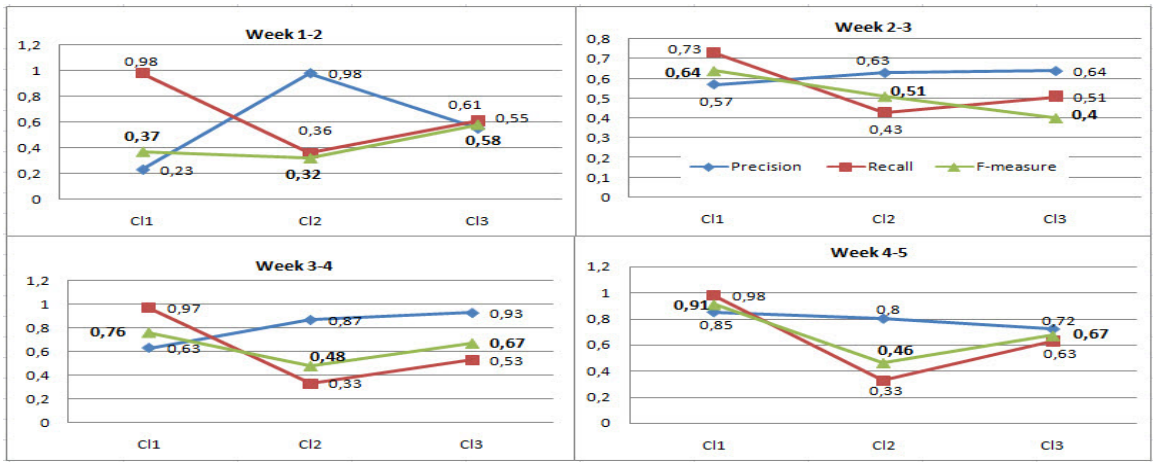

Figure 3: The efficiency measures of the prediction model for each decision class over weeks

- The F-measure relative to the decision class Cl3 of the "Leader Learners" increases over time too. Indeed, from one week to another, these learners increase their participation in the forum, a thing which gives us more information concerning their profile. In addition, the assessment activities provided by the MOOC are increasingly complex over the weeks. Obviously, if compared to a simple Quiz , a complex assessment such as the peer-to-peer activity permits more to better assess a learner. This is justified by the deterioration of the F-measure of the class Cl3 in the week 2-3. In fact, the MOOC on which we tested this method, proposed a quiz at the end of week 2 and a peer to peer activity at the end of week 3. However, students who pass a quiz may hang at a peer-to-peer activity or even abandon the MOOC because of its complexity. This may also affect the quality of the prediction.

In summary, on the basis of these results, we confirm that the DRSA method has achieved very satisfactory results (overall F-measure reached 0.67 for the week 4-5 and 0.61 for the week 1-2). Thanks to the intervention of the pedagogical team, this method succeeds to overcome the problem of imbalanced data because the training sample is based on the quality of the objects it contains and not on their number. The incremental approach also allowed to have a richer and a more diverse sample. Finally, the pessimistic approach permitted to take into account the external factors of a MOOC that can by a way or another incite active learners to abandon the training.

## 6 Conclusion

In this paper, we have proposed a method based on the Dominance- based Rough Set Approach for a periodic incremental prediction. This method permits to overcome the problem of learning in imbalance data. It consists of two phases; first, the construction of a preference model resulting in a set of decision rules; and second, the classification of the "potential objects". The second phase runs periodically at the beginning of each period, on the basis of data provided when implementing the first phase at the end of the previous period. The purpose is to predict the decision class to which each object could belong. This method has been validated on real data coming from a French MOOC proposed by a Business School in France and broadcasted on a French platform. This method has two objectives: (i) minimizing the dropout rate through the early identification of the "At-risk Learners", and so helping them carrying on the MOOC; and (ii) improving the individual appropriation of the exchanged knowledge by the identification of the "Leader Learners" who will be mobilized to support the other learners throughout their training. Experiments showed that the pessimistic incremental approach gives the most efficient preference model with an F-measure that reaches 0.67 and so overcomes the problem of learning in presence of imbalance data. In this work, the prediction concerns only the following week during the MOOC. However, it is more interesting to capitalize in one MOOC in order to predict the exact week during which the learner will be at-risk of dropping similar MOOCs.

# On Solving the Class Imbalance Problem for Clinical Decision Improvement Using Heart Sound Signals

**Arijit Ukil, Soma Bandyopadhyay, Chetanya Puri, Rituraj Singh, Arpan Pal**
TCS Research and Innovation
Kolkata, India
{arijit.ukil, soma.bandyopadhyay, chetanya.puri, singh.rituraj, arpan.pal} @tcs.com

## Abstract

Clinical decision making in data-driven computational method is a challenging task due to the scarcity of negative examples. In clinical analytics, the presence of corruption in terms of motion artifacts, transient noise, instrument defects obscure the inferencing outcome. However, well-represented noisy training datasets are rarely available for model generation although identification and removal of noisy test signals positively impacts the accuracy of clinical decision making. In this paper, we investigate on heart sound recording or PCG (Phonocardiogram) signal to detect cardiac abnormality condition under practical consideration of contaminated inputs. Our results demonstrate that identification and removal of noisy PCG signals significantly improve the clinical decision making outcome. Our solution consists of the following novel components: 1. Rare-class generation method that augments the rarely available negative examples, 2. Feature space optimization on the conditional likelihood maximization over different selection criteria that maximizes both consistency and supremacy over relevancy of the feature set, 3. Jointly optimizing the hyper-parameters; the kernel co-efficient $\gamma$ and rejection rate hyper-parameter $\nu \in (0,1)$ of one-class support vector machine (OC-SVM) to stabilize the decision boundary while maximizing the classification performance, a sub-class of kernel learning problem. Our constructions of rare-class generation, feature space optimization and OC-SVM optimization are completely generic. We have rigorously experimented on publicly available annotated MIT-Physionet database to substatiate our claim of improved clinical decision inference: we have achieved 25% improvement on predicting clinical decision making for cardiac abnormality detection and attained absolute accuracy is 82%.

## 1 Introduction

Clinical inferencing through computational methods is an important step towards automation of basic health screening and remote health monitoring systems. It is an established fact that cardiac disease-related deaths are the the biggest killer worldwide. However, the cardio-vascular diseases are preventable. Proactive or early detection of onset of the disease would commendably deter premature human life loss. In this paper, we show that PCG, a fundamental cardiac condition marker has the potential to accurately detect abnormal cardio-vascular condition. Cardiac auscultation is a fundamental cardiac condition analysis that is performed by doctors through listening the heart sound signal using stethoscope. The change of acceleration of blood flow due to the opening and closure of the heart valves causes vibration at the heart chambers and valves, which results in the generation of heart sound signal or PCG. In the event of cardiac abnormality conditions like coronary artery disease, cardiac arrhythmia, the vibration of heart valves respond differently, as a result the signature of PCG signal changes.

PCG or heart sound signal are audio signal which can be collected cheaply through werable sensors, even by using smartphone microphone. The extracted PCG signal can be analyzed locally through computational devices like smartphones to detect the underlying cardiac abnormality. However, corruption from wide range of sources like transient and ambient noise, motion artifacts extensively distort physiological signals like PCG, PPG (Photoplethysmogram) [Bandyopadhyay et al., 2016a; Bandyopadhyay et al., 2016b]. Such deterioration of signal quality and intelligence impacts the computational analysis and seems to be the vital reason for low accuracy of present state-of-the-art automated PCG based cardiac event detection algorithms. Researchers often attempt to classify PCG signal to indicate cardiac abnormality ignoring the impact of noise [Phy, 2016; Puri and Mukherjee, 2017; Ukil et al., 2016b]. One of the major reasons is the less availability of noisy signal as training data. It still remains widely open area specifically on reducing mis-classification rate of the inductive process in presence of unknown positive-negative distribution at the testing dataset. In the absence of negative examples, one class classification is the obvious choice. A one-class-classifier attempts to find a distinctly sepearted boundary between the class with the given examples (training data) with the rest (1 with N) in the entire feature space. It is an effective method to tackle the class imbalance problem [Manukyan and Ceyhan, 2016].

Researchers have independently worked on removing specific kind of noise or artifact for specific physiological signals [Fraser et al., 2014; Sardouie et al., 2015; Orphanidou et al., 2015]. Few methods that present one-class classification to identify noisy signals, eventually consider true or major examples of noisy training data throughout the method [Fasshauer and Zhang, 2007]. Such approaches fail to justify the consideration of one-class classification and practical implication. In fact, such problems could be solved more emphatically by powerful binary classifiers.

Another major challenge is to properly parameterize the one-class learner kernel. In this work, we conisider one-class SVM (OC-SVM) as the one-class learner. However, the optimality of the parameters; kernel co-efficient $\gamma$ and rejection rate hyper-parameter $\nu$ of the one-class SVM (OC-SVM) play a determinstic role on the performance of the learner model. These parameters ($\gamma$ and $\nu$) are responsible for OC-SVM to form non-linear boundary with the training vectors (positive examples). The variance of the RBF (Radial Basis Function) kernel function is determined by $\gamma$. For larger $\gamma$, the narrower would be the kernel and the corresponding hyper-surface is spiky, which means it is zero in almost everywhere except at the support vectors, while low value of $\gamma$ corresponds to larger RBF bandwidth and the hyper-surface is very flat, whereas $\nu$ represents a lower bound on the support vector and higher bound on number of outliers. We can control $\nu$ to find optimally-fitted smooth decision boundary.

In literature, optimization of $\gamma$ and $\nu$ is solved in isolation and in most cases, cross-validation based solution of optimal $\gamma$ [Fasshauer and Zhang, 2007] and Parzen window estimation of $\nu$ [Arcolano and Rudoy, 2008] are employed. However $\gamma$ and $\nu$ are not independent, isolated attempts of optimizing $\gamma$ and $\nu$ would invariably result in sub-optimal performance. In fact, OC-SVM has high sensitivity with both $\gamma$ and $\nu$.

In this paper, we provide a practical solution to address all the above mentioned research challenges. We propose 1. Negative or rare-class generation method that augments the rarely available negative examples, 2. Effective, discriminative, optimal feature selection, 3. Joint optimization of kernel co-efficient $\gamma$ and rejection hyper-parameter $\nu$ to construct OC-SVM that can accurately form decision boundary among clean and corrupted physiological signals. The proposed decorrupting method is sufficiently generic for its applicability to other physiological signals. Our process of decorruption of physiological signals has two components: 1. Noisy signal identification, 2. Removal of the identified noisy signals in the consequent processing chain. We also demonstrate through available MIT-Physionet annotated PCG database that appropriate and accurate decorruption of physiological signals result in much improved clinical analytics. From the best of our knowledge, this is the first attempt that proposes clinical analytics automation with noisyphysiological signal identification and removal in complete absence of human-in-loop. In order to ensure automated reliable healthcare analytics, clinical significance of the implication needs to be robust and solution has to be developed under practical constraints. In that regard, our proposed solution encompasses augmentation of both technical implications and clinical inferences.

The paper is organized as follows. In Section 2, we describe the problem statement. Typical architecture of the sytem for practical deployment is presented in Section 3. We discuss our major contributions- Rare-class generation method in Section 4, and describe OC-SVM kernel learning with optimal hyper paremeter selection and optimal feature space selection in Section 5. In Section 6, we present our proposed bi-stage learner method for PCG signal analytics. Although, we currently investigate on PCG signals, the methods and algorithms are presented in a generic framework. We present the result of our proposed method in Section 7. Finally, we conclude in Section 8.

## 3 Problem Statement

Let, $\mathcal{X} = (\mathbb{X}_+, \mathbb{X}_-)$, where $\mathbb{X}_+$ be the available known-labeled positive training set and $\mathbb{X}_-$ be the known-labeled rare training set, $\mathbb{X}_+ = \{\mathrm{x}_i^+\}_{i=1}^{\Pi}$, $\mathbb{X}_- = \{\mathrm{x}_i^-\}_{i=1}^{\pi}$, where $\Pi \gg \pi$; $\mathrm{x}_i^+, \mathrm{x}_i^- \in \mathbb{R}^d$. Let, there exists an oracle $\mathcal{O}$ which generates rare/negative training examples $\mathbb{X}_{--} = \{\mathrm{x}_i^{--}\}_{i=1}^{\Pi}$ from $= (\mathbb{X}_+, \mathbb{X}_-)$ by a generator function $\mathcal{G}$:
$$\mathcal{X} = (\mathbb{X}_+, \mathbb{X}_-) \xrightarrow{\mathcal{G}} \mathbb{X}_-$$

Consequently, Let, $\mathcal{F}$ be the feature space and $f(x)$ be the function to construct the decision boundary. So, the primary objective is to find $f^*$ (optimal decision boundary) that minimizes the expected risk of prediction, which is dependent on the hyper-parameters ($\hbar$) and the feature space $\mathcal{F}$:
$$f \xrightarrow{\hbar_{opt}, \mathcal{F}_{opt}(\mathbb{X}_+, \mathbb{X}_{--})} f^*$$
Where, $\hbar_{opt} = \{\nu_{opt}, \gamma_{opt}\}, \mathcal{F}_{opt}$ are the optimal hyper-parameters and selected optimal feature set and the optimization of hyper-parameters and feature set is performed over $\{\mathbb{X}_+, \mathbb{X}_{--}\}$.

We construct kernel-learned OC-SVM with $\hbar_{opt} = \{\nu_{opt}, \gamma_{opt}\}$, which are the optimum values of rejection rate hyper-parameter $\nu \in (0,1)$ and kernel co-efficient $\gamma$ respectively and $f^*$ is to be feature and hyper-parameter optimized such that smooth, optimal-fitted decision boundary can be constructed.

The basic problem is to identify the presence of cardiac abnormality from test PCG signals $\mathcal{T} = \{(t_i, y_i)\}_{i=1}^N, y_i \in \mathcal{Y}_T = \{+1, -1\}, +1 \rightarrow Normal; -1 \rightarrow Abnormal$. Let, training data $\mathcal{X} = \{(x_i, y_i)\}_{i=1}^M$, where $x_i \in \mathbb{R}^d$ is the training instances, $\mathcal{X} = \{(x_i, n_i)\}_{i=1}^M$, $n_i \in \mathfrak{N} = \{clean, noisy\ (rare)\}$. In this work, we declare an uncertain class $\mathcal{U}$ from the test data, signifying that $\mathcal{U}$ is the class of noisy PCG signals. In $\mathcal{U}$ class, we pressume that the physical interpretation of cardiac abnormality detection would be highly error-prone due to severe loss of information. In a nutshell, the problem is:

$$f \xrightarrow{\hbar_{opt}, \mathcal{F}_{opt}(\mathbb{X}_+, \mathbb{X}_{--})} f^* \xrightarrow{\mathcal{T} = \{(t_i)\}_{i=1}^N} \{+1, -1, \mathcal{U}\}.$$

## 3 Typical Architecture

We depict a typical architecture in figure 1 to realize the practical development of an automated (cardiac) healthcare analytics solution. Our main contribution is to identify and discard the noisy PCG signal through rare-class generation by joint parameter optimization in OC-SVM and feature space optimization. Next, we analyze the clean PCG signal to infer whether the signal belongs to pathological normal and abnormal condition. For implementation perspective, the complete analytics system can be part of edge-analytics and the operation can be performed locally, without any requirement of separate infrastructure like server or cloud or Internet.

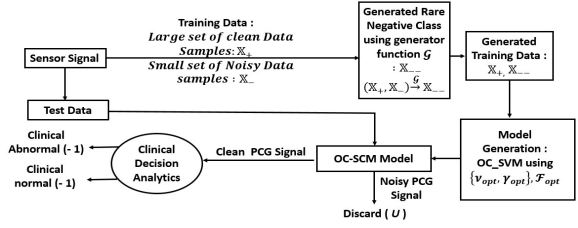

Figure 1 Typical architecture of PCG-based clinical analytics by noise removal.

## 4 Rare-class Generation

The critical problem in the anomaly detection is the complete absence of negative examples or anomaly-labeled training set. In some situations, few negative examples are present. Let $\Pi, \pi$ be the number of positive and negative training examples respectively and $\Pi \gg \pi$.

Let $\pi$ number of negative examples is available. Different investigations show that in many real-life examples such $\pi$ number of negative examples can be collected [Zhou and Li, 2010] through inexpensive ways. In many situations, 'crowd-sourcing' of labeling is performed [Settles, 2011] and nonexpert-labeled examples are generated.

In order to generate reliable validation set in oversampled space that can resemble practical scenarios in absence of negative examples or inadequate negative examples, we require to synthesize noisy datasets such that $\mathcal{X}_{noise} \sim \mathcal{P}_x^{synthetic-noise}$. The familiar synthesis method, SMOTE [Chawla, et al. 2002] describes a mere replication of the rare class examples in feature space through simple oversampling, which is unlikely to form $\mathcal{X}_{rare-class/noise} \sim \mathcal{P}_x^{synthetic-noise}$.

In physiological signal space, typically a noisy signal consists of four segments: clean segment, motion artifact, random noise and power line interference segment, which correspond to measurement, instrumentation and interference plane respectively. We propose markov model based synthetic signal generation. Markov model provides a systematic and stochastic method to model the time varying signals. In markov model the future state only depends upon the current states, not on the predecessor states. This assumption and property makes markov model as the best suited model for the generation of noisy synthetic data.

A stochastic process $\{X_n\}$ is called a markov chain if for all times $n \geq 0$ and all states $i_0, i_1, \ldots j \in S$

$$P(x_{n+1} = j | x_n = i, x_{n-1} = i_{n-1}, \ldots \ldots, x_0 = i_0)$$
$$= P(x_{n+1=j} | x_n = i) = P_{ij}$$

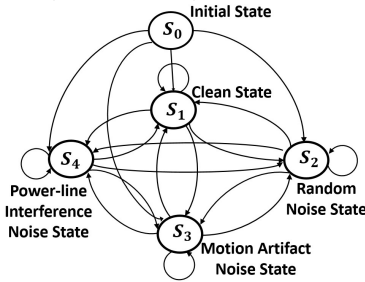

Figure 2 Markov chain model to generate the synthetic noisy signal.

Here, $P_{ij}$ denotes the probability to move from one state to another subject to $\sum P_{ij} = 1$ and is known as one state markov chain. The objective of the model is to generate noisy signals that closely resembles to actual noisy signal present in the real world. The signal generated from the model should be in such a way that percentage of the noisy segments should be larger as compared to the clean segments. In order to generate the large percentage of noisy segments, we need to give higher transition probabilities weights to the noisy states in the model. It is to note that In Table I, transition probabilities moving to noisy states $(S_2, S_3, S_4)$ is higher as compared to the clean state $(S_0)$. Considering this assumption in this case study we have considered the state transition table depicted in table 1. Generalizing to the context, state transition probabilities can be accordingly changed as per the percentage of noisy portions in the signal. Figure 2 represents the markov chain to generate the noisy time series data. The noise model consists of five states (figure 2). The state transitions matrix is formed from noisy signal instances which are extracted from publicly available sources [Sardouie et al., 2015; Fasshauer and Zhang, 2007].

*Table 1 State Transition Probabilities* $(P_{ij})$

|  | $S_0$ | $S_1$ | $S_2$ | $S_3$ | $S_4$ |
|---|---|---|---|---|---|
| $S_1$ | 0.05 | 0.05 | 0.05 | 0.05 | 0.1 |
| $S_2$ | 0.4 | 0.5 | 0.3 | 0.4 | 0.3 |
| $S_3$ | 0.5 | 0.4 | 0.6 | 0.5 | 0.5 |
| $S_4$ | 0.05 | 0.05 | 0.05 | 0.05 | 0.1 |

**Algorithm1** Rare Class (Noisy Data) Generation
**input** : 1. Rare Negative class examples $\mathbb{X}_- = \{x_i^-\}_{i=1}^{\pi}$
     2. Positive class examples $\mathbb{X}_+ = \{x_i^+\}_{i=1}^{\Pi}; \Pi >> \pi$
     3. Prior Knowledge: {Refer Table 1, Figure 2}
       (a) Different transition state : $S = \{S_i\}_{i=1}^n$
       (b) State Transition probabities $P_{ij}; \sum P_{ij} = 1$
**Output** : Generated rare/negative class examples-
     $\mathbb{X}_{--} = \{x_i^{--}\}_{i=1}^{\Pi}$
**algorithm**
1. Let length of the positive class examples be
$L_p = \{l_1^p, l_2^p, \ldots \ldots l_{\Pi}^p\}$
2. Construct markov chain model {Refer Figure 2}
3. $S_0, S_1$: Randomly selected from $\mathbb{X}_+$
  $S_2$ : Additive white noisy signal from gaussian distribution
  $S_3$ : Typical motion artifact samples [Fraser et al., 2014; Fasshauer and Zhang, 2007].
  $S_4$ : Powerline Interference (60 Hz).
4. Construct $L_r = \{l_1^r, l_2^r, \ldots \ldots, l_{\Pi}^r\}$, where $l_i^r, i = 1, 2, \ldots, \Pi$ be the length the generated rare/negative class examples such that $\sum L_p = \sum L_r$.
5. $\mathbb{X}_i^{--} = \{S_0, S_k\}_i$, order in which $S_k; k \in [1, 4]$ selected, based on State Transition probabilities {Refer Table 1} ; E.g. $x_i^{--} = \{S_0, S_2, S_3, S_1, S_4\}_i$ where length $(x_i^{--}) = l_i^r, i = 1, 2, \ldots, \Pi$.

Our intuition behind the the computation of state transition probabilities is that state# 2 ($S_2$ =random noisy segment) and state# 3 ($S_3$ = motion artifact) would contribute maximum in noisy signals. Hence, the corresponding probabilities are taken higher than other states.

In figure 3, we show that our noise synthesis model has the ability to closely capture the real-life noisy physiological signal, which confirms the assumption $\mathcal{X}_{real-life\ noise} \sim \mathcal{P}_x^{synthetic-noise}$. It is observed that morphological identities are closely preserved in the generated synthetic noisy signals and the real-life noisy signals.

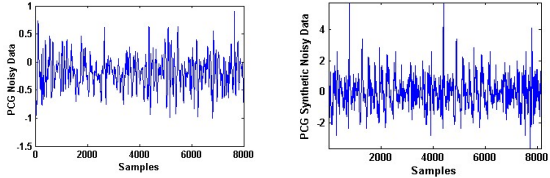

Figure 3 Physiological real-life noisy data vis-à-vis synthetically generated rare-class (noisy PCG signal).

## 5 OC-SVM Kernel Learning with Optimal Hyper Parameter Selection

Traditional classification problem discriminates binary or M-ary classes. In reality, there exists many important set of problems where the discrimination is between one class from any other class(es). Particularly, when only positive example set is available for any reliable training. Decorruption of physiological signals like PCG is one of such application cases. For ensuring clinical inference from such signals without human intervention, appropriate decorruption is of utmost important. It is to be noted that corrupted PCG signal has much less information for appropriate clinical analytics. In fact, such contaminated signals would render high misclassification rate which is completely undesirable in clinical inferencing purposes. Noise or corruption as part of the signal space as well as class noise leads to poor classification outcome independnet on the power of the machine learning technqiues used [Gui, et. al, 2015].

In order to identify noisy PCG signals, the learner model has to be adequately trained by both the clean and noisy examples. However, training set mainly contains clean signals only and it is impractical to train with corrupted signals, as corrupted signal universe can rarely be captured through a smaller set of negative examples. In such scenario, optimal decision boundary construction plays major role to reduce the misclassification error. In that regard, we kernel-optimize the classifier OC-SVM by finding optimal kernel co-efficient $\gamma_{opt}$ and rejection hyper-parameter $\nu_{opt}$. First, we briefly describe OC-SVM and further, we describe our proposed method for hyper-parameter and feature optimization.

### 5.1 OC-SVM: Brief Descrtiption

OC-SVM learns the decision boundary for achieving maximum separation between the points and the origin [Scholkopf et al., 2001, Cohen et al., 2004]. It is more specifically $[1 + N^+]$ class learning problem. There are $N^+$ number of unknown classes available in universe, but we attempt to identify the $1 -$class.

Let, training set $\mathcal{X} = \{x_i\}, i = 1,2, \ldots n$. $x_i \in \mathbb{R}^d$ and $\phi(x_i) = [\mathcal{F}_1(x_i), \mathcal{F}_2(x_i), \ldots, ]^T$ contain the features $\mathcal{F}_i(x_i)$ from $\mathcal{F}$, Where a non-linear function $\phi: \mathcal{X} \to \mathcal{F}$ maps vector $\mathcal{X}$ from input vector space $\mathcal{X}$ to feature space $\mathcal{F}$.

The constraint objective function of OC-SVM is:
$$\min_{\omega,\rho,\xi} \left( \frac{\|\omega\|^2}{2} + \frac{1}{\nu n} \sum_{i=1}^n \xi_i - \rho \right), \quad i = 1,2,\ldots,n \quad (1)$$
subject to:
$$\omega^T \phi(x_i) \geq \rho - \xi_i, \quad \xi_i \geq 0 \quad (2)$$

Where $n$ denotes the number of training vectors, $\xi_i$ is the $i^{th}$ positive slack variable that penalizes the objective function, $\omega$ is the vector perpendicular to the decision boundary and $\rho$ is the bias or offset.

For $(\alpha_1, \ldots \alpha_n)$ non-negative Lagrange multipliers to solve equation (2), the solution is equivalent to its Lagrange duality form and its Wolfe dual representation form [Theodoridis and Koutroumbas, 2001]:
$$minimize \left\{ \frac{1}{2} \sum_{i,j} \alpha_i \alpha_j \langle \phi(x_i), \phi(x_j) \rangle \right\} \quad (3)$$
subject to:
$$0 \leq \alpha_i \leq \frac{1}{\nu n}, \quad \sum_{i=1}^n \alpha_i = 1$$

From equation (3), we find the decision function:
$$f(x) = sgn(\sum_{i=1}^n \alpha_i \langle \phi(x_i), \phi(x) \rangle - \rho) \quad (4)$$
Using the method of "kernel trick" of kernel $\mathcal{K}$ that satisfies Mercer's conditions [Vapnik and Vapnik, 1998]:
$$\mathcal{K}(x_i, x_j) = \langle \phi(x_i), \phi(x_j) \rangle \quad (5)$$
$\mathcal{K}$ is a dot product to fit into the transformed feature space for providing maximum-margin hyper-plane $\mathcal{H}$.
From (4), $f(x) = sgn(\sum_{i=1}^n \alpha_i \mathcal{K}(x_i, x) - \rho)$

Many kernel functions are presented in the literature like polynomial $(\mathcal{K}(x_i, x_j) = (x_i \cdot x_j)^d)$, hyperbolic $(\mathcal{K}(x_i, x_j) = \tan^{-1}(a. x_i. x_j + c), a > 0, c < 0)$, radial basis function (RBF) $(\mathcal{K}(x_i, x_j) = e^{(-\gamma \|x_i - x_j\|^2)}, \gamma > 0)$. We consider RBF kernel for its generalizability characteristics.

### 5.2 Kernel Learning with Hyper-parameter Optimization for OC-SVM

From equation (3) and (5), we notice the optimization function for RBF kernel:
$$minimize \left\{ \frac{1}{2} \sum_{i,j} \alpha_i \alpha_j e^{(-\gamma \|x_i - x_j\|^2)} \right\} \quad (6)$$
Subject to:
$$0 \leq \alpha_i \leq \frac{1}{\nu n}, \quad \sum_{i=1}^n \alpha_i = 1$$

From equation (6), we observe that $\gamma, \nu$ are not independent. Our approach is to jointly optimize $\gamma, \nu$ with the criteria that class separability in terms of performance function maximization while ensuring stability.

Let, Learning algorithm $\mathcal{A}$ (in this case OC-SVM) finds a function $f$ to minimize expected loss $\mathcal{L}(\mathcal{X}, f)$ from the labeled training set distribution $\mathcal{P}_\mathcal{X}$. The objective of optimal parameterization is to find the optimal parameter set $\hbar_{opt} = \{\gamma_{opt}, \nu_{opt}\})$ that minimizes the generalization loss [Peng et al., 2005]:
$$\mathbb{E}_{\mathcal{X} \sim \mathcal{P}_\mathcal{X}} \left[ \mathcal{L}(\mathcal{X}, \mathcal{A}_\hbar(\mathcal{X}_{train})) \right] \quad (7)$$

It is argued in [Bergstra and Bengio, 2012], that efficient algorithm to implement equation (7) is not available and expectation over unknown distribution $\mathcal{P}_\mathcal{X}$ is impossible to find. Cross-validation technique is employed in [Bergstra and Bengio, 2012] with the assumption that $\mathcal{X}_{train}$ elements are drawn i.i.d $\mathcal{X} \sim \mathcal{P}_\mathcal{X}$:

$$\hbar_{opt}^{cv} = \underset{p \in \Lambda}{\operatorname{argmin}} \left[ mean \left( \mathcal{L}\left(\mathcal{X}, \mathcal{A}_p(\mathcal{X}_{train})\right) \right) \right] \quad (7.1)$$

Let $K$-fold cross-validation be performed with trail search $\mathcal{S} = \sum_{k=1}^{K} |\mathcal{L}^k|$. Such search suffers from curse of dimensionality [Bergstra and Bengio, 2012]. Another practical drawback is that such search $\mathcal{S}$ attempts to find the optimal hyper-parameter $\hbar_{opt}$, which is unique and exactly satisfies equation (7.1). Such search would invariably over-fit the training dataset for unbalanced training vectors.

Let $\mathcal{X} = \{\mathcal{X}_{train}, \mathcal{X}_{valid}\}$ be divided into disjoint sets $\mathcal{X}_{train}, \mathcal{X}_{valid}$ for training and validation respectively, the hyper-parameter optimality condition would be [Thornton et al., 2013]:

$$\hbar_{opt} = \underset{p \in \Lambda}{\operatorname{argmin}} \, \mathbb{E}_{\mathcal{X} \sim \mathcal{P}_\mathcal{X}} \left[ \mathcal{L}\left(\mathcal{X}, \mathcal{A}_p(\mathcal{X}_{train}\, \mathcal{X}_{valid})\right) \right] \quad (8)$$

Our approach of finding $\hbar_{opt}$ is to:
- Jointly optimize $\hbar_{opt}$ such that to find the pair $\{\gamma_{opt}, \nu_{opt}\}$.
- Find a close approximation of $\mathcal{P}_x^{synthetic-n\ /\ rareclass}$ such that: $\mathcal{X}_{noise} \sim \mathcal{P}_x^{synthetic-nois\ /\ rareclass}$. From train and validation set we get $\mathcal{X}_{clean}$ and the training, validation set is now: $\mathcal{X}_{train} = \mathcal{X}_{train}^{Clean}$, $\mathcal{X}_{valid} = \{\mathcal{X}_{valid}^{clean}, \mathcal{X}_{valid}^{synthetic-noi\ /\ rareclass}\}$.
- $\hbar_{opt} = \mathcal{R}\left[\mathcal{S}\left(\mathcal{X}, \mathcal{A}_p(\mathcal{X}_{train}\, \mathcal{X}_{valid})\right)\right]$
- This means that we have to find $\hbar_{opt}$ that satisfies the performance criteria $\mathcal{R}$ and stability criteria $\mathcal{S}$, where, $\mathcal{X}_{train} = \mathcal{X}_{train}^{Clean}$. $\mathcal{X}_{valid} = \{\mathcal{X}_{valid}^{clean}, \mathcal{X}_{valid}^{synthetic-no\ /\ rareclass}\}$.

We define two criteria:
1. Performance criteria.
2. Stability criteria.

- **Performance criteria:**

Let $\Delta_\nu, \Delta_\gamma$ be the quantization steps for $\nu, \gamma$ respectively. $N = \{0, \Delta_\nu, 2\Delta_\nu, \dots, 1\}$, $\Gamma = \{\gamma_{min}, \gamma_{min} + \Delta_\gamma, \gamma_{min} + 2\Delta_\gamma, \dots, \gamma_{max}\}$, which forms $N \times \Gamma$ matrix of hyper-parameter $\hbar = \{\hbar_1, \hbar_2, \dots, \hbar_{N\Gamma}\}$, where $\Delta_\nu, \Delta_\gamma$ are such that $\left\lfloor \frac{1}{\Delta_\nu} \right\rfloor \geq N_{thr}, \left\lfloor \frac{1}{\Delta_\gamma} \right\rfloor \geq \Gamma_{thr}$ and typically, $N_{thr}, \Gamma_{thr}$ are to be taken $> 2^7$.

We define performance factor $\varpi$ as a geometric mean of sensitivity $(s_n)$ and specificity $(s_p)$, $\varpi = (s_n \cdot s_p)^{0.5}$. It is to be noted that $\varpi$ provides information about the trade-off between the sensitivity and specificity, such that both of them to be precisely high for $\varpi$ to be considerable. For each of the instances of N and $\Gamma$ of $\hbar$, we get $\varpi$, when training is on clean data $\mathcal{X}_{train} = \mathcal{X}_{train}^{Clean} = \mathbb{X}_+^{train}$ and validation is on clean and synthetically generated noisy signal set, $\mathcal{X}_{valid} = \{\mathcal{X}_{valid}^{clean}, \mathcal{X}_{valid}^{synthetic-n\ /\ rareclass}\} = \{\mathbb{X}_+^{valid}, \mathbb{X}_{--}\}$. If $|N|, |\Gamma|$ be the cardinality of N, $\Gamma$; we get $\varpi = \{\varpi_1, \varpi_2, \dots, \varpi_{|N||\Gamma|}\}$ which forms $|N| \times |\Gamma|$ matrix.

Performance criteria is stated as:
$$\underset{p \in \Lambda}{\operatorname{argmax}} \mathcal{R}(\hbar, \mathbb{L})_\varpi, \quad \mathcal{R} = r_{\{\hbar_1, \hbar_2, \dots, \hbar_{N\Gamma}\}}(\mathbb{L}) \quad (9)$$

Where, $r_{\{\hbar_1, \hbar_2, \dots, \hbar_{N\Gamma}\}}(\mathbb{L})$ represents the performance among $\{\hbar_1, \hbar_2, \dots, \hbar_{|N||\Gamma|}\}$ on $\varpi = \{\varpi_1, \varpi_2, \dots, \varpi_{|N||\Gamma|}\}$ and $\mathbb{L}$ is the class label of clean and synthetic noisy in validation set.

- **Stability criteria:**

$$\underset{p \in \Lambda}{\operatorname{argmax}} \mathcal{S}(\hbar, \mathbb{L})_\varpi, \quad \mathcal{S} = s_{\{\hbar_1, \hbar_2, \dots, \hbar_{N\Gamma}\}}(\mathbb{L}) \quad (10)$$

Where, $s_{\{\hbar_1, \hbar_2, \dots, \hbar_{N\Gamma}\}}(\mathbb{L})$ represents the performance among $\{\hbar_1, \hbar_2, \dots, \hbar_{N\Gamma}\}$ on $\varpi = \{\varpi_1, \varpi_2, \dots, \varpi_{|N||\Gamma|}\}$ and $\mathbb{L}$ is the class label of clean and synthetic noisy in validation set.

We define stability function $\mathcal{S}$ empirically as:

We choose a unit matrix $\mathcal{T}_{mn} = \begin{bmatrix} 1 & \cdots & 1 \\ \vdots & \ddots & \vdots \\ 1 & \cdots & 1 \end{bmatrix}_{mn}$ that depicts the maximum achievable $\varpi = (0,1)$ for each $m \times n$ cross-section, where $m \ll |N|$ and $n \ll |\Gamma|$

We perform closeness between each of the $\Gamma_{mn}$ and $\varpi_{mn}, \forall |N|, |\Gamma|$. We create weighted score matrix $W$ using $\varpi_{|N||\Gamma|}$ and $\mathcal{T}_{mn}$ as sliding weights to find the region having the most consistent and best performance $W$ is a matrix of size $|N| + m - 1$ by $|\Gamma| + n - 1$ given by:

$W_{kl} = \sum_{i=0}^{m-1} \sum_{j=0}^{n-1} GM_{ij} \Gamma_{i-k, j-l} \quad \begin{array}{l} -(m-1) \leq k \leq |N| - 1 \\ -(n-1) \leq l \leq |\Gamma| - 1 \end{array}$

So stability criteria finds one $m \times n$ the region among $\{\hbar_1, \hbar_2, \dots, \hbar_{N\Gamma}\}$, the stability region $s_{mn}$ from $W_{kl}$ that maximizes $\varpi$.

Our main aim is to simltaneously find the maximum performance that provides given stability. Hence, we slect the corresponding $\hbar_{opt}$ that maximizes the performance factor $\varpi$ as per equation (9) while mainitianing equation (10) in the stable region $s_{mn}$,

$$\hbar_{opt} = \underset{p \in \Lambda}{\operatorname{argmin}} \, s_{mn} \quad (11)$$

The intuition behind imposing stability criteria is to ensure that one-off incident of excellent performance that can result in over-fitting can be conveniently ignored. We apply typical grid search mechanism. It is noticed in [Sci,] that hyper-parameter estimation by random search performance is worse than grid search. Figure 4 represents the training example sepration with optimal hyper-parameter optimization and without optimization. The illustration of figure 4 shows that our hyper-parameter optimization and hence, OC-SVM decision boundary construction is unperturbed by the outler training examples.

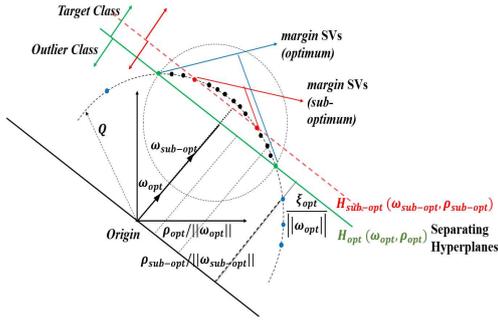

Figure 4 Overview of OC-SVM based hyper-plane sepration optimization in feature space.

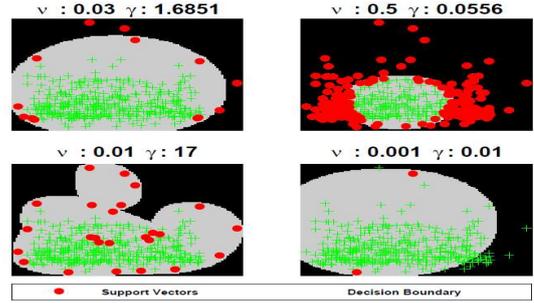

Figure 5 Support vector decision boundaries on different values of ν and γ over training dataset, where $v_{opt}, \gamma_{opt}$ values are 0.03 and 1.6851 respectively.

### 5.3 Feature-space Optimization

The classification process has higher dependency on the selected feature space. The optimal feature space selcetion would be parsimonious in nature whereas it should characterize the training examples as distinctly as possible [Meyer and Bontempi, 2006]. The optimality criteria of a feature selection process is always relative to some objective function and our objective function is to maximize the conditional likelihood of the class labels from the target class given the features [Brown, et al. 2012], where the conditional likelihood function is mutual information $\mathbb{I}(x, y) = \sum_{x \in X} \sum_{y \in Y} p(xy) \log \frac{p(xy)}{p(x)p(y)}$.

We apply Maximum Consistency Maxiamum Dominance (MCMD) based feature optimization method thatoptimizes on the objective function (conditional likelihood maximization) over different selection criteria like mRMR, MIFS, CIFE, ICAP, CONDRED, JMI, CMIM, DISR [Ukil, et al; 2016a], satisfying diverse properties of different selection crietia under common objective function of conditional mutual information optimization.**Maximum Consistency Maximum Dominance (MCMD) property:**

Let $F = \{F_1, F_2, \ldots, F_z\}$ be the features of the given data set $\mathcal{X}$. Feature selection method $\mathbb{M}$ selects $\mathbb{Q}$, $\mathbb{Q} \subseteq \Theta$, where $p$ be the cardinality of $\mathbb{Q}$, $p \leq z$, following the MCMD criteria [Ukil, et al., 2016a]:

*MCMD property [Ukil, et al., 2016a]*:
$\max(\mathbb{U}(\mathbb{Q}, \mathbb{C}|\mathbb{J})) \wedge \max(V(Q, C|J))$
where
$\mathbb{U} = \lambda_{\{\theta_1, \theta_2, \ldots, \theta_p\}}(\mathbb{C})$ and $\mathbb{V} = \beta_{\{\theta_1, \theta_2, \ldots, \theta_p\}}(\mathbb{C})$

where, $\lambda_{\{\theta_1, \theta_2, \ldots, \theta_p\}}(\mathbb{C})$ and $\beta_{\{\theta_1, \theta_2, \ldots, \theta_p\}}(\mathbb{C})$ represents the consistency and dominance between feature subset $Q$ and the target class label $C$, over the conditional likelihood probability $I(x,y)$ respectively for a large set of selection criteria $J$ (mRMR, ICAP, DISR, …).

### 5.4 Smooth Decision Boundary Construction

We validate our claim that support vector decision boundary of $f(x)$, satisfying equation 3,4,5,9,10 11 with our derived optimal hyper-parameter $k_{opt} = \{\gamma_{opt}, v_{opt}\}$ would result in intended smooth, properly fitted decision boundary, shown in figure 5. We have considred MIT-Physionet 2016 PCG database [Phy, 2016].

## 6 Our Clinical Analytics Method: Bi-stage Cardiac Abnormality Detection

Given a PCG time series, a set of 54 features is generated. These features pertain to time domain, frequency domain and wavelet domain, where few of the features are domain-knowledge dependent. Our Abnormality classification algorithm primarily consists of four stages. These 54 features from time, frequency and statistical domain and followed the procedure of cardiac condition detection from PCG signal as proposed in [Puri et al., 2016]. 'Clinical decision analytics' block (refer figure 1) enables the cardiac condition detection or identifying whether the PCG signal indicates cardiac abnormality. In this block, training is done to purely to classify cardiac normal or abnormal irrespective of the other states of the data (clean or noisy). The input to this block is presumably clean as the test or training PCG signals are sanitized by decorruption block.

The main contribution that we bring is to ensure more accuracy in clinical decision making by uncertainty elimination. It is prudent to declare that 'decision cannot be made' than to make wrong decision. When a test PCG signal is declared as 'Uncertain (U) class', the signal extraction process is to be retaken. We summarize the scheme below in figure 6.

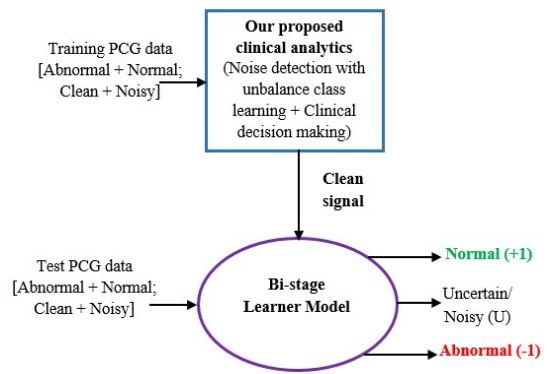

Figure 6 Summarized description of our bi-stage learning method for clinical decision making using PCG signal that eliminates decision making uncertainty by introducing unceratin class.

## 7 Results: Clinical Analytics Improvement

The major impact our research work is on the improvement of clinical analytics outcome, where pathological normal/ abnormal conditions are detected from PCG signals. When the physiological signal is corrupted by different noise sources, our method first detects and discards those noisy signals and subsequently analyze on the clean signal to declare whether the physiological signal corresponds to abnormal condition. We demonstarte the efficacy of our proposed solution in two stages.

### 7.1 Efficacy Towards Noisy PCG signal Detection

We have experimented from publicly available PCG database, where 3153 PCG waveforms are available [Phy, 2016] with annotations pertaining to presence of cardiac abnormality as well as signal quality annotation of these wave forms. We randomly select 1203 clippings as training from the complete 3153 PCG signal database, out of which 960 clean clean signals are used for training [Phy, 2016].

In figure 7 and 8, we show optimum values of hyper parameter and the noisy PCG signal detection accuracy in terms of sensitivity, specificity and $\alpha$-accuracy with default respectively, $\alpha$-Accuracy is defined as $\alpha - Acc = \frac{TP+TN}{TP+FN+TN+\cdot FP}$ where we penalize false positives more.

**Impact of proposed $k_{opt}$ and other $k$ ($k_{default}$)**

A contour plot of the performance on $\nu - \gamma$ plane is depicted in figure 7. The blind or default selection of $k_{default} = \left[\nu = 0.5, \gamma = \frac{1}{F_{selected}}\right]$ [Sci.,]. We observe in figure 8 that both $k_{opt}$ and MCMD feature selection play major role for ensuring better corrupt/ noisy PCG signal detection.

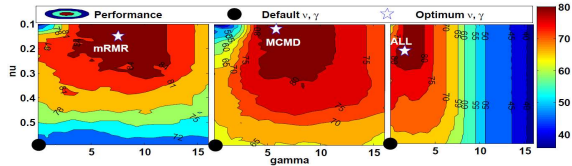

Figure 7 Contour plot depicting a) mRMR, b) MCMD, c) Complete feature selection method over hyper-parameters ($\nu$, $\gamma$). Objective: to reach center of the high performance contour with highest performance.

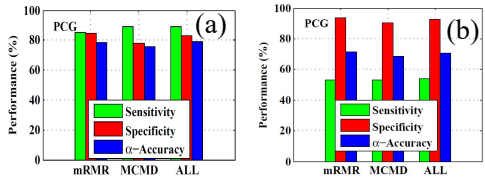

Figure 8 Performance comparison of (a) our scheme with set of different features (mRMR, MCMD, All) with $k_{opt}$ and (b) $k_{default} = [\nu = 0.5, \gamma = 1/|F_{selected}|\ ]$ for noisy PCG signal detection.

### 7.2 Clinical Decision Improvement by Our Proposed method

Out of the complete 3153 datasets in [Phy, 2016], 853 datasets are used for building the normal-abnormal classifier and the remaining 2300 datasets were kept as hidden datasets for testing. It is noticed that decorrupting has the potential to augment the clinical utility of automated cardiac condition screening and for 2300 hidden test datasets, we find our proposed decorrupting approach ensures more than 25% relative performance gain in comparison to the approach in practice given the 'clinical decision analytics' methods are identical as shown below in figure 9.

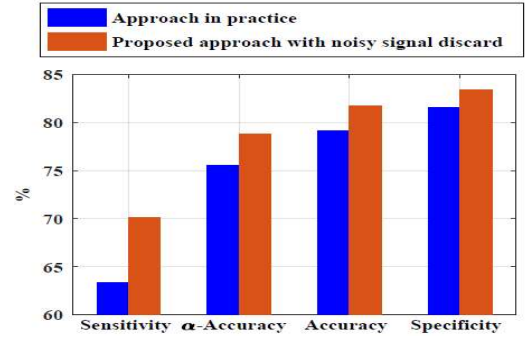

Figure 9 Performance improvement of cardiac abnormality detection with and without noisy signal discard. It shows that our proposed method improves the clinical decision making by incorporating the decorruption component.

## 8 Conclusion

In this paper, we have demonstrated that absence of appropriate data cleaning or decorruption is one of the primary reasons for the failure of automated cardiac management systems using physiological signals, like heart sound or PCG. One of our major contributions is developing highly accurate noisy physiological signal detection method by minimizing the uncertainty in the clinical decision making process. Our rare-class generation method would provide sufficient impetus to elegantly tackle the class imbalance problem of identifying noisy PCG signals. We achieve more than 82% accuracy to detect cardaic abnormality condition from heart sound signals from our extensive experimentaion over publicly available dataasets. Our proposed optimal feature based hyper-parameter optimized one-class classification for decorruption along with pathological condition determination under generic noise synthesis model would potentially usher the realization of reliable computational cardiac health management system.

# An Oversampling Method based on Shapelet Extraction for Imbalanced Time Series Classification

Qiuyan Yan, Fanrong Meng, Qifa Sun

Computer Science and Technology, China University and Mining Technology, Xuzhou, China
yanqy@cumt.edu.cn,mengfr@cumt.edu.cn,sunqifa@live.com

## Abstract

Time series classification (TSC) is a critical problem in data mining communities that becomes more challenging when a data distribution is imbalanced. Owing to high noise and dimensionality, the selection or extraction of representative and distinctive features is highly important for TSC. However, the current widely used sampling-based rebalanced method is applicable only to raw data, not to the proper feature space. To address this problem, this paper proposes a novel oversampling method based on feature extraction for imbalanced TSC. First, we extract the representative pattern, shapelets, as the distinguished characteristic between time series.Also, we replace the classic shapelets quality measure , information gain, with the AUC measure which is more suitable for imbalanced data. Second, to remove redundant shapelets and acquire the most distinctive characteristics of each time series, we construct a diversified graph of a shapelet candidate set and apply diversified top-k queries to the diversified graph. These diversified top-k shapelets are used to construct the new feature space by calculating the distance between each time series and shapelet candidate. Last, the SMOTE oversampling method is applied to the new feature space to generate synthetic data. The competitiveness of the proposed methods is demonstrated through experiments on benchmark data, and the results of the experiments verify that our proposed method outperforms the compared algorithms.

## 1  Introduction

Imbalanced learning aims to tackle the adverse influence on learning algorithms caused by the (relative or absolute) bias of the size of different classes [Gong and Chen, 2016]. An imbalanced dataset has at least one class in which the number of examples is far less than others.The imbalanced TSC problem is more significant for the traditional TSC algorithm. For example, in mine and gas outburst monitoring, an outburst accident is very rare; thus, the amount of that type of accident data is much smaller than the amount of normal data. However, recognizing any ominous accident data accurately can help to prevent accidents from occurring; thus, in these cases, identifying the minority samples is more meaningful than simply obtaining a high classification accuracy.

The approaches of imbalanced learning can broadly be divided into three main streams: algorithm level ,data level and hybrid methods. Algorithm-level methods concentrate on modifying existing algorithms to fit a biased data distribution. Cost-sensitive methods [Cheng *et al.*, 2017] are typical of this category. A large penalty cost will be incurred by the misclassification of a minority class. It is difficult to determine the actual values that should be used in a cost matrix. Data-level methods mainly involve oversampling of the minority classes and undersampling of the majority classes. Resampling methods can be further divided into random sampling and synthetic sampling methods. Overall, the synthetic oversampling methods best consider the original distribution of a minority class. The most widely used oversampling approaches include SMOTE [Chawla *et al.*, 2002], BorderSMOTE [Han *et al.*, 2005], ADASYN [He *et al.*, 2008] and INOS [Cao *et al.*, 2013]. Hybrid methods [Sun *et al.*, 2015] combine both algorithm-level methods and data-level methods to take advantage of the strengths of those methods and minimize their weaknesses. The use of an ensemble of classifiers is effective for improving prediction accuracy in resolving the imbalance problem; however, it is difficult to define and increase the diversity.

Time series classification (TSC) problems are different from traditional classification problems because the attributes of TSC problems are time ordered. Owing to high noise and dimensionality, feature selection is a key procedure in time series data classification. The selection or extraction of representative and distinctive features is even more important than the type of classification algorithm used. Unfortunately, sampling methods have rarely been proposed for time series imbalance classification considering feature selection. As discussed above, the widely used oversampling methods, including SMOTE, BorderSMOTE and ADASYN, are applied directly to raw data. INOS calculates the covariance among different time steps to maintain the original covariance structure. However, INOS requires that the sequences be univariant and of equal length; these requirements might be difficult to achieve in most real-world applications.

Shapelets[Ye and Keogh, 2009] are discriminative patterns in time series that best predict the target variable when their distances to the respective time series are used as features for a classifier. The utility of shapelets has been confirmed and extended by many independent groups of researchers. In this paper, we focus on an oversampling method based on shapelets extraction for imbalanced time series classification. Firstly, we extract the representative pattern, shapelets, for all time series objects.Specifically, to correct the measure of classification quality for imbalanced data, we replace the information gain of classic shapelets measure with the AUC measure, which is widely used as the quality metric for imbalanced data. Then, to select the most representative and distinctive shapelet pattern set, we apply the diversified top-k query method to a shapelet candidate set. Finally, we transform the training data by calculating the distances between subsequences and shapelets , and the distance matrix is considered as the new feature space. The SMOTE oversampling method is applied to the new feature space.

This paper is organized as follows. Section 2 introduces the background. Section 3 details the proposed algorithm. In Section 4, experiments are performed with benchmark data. Section 5 concludes the paper.

## 2 Related methods

In this section, the basic concepts regarding shapelets and the SMOTE sampling method, which are used in our proposed method, are briefly described.

### 2.1 Shapelets

The original shapelet-based classifier embeds the shapelet discovery algorithm in a decision tree method, and information gain is adopted to assess the quality of candidate shapelets [Ye and Keogh, 2009]. The shapelet transformation classification method was introduced to separate the processes of shapelet selection and classification [Lines *et al.*, 2012]. In this category, distances between time series and shapelets can be viewed as new classification predictors. It has been shown by various researchers that shapelet-derived predictors increase the classification accuracy [Mueen *et al.*, 2011]. In addition, shapelets also provide interpretive features that help domain experts to understand the differences between target classes.

Research challenges with respect to shapelet technology include the following: candidate shapelet selection is time consuming, and large quantities of redundant shapelets decrease the accuracy of classification. Some studies have sought to address this issue by introducing clustering [Rakthanmanon and Keogh, 2013], the diversified query method [Yan *et al.*, 2016] or the generalized learning method [Hou *et al.*, 2016] to reduce the redundancy. However, all those studies have focused on balanced time series datasets.

### 2.2 SMOTE oversampling method

The SMOTE algorithm [Han *et al.*, 2005] is mainly used to balance datasets by increasing the number of minority samples in a dataset during processing of an imbalanced dataset. The SMOTE algorithm creates new "synthetic" samples between existing minority samples. It solves the problem of overfitting of the learning model, because the minority samples that are simply copied will be too special and hence cannot be generalized. The SMOTE algorithm is as follows:

(1) For each sample of the minority classes $x \in S_{\min}$, calculate the distance between the sample and all the samples in the minority class set by using the Euclidean distance as the standard and obtain the k-nearest neighbors of the sample.

(2) According to the imbalance ratio of the samples, a sampling ratio is set to determine the sampling rate $n$. For each sample $x \in S_{\min}$ of the minority class, $n$ samples are randomly selected from its k-nearest neighbors, and the selected $n$ samples are $\{y_1, y_2, \ldots, y_n\}$.

(3) For the set of samples $\{y_1, y_2, \ldots, y_n\}$ generated in the previous step and $x \in S_{\min}$, construct new samples according to the following interpolation formula so that each of the minority class will have $n$ new minority samples.

$$x_{new} = x + rand(0,1) \times (y_i - x), i = 1, 2, \ldots, n \quad (1)$$

$rand(0,1)$ in the above formula produces a random number in the range of $(0,1)$. Regarding the sampling rate $n$, its value depends on the imbalance ratio (IR) of the dataset, as shown in equation 2:

$$n = round(IR) - 1 \quad (2)$$

In equation 2, the $IR$ is calculated by dividing the number of majority samples by the number of minority samples. $round(IR)$ is an integer value obtained by rounding the $IR$. In this way, the SMOTE algorithm can balance the number of samples in the dataset, also resolve the overfitting problem of traditional oversampling methods.

## 3 Our proposed method

### 3.1 AUC-based diversified top-k shapelets selection

For the classic shapelets, information gain is used as the standard measure to evaluate candidate. The information gain for a split point $<s, d>$ is defined as [Ye and Keogh, 2009]

$$IG(<s,d>) = E(D) - \frac{n_L}{n}E(D_L) - \frac{n_R}{n}E(D_R) \quad (3)$$

$E(A)$ means the entropy of dataset A. As shown in the above equation, the biggest problem in information gain is that it can study the contribution of a feature to only the whole system, not to a specific class. Hence, when an imbalanced dataset is classified by using information gain as the measure of shapelets, the minority samples might not be classified correctly if the imbalance ratio of the dataset is too large. Thus, for imbalanced datasets, the information gain is no longer effective, and other metrics that are not sensitive to datasets distribution are needed.

To solve the problem of imbalanced time series classification, in this paper, the AUC is used to evaluate the performance of candidate shapelets, and an imbalanced time series classification method based on diversified top-k shapelets (DivIMShapelet) is proposed. The DivIMShapelet algorithm first uses the SAX[Eamonn Keogh and Fu, 2005] technique to reduce the original time series data. When generating the candidate shapelets, we use the imbalanced data evaluation standard AUC to evaluate each candidate shapelet and calculate the corresponding AUC value. After obtaining the AUC value of the candidate shapelets, we construct the diversified top-k shapelet graphs and obtain the diversified top-k shapelets on the graph. The pseudo-code of the DivIMShapelet algorithm to produce diversified top-k shapelets using AUC measure is as in Algorithm 1. The funtion **DivTopKShapeletQuery()** is described in Algorithm 2. We omit the specific explanation of the definition about diversified shapelets query owning to the article length restriction ,and please reference [Yan et al., 2016] for more information.

---

**Algorithm 1** DivIMShapelet ($T$, min, max)

**Input:** training set $T$, minimum length min, maximum length max
**Output:** all candidate shapelets
1: *allShapelets* = ∅
2: **for** all time series $T_i$ in $T$ **do**
3:     *shapelets* = ∅
4:     **for** $l$ = min to max **do**
5:         $W_{i,l}$ = *generateCandidates*($T_i$, min, max)
6:         **for** all subsequences $S$ in $W_{i,l}$ **do**
7:             **for** $m$ = 0 to $|T|$ **do**
8:                 $D_s$ = *calcDistance*($S$, $T_m$)
9:             **end for**
10:             sort($D_s$)
11:             *AUCvalue* = 0
12:             **for** $i$ = 0 to $|D_s|$ **do**
13:                 **for** $j$ = 0 to $i$ **do**
14:                     $T_n$ = $D_s[j].ts$
15:                     **if** $T_n$.label = positive **then**
16:                         $tp = tp + 1$
17:                     **else**
18:                         $fp = fp + 1$
19:                     **end if**
20:                 **end for**
21:                 **for** $j$ = i+1 to $|D_s|$ **do**
22:                     $T_n$ = $D_s[j].ts$
23:                     **if** $T_n$.label = negative **then**
24:                         $tp = tp + 1$
25:                     **else**
26:                         $fp = fp + 1$
27:                     **end if**
28:                 **end for**
29:                 *tpr[i]* = *tp*/#positive
30:                 *fpr[i]* = *fp*/#negative
31:             **end for**
32:             *AUCvalue* = *calcAUC*(*tpr, fpr*)
33:             *allShapelets.add*(*shapelets, AUCvalue*)
34:         **end for**
35:     **end for**
36: **end for**
37: *Graph* = conShapeletGraph(*allShapelets*)
38: *kShapelets* = DivTopKShapeletQuery(*Graph*, $k$)
39: **return** *kShapelets*

---

The first line of algorithm 1 is the initialization of shapelets candidate set; lines 2-5 generate candidate shapelets with a minimum length and a maximum length for all time series in the training set. The function generateCandidates() in line 5 uses the SAX technique to reduce the dimension of the time series; lines 6-10 calculate the distances between all candidate shapelets and each time series in the training set and then sort the distances from smallest to largest; lines 12-31 increase the threshold and calculate the TPR and FPR value. The calculation method is as follows: because the distances in $D_s$ have been ranked, distance is used as the feature and the distance in $D_s$ is chosen as a threshold in turn. If the distance of time series is less than or equal to the threshold, the sequence is predicted to be positive, and the rest are predicted to be negative. In this way, the number of TPs and FPs is calculated, and then the TPR and FPR values are obtained. Lines 32 and 33 calculate the AUC value of each candidate shapelet on the basis of the previous multiple TPR and FPR values and input the shapelets into allShapelets. Line 37 constructs the diversified graph and Line 38 uses algorithm 2 to query and obtain a collection containing k diviersified shapelets and then returns the k shapelets $kShapelets$ to transform the dataset.

---

**Algorithm 2** DivTopKShapeletQuery ($Graph, k$)

**Input:** Diversified Graph of Shapelets Candidates Set $Graph$, top k distinctive shapelets patterns $k$
**Output:** Diversified top-k Shapelets Set *kShapelets*
1: *kshapelets* = ∅, $n$ = |Graph.$V$|
2: *kshapelets.add*($v_1$)
3: **while** |kShapelets| <$k$ **do**
4:     **for** $i$ = 2 to $n$ **do**
5:         **if** $Graph[i] \cap kshapelets = \emptyset$ **then**
6:             *kshapelets.add*($v_i$)
7:         **end if**
8:     **end for**
9: **end while**
10: **return** *kShapelets*

---

### 3.2 A new Oversampling method

After obtaining the top-k shapelet set, we transform the training data by calculating the distance between each time series object in the dataset and each shapelet separately. The new distance matrix can be seen as a new variable space based on the extracted shapelet patterns. Then, to further enhance the classification effect of the algorithm, the SMOTE algorithm is used to oversample the distance matrix of the training minority data. For convenience, this article refers to the proposed algorithm as DivIMShapelet + SMOTE, and the specific process of the algorithm is as in Algorithm 3.

The input of the DivIMShapelet + SMOTE algorithm is the minority training set $T$, the sampling rate $n$, the number of nearest neighbors $m$, and the diversified top-k shapelets $kShapelets$. Each of the top-k shapelets is used to transform each time series in the original dataset into an instance with k attributes, and the value of each attribute is the distance between time series and shapelets. After the training

**Algorithm 3** DivIMShapelet + SMOTE $(T, n, m)$

**Input:** The minority samples in transformed training set $T$, the sampling rate $n$, the number of nearest neighbors $m$, the diversified top-k shapelets *kShapelets*
**Output:** $n * |T|$ minority samples *syntheticData*

1: $syntheticData = \emptyset$
2: $newIndex = 0$
3: **for** $i = 1$ to $|T|$ **do**
4:     $NN = nearestNeighbors(T_i)$
5:     **while** $n \neq 0$ **do**
6:         $j = \text{random}(1, n)$
7:         **for** $k = 1$ to $|kShapelets|$ **do**
8:             $dif = T[i][k] - NN[j][k]$
9:             $gap = \text{rand}(0, 1)$
10:            $syntheticData[newIndex][attr] = T[i][k] + gap * dif$
11:         **end for**
12:         $newIndex$++
13:         $n = n - 1$
14:     **end while**
15: **end for**
16: **return** *syntheticData*

set is transformed by using the diversified top-k shapelets, the transformed minority samples in the training set are then selected as $T$. For each sample in $T$, the m-nearest neighbors are calculated (lines 3-4). Then, n nearest neighbors are randomly selected (lines 5-6) from the m-nearest neighbors (m >n). After that, the selected nearest neighbor sample is interpolated with the samples in $T$ (lines 7-11), according to the formula 1, so that each sample in $T$ generates n artificial samples. Finally, the generated artificial sample set *syntheticData* is returned (line 16).

## 4 Experiments and results

### 4.1 Datasets and experimental design

We select 4 datasets from the UCR time series repository [Chen *et al.*, 2015], including ECG200, Wafer, Fish and Adiac. Table 1 shows the information of the datasets used in this experiment, including the names of the datasets, the lengths of time series, the number of classes, the number of minority classes used, the number of samples in minority and majority class and the imbalance ratio *IR*.

In general, when the ratio of majority samples to minority samples is greater than 2:1, a dataset is considered to be imbalanced. Of the four datasets listed in Table 1, ECG200 and Wafer contain two classes, with one of the classes being a minority class; the imbalance ratios are 2.2 and 9.3, respectively. Fish and Adiac contain multiple classes. Here, we design the following strategy: when a dataset contains multiple classes, we take each class comprising less than 1/3 of the total number of samples as a minority class and the rest as the majority class. If there is more than one minority class, we treat each class as a minority class in turn and average the results. For example, the Adiac dataset contains 37 classes. For each class, because the number of samples per class is between 4 and 15, each class can be treated as a minority class, and the remainder can be regarded as majority classes. The number of majority class samples is between 375-386, and the imbalance ratio of the dataset is between 25 and 96.5.

To illustrate the validity of the DivIMShapelet algorithm and the DivIMShapelet + SMOTE algorithm, this paper tests them by using the four datasets mentioned above. The two algorithms will be compared with DivTopKShapelet [Yan *et al.*, 2016], which also uses diversified top-k shapelets to improve the performance of the time series classification but does not consider the imbalance problem. These three methods are all combined with six classic time series classifiers to determine the performance of each method. We also compare our methods with the INOS + SVM algorithm [Cao *et al.*, 2013], which has been shown to have some advantages in time series classification. Next, we analyze the Accuracy, AUC value and F-measure of each algorithm. Each algorithm is run 20 times on the same dataset, and the average is taken as the final result.

### 4.2 Accuracy analysis

Table 2 shows the classification accuracy of the DivTopKShapelet, DivIMShapelet, DivIMShapelet + SMOTE, and INOS + SVM algorithms. The bold numbers in the table indicate the highest classification accuracy achieved for the dataset.

In Table 2, the DivIMShapelet and DivIMShapelet + SMOTE algorithms are shown to have a better effect on the classification accuracy than the DivTopKShapelet algorithm. The DivIMShapelet + SMOTE algorithm works best. Compared with that of the INOS + SVM algorithm, the accuracy of the DivIMShapelet algorithm on the ECG200, Wafer and Fish datasets is very similar, but the accuracy of the Adiac dataset is low because the imbalance ratio of Adiac is too high for DivIMShapelet to adapt to. After the SMOTE algorithm is used to oversample the transformed training set, the classification accuracy of the DivIMShapelet + SMOTE algorithm is better than that of the INOS + SVM algorithm, owing to the increased number of representative features of the minority class in the training sets.

For the ECG200 dataset, if it is directly classified by using the DivTopKShapelet algorithm combined with the Random Forest classifier, the best classification accuracy is 82.0%. After the AUC value is used as the measure of shapelets, the best classification accuracy of DivIMShapelet combined with the Random Forest classifier is 88.0%, a percentage similar to that of the INOS + SVM algorithm. After the SMOTE algorithm is used to balance the transformed dataset, the classification accuracy of the Random Forest classifier is 99.2%. Similar results can also be derived for the Fish and Adiac datasets. For the Wafer dataset, the accuracy of the DivIMShapelet + SMOTE algorithm combined with the 1NN and Rotation Forest classifiers is 100%, thereby indicating that all samples in the dataset are classified correctly. Therefore, the DivIMShapelet + SMOTE algorithm can significantly improve the classification accuracy.

### 4.3 AUC analysis

Next, we compare the AUC values of the DivTopKShapelet, DivIMShapelet, DivIMShapelet + SMOTE, and INOS +

Table 1: Description of the Datasets

| Dataset | Length | Classes | Minority classes | Size of training data | | | Size of testing data | |
|---|---|---|---|---|---|---|---|---|
| | | | | minority | majority | *IR* | minority | majority |
| ECG200 | 96 | 2 | 1 | 31 | 69 | 2.2 | 36 | 64 |
| Wafer | 152 | 2 | 1 | 97 | 903 | 9.3 | 665 | 5509 |
| Fish | 463 | 7 | 7 | 21–28 | 147–154 | 5.3–7.3 | 22–29 | 146–153 |
| Adiac | 176 | 37 | 37 | 4–15 | 375–386 | 25–96.5 | 6–16 | 375–385 |

Table 2: Classification accuracy of different algorithms(%)

| algorithm | classifier | Dataset | | | |
|---|---|---|---|---|---|
| | | ECG200 | Wafer | Fish | Adiac |
| DivTopKShapelet | C4.5 | 79.0 | 96.4 | 60.8 | 49.4 |
| | 1NN | 78.0 | 97.9 | 80.1 | 56.3 |
| | Naive Bayes | 80.0 | 89.0 | 71.9 | 57.5 |
| | Bayesian Network | 81.0 | 96.9 | 65.1 | 38.9 |
| | Random Forest | 82.0 | 97.2 | 74.9 | 58.8 |
| | Rotation Forest | 80.0 | 98.0 | 85.5 | 60.1 |
| DivIMShapelet | C4.5 | 86.0 | 97.9 | 75.7 | 65.0 |
| | 1NN | 84.0 | 99.4 | 89.7 | 66.2 |
| | Naive Bayes | 81.0 | 89.0 | 85.3 | 70.9 |
| | Bayesian Network | 83.0 | 95.2 | 74.4 | 62.5 |
| | Random Forest | 88.0 | 97.3 | 79.9 | 61.6 |
| | Rotation Forest | 87.1 | 99.6 | 89.5 | 63.3 |
| DivIMShapelet + SMOTE | C4.5 | 90.2 | 98.7 | 85.1 | 88.4 |
| | 1NN | 95.0 | **100** | **93.3** | 80.6 |
| | Naive Bayes | 98.0 | 98.2 | 90.9 | **96.7** |
| | Bayesian Network | 92.0 | 94.9 | 89.3 | 94.7 |
| | Random Forest | **99.2** | 99.4 | 89.8 | 97.0 |
| | Rotation Forest | 96.0 | **100** | 87.0 | **96.7** |
| INOS | SVM | 86.0 | 97.6 | 85.9 | 82.2 |

Table 3: AUC values of different algorithms

| algorithm | classifier | Dataset | | | |
|---|---|---|---|---|---|
| | | ECG200 | Wafer | Fish | Adiac |
| DivTopKShapelet | C4.5 | 0.66 | 0.55 | 0.75 | 0.78 |
| | 1NN | 0.87 | 0.60 | 0.79 | 0.79 |
| | Naive Bayes | 0.88 | 0.64 | 0.76 | 0.81 |
| | Bayesian Network | 0.79 | 0.65 | 0.72 | 0.85 |
| | Random Forest | 0.86 | 0.61 | 0.86 | 0.77 |
| | Rotation Forest | 0.90 | 0.61 | 0.82 | 0.81 |
| DivIMShapelet | C4.5 | 0.78 | 0.75 | 0.79 | 0.83 |
| | 1NN | 0.89 | 0.79 | 0.84 | 0.84 |
| | Naive Bayes | 0.94 | 0.83 | 0.85 | 0.83 |
| | Bayesian Network | 0.89 | 0.72 | 0.86 | 0.88 |
| | Random Forest | 0.90 | 0.76 | 0.80 | 0.80 |
| | Rotation Forest | 0.91 | 0.74 | 0.87 | 0.84 |
| DivIMShapelet + SMOTE | C4.5 | 0.92 | 0.95 | 0.90 | 0.96 |
| | 1NN | 0.93 | **1** | 0.93 | **0.99** |
| | Naive Bayes | 0.98 | 0.96 | 0.95 | 0.94 |
| | Bayesian Network | 0.94 | 0.96 | 0.96 | 0.97 |
| | Random Forest | **0.95** | 0.99 | 0.95 | 0.97 |
| | Rotation Forest | **0.95** | **1** | **0.97** | 0.98 |
| INOS | SVM | 0.90 | 0.93 | 0.88 | 0.89 |

Table 4: F-measure values of different algorithms

| algorithm | classifier | Dataset | | | |
|---|---|---|---|---|---|
| | | ECG200 | Wafer | Fish | Adiac |
| DivTopKShapelet | C4.5 | 0.72 | 0.61 | 0.78 | 0.80 |
| | 1NN | 0.83 | 0.62 | 0.76 | 0.81 |
| | Naive Bayes | 0.78 | 0.65 | 0.77 | 0.85 |
| | Bayesian Network | 0.80 | 0.63 | 0.71 | 0.86 |
| | Random Forest | 0.85 | 0.68 | 0.88 | 0.79 |
| | Rotation Forest | 0.82 | 0.60 | 0.80 | 0.78 |
| DivIMShapelet | C4.5 | 0.87 | 0.78 | 0.80 | 0.82 |
| | 1NN | 0.88 | 0.80 | 0.82 | 0.85 |
| | Naive Bayes | 0.93 | 0.85 | 0.86 | 0.86 |
| | Bayesian Network | 0.90 | 0.82 | 0.85 | 0.81 |
| | Random Forest | 0.91 | 0.78 | 0.83 | 0.88 |
| | Rotation Forest | 0.92 | 0.76 | 0.78 | 0.83 |
| DivIMShapelet + SMOTE | C4.5 | 0.93 | 0.95 | 0.94 | 0.95 |
| | 1NN | 0.92 | **1** | 0.95 | **0.98** |
| | Naive Bayes | **0.97** | 0.95 | 0.92 | 0.92 |
| | Bayesian Network | 0.95 | 0.97 | 0.93 | 0.93 |
| | Random Forest | 0.96 | 0.98 | 0.96 | 0.94 |
| | Rotation Forest | 0.95 | **1** | **0.97** | 0.96 |
| INOS | SVM | 0.90 | 0.96 | 0.83 | 0.80 |

SVM algorithms. Table 3 shows the AUC results for these four algorithms. The bold numbers in the table indicate the maximum AUC value achieved for a dataset.

It is clearly seen in Table 3 that the DivIMShapelet algorithm has a higher AUC value than the DivTopKShapelet algorithm, but as compared with the INOS + SVM algorithm, the values are similar, and the distinction is not obvious. After use of the SMOTE algorithm, the AUC value of the DivIMShapelet + SMOTE algorithm improves significantly, and DivIMShapelet + SMOTE has the best classification results among all four algorithms.

For the Wafer dataset, the DivIMShapelet + SMOTE algorithm, combined with the 1NN and Rotation Forest classifiers, has an AUC value of 1, which indicates that all samples in the test set are correctly classified. For the ECG200 dataset, the maximum AUC is 0.95, which is obtained through combination with the Random Forest and Rotation Forest classifiers, and the highest AUC values for the Fish and Adiac datasets are 0.97 and 0.99, respectively.

### 4.4 F-measure analysis

Similarly to the previous section, this section compares the F-measure value of the DivTopKShapelet algorithm, the DivIMShapelet algorithm, the DivIMShapelet + SMOTE algorithm, and the INOS + SVM algorithm. Table 4 shows the F-measure for these four algorithms. The bold numbers in the table represent the largest F-measure value achieved for a dataset.

As shown in Table 4, the DivIMShapelet algorithm is significantly better than the DivTopKShapelet algorithm when the evaluation criteria for shapelets is changed to AUC, and the optimal F-measure values of the ECG200, Fish and Adiac datasets are greater than that of the INOS + SVM algorithm, thus demonstrating that the DivIMShapelet algorithm has a better adaptability to imbalanced data classification. Furthermore, after oversampling the transformed dataset by using SMOTE, the F-measure value of the DivIMShapelet + SMOTE algorithm is improved significantly.

For the ECG200 dataset, the maximum F-measure value of the DivIMShapelet + SMOTE algorithm is 0.97, which is obtained after combining with the Naive Bayes classifier. The F-measure value of the DivIMShapelet + SMOTE algorithm, combined with the 1NN and Rotation Forest classifiers, reaches 1 for the Wafer dataset; the maximum F-measure values for the Fish and Adiac datasets are 0.97 and 0.98, respectively.

In summary, for imbalanced time series classification, the DivTopKShapelet algorithm cannot achieve a satisfactory result; the DivIMShapelet algorithm is only slightly better than the DivTopKShapelet algorithm, and its effect is similar to that of the INOS + SVM algorithm. The results show that the DivMAXShapelet + SMOTE algorithm has the best effect after the transformed dataset is oversampled by using the SMOTE algorithm, and this approach significantly improves the classification accuracy, AUC value and F-measure.

## 5 Conclusions

In this paper, we presents a new oversampling method for imbalanced TSC. In order to extract the most representative and distinctive features for time series objects, **DivIMShapelet** algorithm is firstly proposed, which is based on diversified top-k shapelets and adopts the AUC as the metric to measure the shapelets quality. Using **DivIMShapelet** algorithm, we transform the original training data to the new feature space. Then, a new SMOTE method, named as **DivIMShapelet+SMOTE** is used to oversample the feature vectors in the feature space and achieve the rebalance of the original dataset. Experiments are carried out to validate our proposed methods performance.

## Acknowledgments

This work is supported by the Youth Science Foundation of China University of Mining and Technology under Grant No (2013QNB16), Natural Science Foundation of Jiangsu Province of China (BK20140192).

# Predicting Concept Drift Severity

**Ruolin Jia**
The University of Auckland,
New Zealand
rjia477@aucklanduni.ac.nz

**Yun Sing Koh**
The University of Auckland,
New Zealand
ykoh@cs.auckland.ac.nz

**Gillian Dobbie**
The University of Auckland,
New Zealand
gill@cs.auckland.ac.nz

## Abstract

In data streams, concept drift severity refers to the amount of change of a concept. Low severity concept drifts are hard to detect using drift detectors, and in these cases, a high detection sensitivity is expected. However, high detection sensitivity comes with a higher false positive rate that degrades the performance of the drift detector. In this paper, we present PRESS (PREdictive Severity Seed) detector. PRESS learns the severity trends of a stream with recurrent volatility (the frequency of experiencing concept drifts) and predicts the severity of future concept drifts with a probability network. Experiments show that PRESS outperformed other existing drift detectors by reducing false positive rates while maintaining true positive rates. To the best of our knowledge, it is the first algorithm predicting the severity of concept drifts to improve the detection performance. In addition, we provide a simple yet effective severity measurement method, which does not require a detection warning period, for Seed and ADWIN detectors.

## 1 Introduction

In real-world cases, data streams experience frequent change of the underlying distribution, and this is known as concept drift. To mine a data stream, it is important to detect concept drifts to maintain the quality of the models. Various drift detection methods [Page, 1954; 1954; Gama *et al.*, 2004; Baena-Garcıa *et al.*, 2006; Bifet and Gavalda, 2007] have gained attention. Concept drifts have different properties that can characterise them. One property is concept drift volatility that represents the frequency of experiencing concept drifts. The other property is the amount or magnitude of a concept drift, known as drift severity [Kosina *et al.*, 2010]. Different levels of drift severity can influence the behaviour of a drift detector, thus tracking drift severity of a data stream is necessary. Although various methods have been proposed to track drift severity [Kosina *et al.*, 2010; Chen *et al.*, 2015; Webb *et al.*, 2016], until now, there are no existing methods for predicting the severity trends in a stream.

In this paper, we investigated how drift severity affects the behaviours of drift detectors. We found that it is difficult for current well-known drift detectors such as Seed [Huang *et al.*, 2014] and ADWIN [Bifet and Gavalda, 2007] to detect concept drifts if the drifts have low severity in the stream. In this case, a detector with high detection sensitivity is needed. The tradeoff of a current detector with high sensitivity is it may result in more false alarms. To overcome these challenges, we proposed the PRESS (PREdictive Severity Seed) drift detector. We assume that drift severity follows the same distribution if the frequency of occurrences (volatility) is constant. PRESS has two phases: training phase and testing phase. In the training phase, PRESS uses a probability network to learn the volatility changing patterns of a stream. Then, for each volatility pattern, it samples the corresponding severity values. In the testing phase, by using learned severity values, PRESS applies high detection sensitivity only when the severity is low. This technique has been shown, in most cases, to effectively reduce false positive rates compared with existing detectors while maintaining similar true positive rates and detection delays. In addition to PRESS, we also proposed a novel way of measuring drift severity for ADWIN and Seed detectors.

PRESS works under the assumption that many data streams concept drifts in the real-world follow behavioural patterns. This characteristic has been exploited by previous research in the area method to capture the regularity of these concept drifts [Kosina *et al.*, 2010; Chen *et al.*, 2015]. In such real-world cases with regular concept drifts, each concept drift is caused by some events. For example, in a manufacturing production line, concept drifts can arise due to the regular machine faults and these events are also regular based on the life span of the machine parts. This assumption exist in many different real world application that follows pattern regularity. Concept drift caused by the same events also share the same severity. In the case of regular concept drifts, the severity and volatility both follow some distribution patterns. In the event that no behavioural patterns exist, PRESS defaults to using a base state-of-the-art detector, where the accuracy of its results are no worse than the current base technique such as Seed.

In Section 2 we review data stream drift detection. In Section 4, we discuss background before presenting our methods. In Sections 5 and 6, we describe our PRESS algorithm and demonstrate its performance through experimentation. We then conclude our paper.

## 2 Related Work

In data stream mining, concept drift occurs when the underlying distribution of data streams changes. Formally, a concept drift is a change in the joint probability $P(\vec{x}, y) = P(y|\vec{x}) \times P(\vec{x})$ in which $\vec{x}$ is the input attributes vector and $y$ is the dependent class label [Gama, 2010]. Drift detectors are applied to detect when a concept drift occurs. Detectors monitor the stream of some performance indicators, such as the prediction accuracy over time, of the predictor and alarm a drift when finding a change in the corresponding indicator [Klinkenberg and Renz, 1998]. Several drift detection algorithms [Page, 1954; Gama et al., 2004; Baena-Garcıa et al., 2006; Bifet and Gavalda, 2007] monitor such streams of indicators. In reality, these detectors often monitor the error stream of a predictor and signal a drift when the error increases significantly. Because drift detectors cannot make the perfect approximation of a stream and real-world streams are largely susceptible to noise, detectors may signal a false alarm, known as false positive, when in fact, there is no actual concept drift in the stream. To reduce false positive rates, Huang et al. [2015] use stream volatility, a measure of how frequent the concept drift occurrence is in a stream, to capture the trend of concept drifts. Based on these trends, their algorithms predicts the future drift positions and probability. Then it reduces the sensitivity of the drift detector when the concept drift is unlikely to happen. Thus, the false positive rate is reduced. An algorithm sharing the same idea is ProSeed [Chen et al., 2016]. It builds a probability network to capture the recurrent volatility of a data stream that transits among a few volatility patterns. Then it uses the network to predict the position of future drifts and only tries to find a drift in those instances where drifts are likely.

A data stream can experience different amounts of drift represented as drift severity. Kosina et al. [2010] presented a metric for severity measurements: when the stream with input space $S$ changes from distribution $C_1$ to $C_2$, the percentage of $S$ in $C_2$ with different class labels from it in $C_1$ is the drift severity. To estimate this percentage, they enable a warning period of the drift detector when it detects the stream is going to, but yet to, experience a drift, and it counts the number of misclassifications in this period. After a drift is detected, it computes the drift severity as

$$Severity = \frac{\#misclassified}{\#instances\ in\ warningperiod}.$$

Generally, this approach measures the amount of change in the one-dimensional indicator stream of the classifier. Additionally, to apply this metric measurement, one also needs to enable the existing detector to detect the warning period. Kosina et al. [2010] use two different thresholds: the drift warning can be detected with the lower threshold before detecting an actual drift. Recently Chen et al. [2015] proposed the MagSeed algorithm that is a warning level enabled version of the original Seed detector. Similarly, it applies a relaxed threshold to detect a warning level prior to detecting the actual drift with the drift threshold. Then it calculates the severity of a stream. However, to the best of our knowledge, all current approaches for tracking drift severity only capture the severity of previous drifts, and they do not proactively predict the severity trends.

## 3 Severity Detection Method Without Warning Periods

In this paper, we follow the general definition of concept drift severity proposed by Webb et al. [2016]. At any time $t$ and $u$, the concept drift severity can be calculated by $Severity_{t,u} = D(t, u)$ where $D$ is some distance function returning a non-negative value. For example, one can compute the joint probability distance between $P_t(X, Y)$ and $P_u(X, Y)$ to show the concept drift magnitude in a classification task.

Methods building upon the severity metric by Kosina et al. [2010] requires a warning period. A major limitation of the warning period is that one needs to set an appropriate warning threshold. A poorly set threshold may result in either the detector directly signalling a drift without encountering any warning periods, or it enters the warning period that is not followed by an actual drift (false warning alarm). In this section, we present a method that is suitable to be used with current detectors, such as ADWIN and Seed. Our method detects concept drift severity without requiring a warning period. Our technique compares the distance between two concept drifts in a predictor's indicator stream, for example, an error stream. We calculate the distance between stream distributions after two consecutive drifts $d_1$ and $d_2$ as the drift severity of drift $d_2$.

ADWIN and Seed both signal a drift when they detect a significant difference between two sub-window $W_1$ and $W_2$, where $W_2$ represents the most recent instances after a drift occurs. Whenever a drift happens, we store the mean of all instances in $W_2$ in the variable called "snapshot" $S$. When a new drift is detected, we then compute the severity (the distance) as $|\overline{W_2} - S|$, and then assign the mean of all instances in $W_2$ to $S$. We repeat this process when the detector performs its task.

## 4 Relationship between Severity and Detector Performances

Different levels of severity affect a drift detector's performance. We show the influences of these through experimentation. We use Seed and ADWIN detectors with different confidence levels as the base drift detector for this experiment. We run those two detectors on our synthetic data streams. The detail about generating these data streams are discussed in Section 6.

To show our findings, we plot the relationship between severity v.s. false negatives and detection delays for the Seed detector in Figure 1. We use different colours to represent changing confidence levels. The experiment concludes that, a concept drift with higher severity is "easier" to be detected when the severity is relatively high, because these experiments show that the detector experiences lower delay and false negative rate when processing a high severity drift. Thus, it is not necessary to set a relatively high detection sensitivity in the high severity case, given that a high detection sensitivity will result in an increasing false positive rates. A similar phenomenon was also seen using the ADWIN detector.

In general, the result suggests that it is feasible to attempt to improve the performance of drift detection by reducing the sensitivity of the drift detector when the severity is high.

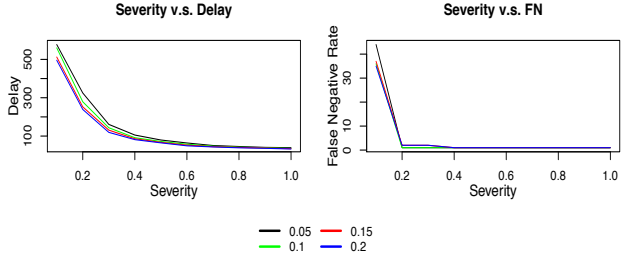

Figure 1: Severity and Drift Detector Behaviour

## 5 PRESS Algorithm

In this section we review the drift detector SEED, which we use in conjunction with our method to demonstrate the advantages of PRESS. We then detail our PRESS algorithm.

### 5.1 Review of Seed

Because PRESS uses the Seed [Huang *et al.*, 2014] detector as its component, we firstly review Seed. The Seed algorithm uses a sliding window storing the most recent instances seen in a stream. In addition, it accumulates instances in the window into fixed-size blocks. Seed assumes that instances of one block follow the same concept. To detect a drift, Seed compares the statistical difference between each of all the two possible sub-windows containing those blocks and signals a drift when the difference is greater than the threshold. For example, if Seed has three blocks $B_1, B_2, B_3$. It compares the difference between sub-window $\{B_1\}$ and sub-window $\{B_2, B_3\}$, and between $\{B_1, B_2\}$ and $\{B_3\}$.

In any partition of Seed's window, the Seed detector alarms a drift when $|\mu_{\hat{W}_0} - \mu_{\hat{W}_1}| > \epsilon$ where $\mu_{\hat{W}_0}$ and $\mu_{\hat{W}_1}$ are two sub windows' means and $\epsilon$ is a threshold value calculated by:

$$\epsilon = \sqrt{\frac{2}{m} \cdot \sigma_W^2 \cdot \ln \frac{2}{\delta'}} + \frac{2}{3m} \cdot \ln \frac{2}{\delta'}, \; \delta' = \frac{\delta}{n},$$
$$m = \frac{1}{1/n_0 + 1/n_1}$$

where $\sigma_W^2$ is the variance of the window; $n_0$ and $n_1$ are instance counts of two sub-windows of the partition; $\delta$ is a user set confidence level in range $(0, 1)$.

### 5.2 PRESS

In this section, we present the PRESS (PREdictive Severity Seed) algorithm.

We make the following assumption about the drift severity: in a stream fluctuating among a few fixed volatility patterns, if the volatility changes from a pattern $p_1$ to a pattern $p_2$ at time $t_1$, the severity of $p_2$ has the same distribution as $p_2$ when it changes from $p_1$ at another time $t_2$. In short, the most recent volatility transition determines the current pattern's concept drift severity at any time. In this case, severity changes also follow some fixed patterns. The assumption is reasonable if the same volatility transitions are caused by the same event. For example, the data generated by a manufacturing equipment monitoring sensor may experience recurrent volatility because the equipment may experience regular faults. A faulty event will cause a volatility transition such that the concept drift is becoming more frequent. In addition, upcoming concept drifts in the new volatility pattern are caused by the same factor - the faulty event, so they will share the same concept severity.

We use an example shown in Figure 2 to discuss our technique. In the two plots, the time element is signalled by the arrival of new instances. In the top plot (volatility v.s. instance), we observed that the volatility pattern changes among three patterns $p_1$, $p_2$, and $p_3$. Whenever we experience the same pattern transition we assume that the severity after the transition would be similar. In our example, there are two transitions from $p_1$ to $p_2$, so in the bottom plot (severity v.s. instance), the severity level denoted by "$p_1$ to $p_2$" in both of the cases are similar.

The intuition of PRESS is to learn the volatility transition using a probability network, and sample the concept drift severity for each volatility pattern by using reservoir sampling [Vitter, 1985]. Then, PRESS predicts the severity of future drifts and adjusts the Seed detector threshold correspondingly, so that it reduces the detector sensitivity by augmenting the threshold when the predicted severity is relatively low.

PRESS uses the volatility change detector presented by Huang et al. [2014]. It takes input of drift intervals. Whenever a drift is detected, it calculates the interval between the most recent drift and its previous drift. Then it inputs the interval value to the volatility change detector. The detector will alarm a change of volatility when it finds a significant difference between recent intervals and previous ones.

PRESS has a training phase and a testing phase. In the training phase, PRESS learns a probability network of stream volatility changes using the same approach as Chen et al. [2016]. In the network, each node is a certain volatility pattern containing samples from the distribution of one volatility state. One example of the pattern is $p = \{110, 108, 99\}$. Each number in $p$ denotes the interval between two drifts. All numbers in $p$ are assumed to be from one volatility distribution. Each edge of the network represents the probability of transiting from one volatility state to another state. Whenever a volatility change is detected, the detector can also return the most recent volatility pattern. Then, the most recently seen volatility pattern is compared with each pattern in the network using the Kolmogorov-Smirnov test. If the most recently seen pattern is not similar to any pattern in the network, we add it to the network along with a new transition edge. For example, the current network contains two patterns $p_1 = \{110, 108, 99\}$ and $p_2 = \{210, 190, 200\}$. The stream is currently in pattern $p_2$. When the volatility change detector detects a new pattern $p_3 = \{500, 490, 470\}$, it decides that $p_3$ is not statistically similar to any existing pattern, so $p_3$ will be added to the network and we add the transition edge from $p_2$ to $p_3$. In the other case, if we find a match of patterns, we simply insert samples of the most recently seen intervals to the matched patterns in the network and update the transition probability. By default, each pattern in the network can have up to 100 intervals. If the pattern is full, the oldest intervals are deleted to create space for new ones. The probability update is calculated based on the frequency of transitions. For example, we have seen that pattern $p_2$ has transited 3 times to $p_3$ and 2 times to $p_1$, so the transition probability from $p_2$ to $p_3$ and $p_1$ are 0.6 and 0.4 respectively.

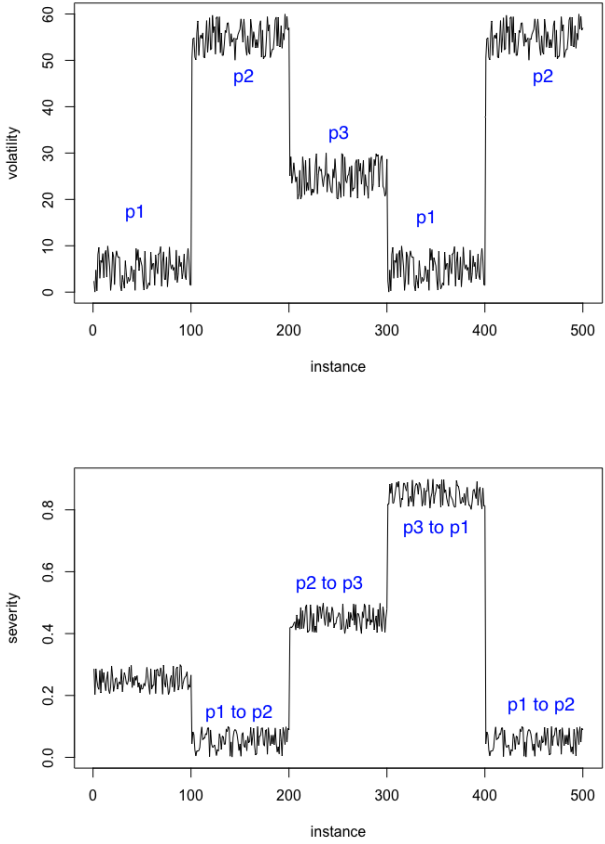

Figure 2: Relationship of Volatility and Severity

When learning the network, PRESS also maintains a severity buffer of severity and a set of reservoirs denoted by $R(x, y)$ where $x$ and $y$ are any two volatility patterns. The severity buffer is a sliding window storing recently seen concept drift severity. It records a severity value whenever a drift occurs. When a volatility change is detected, PRESS inserts all samples in the buffer to $R(p_j, p_i)$ where $p_i$ is the most recently seen pattern and $p_j$ is the pattern seen before $p_i$. Then, it clears the buffer and repeats the process. For example, before detecting a volatility change, the stream is in pattern $p_1$, and the volatility change detector detects a volatility change to $p_2$, we add all severity values in the buffer to $R(p_1, p_2)$. PRESS uses the severity detection method in Section 3.

In the testing phase, PRESS is able to predict the severity of future concept drifts by using the network and severity reservoirs. It adjusts the threshold based on the predicted future severity. If the predicted severity in upcoming drifts is relatively high, it reduces the sensitivity by increasing the detection threshold. We calculate the threshold $\epsilon$ of PRESS using the following formula:

$$\epsilon = c \cdot \sqrt{\frac{2}{m} \cdot \sigma_W^2 \cdot \ln \frac{2}{\delta'}} + \frac{2}{3m} \cdot \ln \frac{2}{\delta'}$$

This is a variation of the Hoeffding bound. Compared with the previous Seed detector's threshold formula, it has an additional $c$ coefficient multiplying its first term. $c$ is the threshold augmenting coefficient computed by $c = 1 + \beta \cdot \phi$. In this formula, $\beta$ is a user set parameter controlling the magnitude of threshold augmentation. $\phi$ is the predicted severity value in range $(0, 1)$.

---

**Algorithm 1:** Learning Phase

**input** : $S$: Stream;
1 Let $Seed$ be the Seed Detector;
2 Let $VolDect$ be the Volatility Change Detector;
3 Let $B$ be the severity buffer;
4 Let $R(x, y)$ denotes the severity reservoir of pattern $x$ if $x$ is transited from pattern $y$;
5 Let $N$ be the probability network;
6 **begin**
7     **foreach** $s \in S$ **do**
8         Input $s$ in $Seed$;
9         **if** $Seed$ detects a drift **then**
10             Let $i$ be the interval between the most recent two drifts;
11             Input $i$ in $VolDect$;
12             Compute the drift severity $sv$ and input $sv$ in $B$;
13             **if** $VolDec$ detects a volatility change **then**
14                 Let $p_i$ be the most recent seen pattern;
15                 Let $p_j$ be the pattern seen before $p_i$;
16                 **if** node $p_i$ does not exist in $N$ **then**
17                     add node $p_i$ to $N$;
18                     add edge $E(p_j, p_i)$ to $N$;
19                 increment probability of $E(p_j, p_i)$;
20                 **if** $R(p_j, p_i)$ does not exist **then**
21                     Create empty reservoir $R(p_j, p_i)$;
22                 Input all instances of $B$ to $R(p_j, p_i)$;
23                 Clear $B$;
24     **end for**
25 **end**

---

Predicted severity $\phi$ is the expected severity value of the

**Algorithm 2:** Predicting Phase

**input** : $S$: Stream;
1 Let $Seed$ be the Seed Detector;
2 Let $VolDect$ be the Volatility Change Detector;
3 Let $B$ be the severity buffer;
4 Let $R(x, y)$ denotes the severity reservoir of pattern $x$ if $x$ is transited from pattern $y$;
5 Let $\overline{R}(x, y)$ denotes the mean of $R(x, y)$;
6 Let $Prob(y|x)$ denotes the probability of transiting to $y$ given pattern $x$;
7 Let $\beta$ be a parameter in range $(0, 1)$;
8 Let $Seed.c$ be the threshold coefficient of $Seed$ detector;
9 **begin**
10     Normalise all severity samples in $R(x, y)$ in range $(0, 1)$ for any pattern x and y;
11     **foreach** $s \in S$ **do**
12         Input $s$ in $Seed$;
13         **if** $Seed$ detects a drift **then**
14             Let $i$ be the interval between the most recent two drifts;
15             Input $i$ in $VolDect$;
16             **if** $VolDec$ detects a volatility change **then**
17                 Let $p_i$ be the most recent seen pattern;
18                 Let $H$ be the set of nodes where $Prob(H_j|p_i) > 0$;
19                 Set $\phi = \sum_{j=1}^{|H|}(Prob(H_j|p_i) \cdot \overline{R}(p_i, H_j))$;
20                 Set $Seed.c$ as $(1 + \beta \cdot \phi)$;
21     **end for**
22 **end**

incoming concept drifts calculated by:

$$\phi = \sum_{j=1}^{k}(Prob(H_j|p_i) \cdot \overline{R}(p_i, H_j))$$

where $p_i$ is the most recently seen volatility pattern and $H$ is the set of all patterns transited from $p_i$ obtained from the probability network. $Prob(y|x)$ denotes the probability of transiting to pattern $y$ given the current pattern $x$. $\overline{R}$ is the mean of the severity reservoir. We show the pseudo-code in Algorithms 1 and 2.

The learning phase can be ran online. Every time a new instance arrives, the probability network structure is updated. The drawback of this is a slightly high computational cost. In reality batch processing, *i.e.* only updating the network after a number of instances, can be applied because it is likely that the same concept drift pattern will be constant for a certain period and it is unnecessary to update the network for every instance. In batch processing, we can drop also the old network after a period and then re-train a new one, so the algorithm will switch between the training phase and testing phase, to capture the most recent models.

Caveat: PRESS algorithm works well when there are some regularity of concept drift patterns. However it may not work well on some data streams, especially those without regular concept drift patterns. In these cases, the results will default to the base detector used. Thus performing no worst in terms of accuracy when compared current state-of-the-art techniques. Despite this many data streams concept drifts in the real-world follow behavioral patterns. In such real-world cases with regular concept drift, each concept drift is caused by some event (for example, concept drifts arise due to regular machine faults) and these events are also regular. Concept drift caused by the same event also share the same severity. So in the case of regular concept drifts, the severity and volatility both follow some distribution.

# 6 Experiments

Our experiments compare the performances of four drift detectors: PRESS, ProSeed, Seed, ADWIN. We test each detector with confidence level of 0.05, 0.1, 0.15, 0.2, 0.25. We carried out testing in both real world and synthetic datasets.

## 6.1 Synthetic Data

In this set of experiments, we tested our drift detectors with our synthetic data. Our synthetic generator outputs a series of values simulating the performance indicator of the predictor. The output of the stream is generated by formula $\mu + G$ where $\mu$ is an integer and $G$ is a noise term, a Gaussian random variable with mean 0 and standard deviation 1.

To create drifts with different severity, we add different magnitudes of change to $\mu$. We create datasets with both abrupt drift and gradual drift. Our synthetic datasets contain drifts with recurrent volatility and severity. For both abrupt and gradual concept drift streams, we generate datasets with different amounts of volatility patterns including 3 patterns, 5 patterns, and 10 patterns. The severity of the concept drift is determined by the transition of the previous pattern to the current pattern. (We follow the assumption in Section 5.2). Each dataset contains 1 million instances with 50000 concept drifts. We generate 50 copies for each dataset with different random seeds.

We measured False Positive Rate (FP), True Positive Rate (TP) and Delay. False positives is the count of false alarms in in a stream. The true positive is the count of actual drift that is successfully detected. Delay is the average count of instances between the actual drift place and the time of detecting it. For PRESS, we set $\beta = 0.4$. We set 32 as the block size for Seed and the volatility change detector.

The numbers in all results tables are average results across the 50 runs, and the standard deviation of the results are provided in the round brackets.

## 6.2 Evaluation of the True Positive Rate (TP) and Delay

We firstly evaluate the True Positive Rate (TP) and detection delay among four algorithms. We expect PRESS to have similar TP and Delay to other algorithms. TP in results tables is a rate of corrected detected concept drifts out of all real concept drifts in a stream.

For abrupt drift, we show the results for 10 patterns (Table 1). We note that the results for 3 and 5 patterns' results follow the same trend. We observe perfect TP (approximately

Table 1: 10 patterns, abrupt change

**PRESS**

| $\delta$ | Avg. FP (SD) | Avg. TP (SD) | Delay (SD) |
|---|---|---|---|
| 0.05 | $1.28\times10^{-4}$ ($8\times10^{-6}$) | 100 (0.0014) | 19.54(0.2) |
| 0.10 | $1.57\times10^{-4}$ ($1.3\times10^{-5}$) | 100 (0.0016) | 19.04(0.15) |
| 0.15 | $1.98\times10^{-4}$ ($1.4\times10^{-5}$) | 100 (0.0016) | 18.86(0.13) |
| 0.20 | $2.38\times10^{-4}$ ($2.3\times10^{-5}$) | 100 (0.0015) | 18.63(0.17) |
| 0.25 | $2.84\times10^{-4}$ ($1.7\times10^{-5}$) | 100 (0.0018) | 18.58(0.08) |

**ProSeed**

| $\delta$ | Avg. FP (SD) | Avg. TP (SD) | Delay (SD) |
|---|---|---|---|
| 0.05 | $1.25\times10^{-4}$ ($7\times10^{-6}$) | 100 (0.0016) | 18.98(0.16) |
| 0.10 | $1.89\times10^{-4}$ ($1.9\times10^{-5}$) | 100 (0.002) | 18.53(0.16) |
| 0.15 | $2.65\times10^{-4}$ ($1.3\times10^{-5}$) | 100 (0.0025) | 18.43(0.09) |
| 0.20 | $3.56\times10^{-4}$ ($2.1\times10^{-5}$) | 100 (0.0014) | 18.31(0.12) |
| 0.25 | $4.57\times10^{-4}$ ($2.5\times10^{-5}$) | 100 (0.003) | 18.18(0.16) |

**Seed**

| $\delta$ | Avg. FP (SD) | Avg. TP (SD) | Delay (SD) |
|---|---|---|---|
| 0.05 | $5.15\times10^{-4}$ ($4\times10^{-4}$) | 100 (0.0014) | 18.87(0.2) |
| 0.10 | $6.11\times10^{-4}$ ($4.33\times10^{-4}$) | 100 (0.0015) | 18.48(0.16) |
| 0.15 | $7.51\times10^{-4}$ ($5\times10^{-4}$) | 100 (0.002) | 18.39(0.12) |
| 0.20 | $9.08\times10^{-4}$ ($5.67\times10^{-4}$) | 100 (0.002) | 18.22(0.16) |
| 0.25 | $1.08\times10^{-3}$ ($6.41\times10^{-4}$) | 100 (0.0022) | 18.1(0.21) |

**ADWIN**

| $\delta$ | Avg. FP (SD) | Avg. TP (SD) | Delay (SD) |
|---|---|---|---|
| 0.05 | $7.27\times10^{-4}$ ($8\times10^{-6}$) | 100 (0) | 17.76(0.08) |
| 0.10 | $1.02\times10^{-3}$ ($1.1\times10^{-5}$) | 100 (0.0008) | 17.62(0.15) |
| 0.15 | $1.38\times10^{-3}$ ($1.3\times10^{-5}$) | 100 (0.001) | 17.34(0.16) |
| 0.20 | $1.81\times10^{-3}$ ($1.1\times10^{-5}$) | 100 (0.0011) | 17.29(0.13) |
| 0.25 | $2.28\times10^{-3}$ ($1.8\times10^{-5}$) | 100 (0.001) | 17.16(0.15) |

Table 2: 3 patterns, gradual change

**PRESS**

| $\delta$ | Avg. FP (SD) | Avg. TP (SD) | Delay (SD) |
|---|---|---|---|
| 0.05 | $3.73\times10^{-3}$ ($1.6\times10^{-5}$) | 84.69 (0.54) | 58.56(0.38) |
| 0.10 | $3.71\times10^{-3}$ ($5.2\times10^{-5}$) | 87.5 (0.85) | 57.7(0.4) |
| 0.15 | $3.72\times10^{-3}$ ($2.5\times10^{-5}$) | 88.76 (0.88) | 57.1(0.41) |
| 0.20 | $3.76\times10^{-3}$ ($5.4\times10^{-5}$) | 89.63 (0.67) | 56.88(0.35) |
| 0.25 | $3.76\times10^{-3}$ ($3.6\times10^{-5}$) | 90.62 (0.91) | 56.26(0.5) |

**ProSeed**

| $\delta$ | Avg. FP (SD) | Avg. TP (SD) | Delay (SD) |
|---|---|---|---|
| 0.05 | $3.67\times10^{-3}$ ($1.6\times10^{-5}$) | 90.77 (0.42) | 56.63(0.43) |
| 0.10 | $3.90\times10^{-3}$ ($1.6\times10^{-5}$) | 93.24 (0.49) | 55.37(0.36) |
| 0.15 | $4.22\times10^{-3}$ ($2.2\times10^{-5}$) | 94.35 (0.25) | 54.5(0.32) |
| 0.20 | $4.59\times10^{-3}$ ($2.9\times10^{-5}$) | 95.21 (0.28) | 53.64(0.4) |
| 0.25 | $5.02\times10^{-3}$ ($2.8\times10^{-5}$) | 95.64 (0.3) | 53.08(0.53) |

**Seed**

| $\delta$ | Avg. FP (SD) | Avg. TP (SD) | Delay (SD) |
|---|---|---|---|
| 0.05 | $4.17\times10^{-3}$ ($5.14\times10^{-4}$) | 91.08 (0.6) | 56.43(0.41) |
| 0.10 | $4.50\times10^{-3}$ ($6.2\times10^{-4}$) | 93.44 (0.44) | 55.13(0.41) |
| 0.15 | $4.91\times10^{-3}$ ($7.13\times10^{-4}$) | 94.56 (0.31) | 54.33(0.34) |
| 0.20 | $5.3\times10^{-3}$ ($7.95\times10^{-4}$) | 95.26 (0.29) | 53.57(0.43) |
| 0.25 | $5.87\times10^{-3}$ ($8.72\times10^{-4}$) | 95.81 (0.28) | 52.85(0.46) |

**ADWIN**

| $\delta$ | Avg. FP (SD) | Avg. TP (SD) | Delay (SD) |
|---|---|---|---|
| 0.05 | $1.09\times10^{-2}$ ($1.08\times10^{-4}$) | 91.39 (0.39) | 56(0.32) |
| 0.10 | $1.07\times10^{-2}$ ($1.53\times10^{-4}$) | 93.52 (0.33) | 54.79(0.62) |
| 0.15 | $1.07\times10^{-2}$ ($7.1\times10^{-5}$) | 94.95 (0.20) | 53.45(0.25) |
| 0.20 | $1.10\times10^{-2}$ ($8.4\times10^{-5}$) | 95.67 (0.18) | 52.86(0.38) |
| 0.25 | $1.13\times10^{-2}$ ($1.2\times10^{-4}$) | 96.16 (0.22) | 52.36(0.53) |

Table 3: 5 patterns, gradual change

**PRESS**

| $\delta$ | Avg. FP (SD) | Avg. TP (SD) | Delay (SD) |
|---|---|---|---|
| 0.05 | $2.613\times10^{-3}$ ($2.7\times10^{-5}$) | 97.76 (0.57) | 50.08(0.62) |
| 0.10 | $2.729\times10^{-3}$ ($4\times10^{-5}$) | 98.54 (0.28) | 48.53(0.38) |
| 0.15 | $2.862\times10^{-3}$ ($4.3\times10^{-5}$) | 98.64 (0.38) | 47.84(0.46) |
| 0.20 | $2.951\times10^{-3}$ ($2.3\times10^{-5}$) | 98.96 (0.27) | 47.04(0.28) |
| 0.25 | $3.07\times10^{-3}$ ($3.9\times10^{-5}$) | 98.88 (0.22) | 46.69(0.38) |

**ProSeed**

| $\delta$ | Avg. FP (SD) | Avg. TP (SD) | Delay (SD) |
|---|---|---|---|
| 0.05 | $2.957\times10^{-3}$ ($2.8\times10^{-5}$) | 98.69 (0.34) | 46.94(0.55) |
| 0.10 | $3.344\times10^{-3}$ ($1.5\times10^{-5}$) | 99.02 (0.18) | 45.56(0.46) |
| 0.15 | $3.739\times10^{-3}$ ($2.5\times10^{-5}$) | 99.1 (0.11) | 44.86(0.39) |
| 0.20 | $4.162\times10^{-3}$ ($1.5\times10^{-5}$) | 99.32 (0.09) | 44.06(0.38) |
| 0.25 | $4.641\times10^{-3}$ ($3.1\times10^{-5}$) | 99.44 (0.14) | 43.47(0.51) |

**Seed**

| $\delta$ | Avg. FP (SD) | Avg. TP (SD) | Delay (SD) |
|---|---|---|---|
| 0.05 | $3.383\times10^{-3}$ ($4.37\times10^{-4}$) | 98.64 (0.31) | 46.82(0.59) |
| 0.10 | $3.842\times10^{-3}$ ($5.12\times10^{-4}$) | 99.05 (0.19) | 45.5(0.48) |
| 0.15 | $4.334\times10^{-3}$ ($6.11\times10^{-4}$) | 99.16 (0.15) | 44.81(0.35) |
| 0.20 | $4.825\times10^{-3}$ ($6.81\times10^{-4}$) | 99.36 (0.12) | 43.92(0.43) |
| 0.25 | $5.354\times10^{-3}$ ($7.32\times10^{-4}$) | 99.43 (0.13) | 43.34(0.49) |

**ADWIN**

| $\delta$ | Avg. FP (SD) | Avg. TP (SD) | Delay (SD) |
|---|---|---|---|
| 0.05 | $6.685\times10^{-3}$ ($7\times10^{-5}$) | 98.82 (0.16) | 45.94(0.34) |
| 0.10 | $6.881\times10^{-3}$ ($1.04\times10^{-4}$) | 99.01 (0.23) | 44.75(0.46) |
| 0.15 | $7.093\times10^{-3}$ ($1.02\times10^{-4}$) | 99.29 (0.20) | 43.57(0.43) |
| 0.20 | $7.453\times10^{-3}$ ($8.9\times10^{-5}$) | 99.32 (0.14) | 42.96(0.4) |
| 0.25 | $7.827\times10^{-3}$ ($1.04\times10^{-4}$) | 99.46 (0.15) | 42.3(0.52) |

Table 4: 10 patterns, gradual change

**PRESS**

| $\delta$ | Avg. FP (SD) | Avg. TP (SD) | Delay (SD) |
|---|---|---|---|
| 0.05 | $2.917\times10^{-3}$ ($4.5\times10^{-5}$) | 99.13 (0.2325) | 41.56(0.52) |
| 0.10 | $3.093\times10^{-3}$ ($5.8\times10^{-5}$) | 99.46 (0.2108) | 40.26(0.47) |
| 0.15 | $3.226\times10^{-3}$ ($4.7\times10^{-5}$) | 99.55 (0.1013) | 39.68(0.4) |
| 0.20 | $3.316\times10^{-3}$ ($4.5\times10^{-5}$) | 99.69 (0.0839) | 39.01(0.5) |
| 0.25 | $3.44\times10^{-3}$ ($6.8\times10^{-5}$) | 99.58 (0.1767) | 38.83(0.27) |

**ProSeed**

| $\delta$ | Avg. FP (SD) | Avg. TP (SD) | Delay (SD) |
|---|---|---|---|
| 0.05 | $3.331\times10^{-3}$ ($3.1\times10^{-5}$) | 99.43 (0.1563) | 39.31(0.45) |
| 0.10 | $3.695\times10^{-3}$ ($3.9\times10^{-5}$) | 99.67 (0.111) | 37.87(0.5) |
| 0.15 | $4.101\times10^{-3}$ ($3.7\times10^{-5}$) | 99.71 (0.1142) | 37.49(0.29) |
| 0.20 | $4.514\times10^{-3}$ ($3.4\times10^{-5}$) | 99.72 (0.0644) | 36.92(0.41) |
| 0.25 | $4.98\times10^{-3}$ ($5.5\times10^{-5}$) | 99.79 (0.0753) | 36.56(0.46) |

**Seed**

| $\delta$ | Avg. FP (SD) | Avg. TP (SD) | Delay (SD) |
|---|---|---|---|
| 0.05 | $3.793\times10^{-3}$ ($4.76\times10^{-4}$) | 99.42 (0.1594) | 39.18(0.54) |
| 0.10 | $4.209\times10^{-3}$ ($5.29\times10^{-4}$) | 99.65 (0.0919) | 37.94(0.48) |
| 0.15 | $4.67\times10^{-3}$ ($5.85\times10^{-4}$) | 99.7 (0.0961) | 37.5(0.34) |
| 0.20 | $5.143\times10^{-3}$ ($6.47\times10^{-4}$) | 99.75 (0.0682) | 36.85(0.43) |
| 0.25 | $5.64\times10^{-3}$ ($6.79\times10^{-4}$) | 99.79 (0.0947) | 36.44(0.57) |

**ADWIN**

| $\delta$ | Avg. FP (SD) | Avg. TP (SD) | Delay (SD) |
|---|---|---|---|
| 0.05 | $6.21\times10^{-3}$ ($6.4\times10^{-5}$) | 99.49 (0.1018) | 37.66(0.25) |
| 0.10 | $6.422\times10^{-3}$ ($1.06\times10^{-4}$) | 99.57 (0.1689) | 37.02(0.42) |
| 0.15 | $6.651\times10^{-3}$ ($1.21\times10^{-4}$) | 99.75 (0.1128) | 36.03(0.48) |
| 0.20 | $7.031\times10^{-3}$ ($1.08\times10^{-4}$) | 99.74 (0.0652) | 35.68(0.41) |
| 0.25 | $7.42\times10^{-3}$ ($1.2\times10^{-4}$) | 99.78 (0.1062) | 35.16(0.49) |

100%) for all algorithms, so we conclude that PRESS has a similar TP to others. In addition, all four algorithms have similar detection delays. For gradual drifts, we show results for 3, 5, and 10 patterns (Tables 2 to 4), we observed that all detectors have a similar detection delay.

In these experiments, we ran 1 million training instances to learn the network to train the the PRESS network. Then we switch PRESS to the testing phase, we ran it on the subsequent incoming 1 million instances generated by using the same random seed. In the real-world application, deciding when to switch the phase depends on the data stream, we suggest to drop the old network and re-initiate the training phase when the detector's performances drop, for example, when the overall false positive rates starts to increase. Alertnatively, a weighted mechanisme for expiring older informa-tion in the network can be used, this allows the information in the network to be up to date.

Table 5: False Postive Results on 3, 5, 10 Patterns

| | $\delta = 0.05$, abrupt changes | | |
|---|---|---|---|
| Patterns | 3 patterns | 5 patterns | 10 patterns |
| PRESS | $4.6\times10^{-5}$ ($1.9\times10^{-5}$) | $1.1\times10^{-4}$ ($1.6\times10^{-5}$) | $1.28\times10^{-4}$ ($8\times10^{-6}$) |
| ProSeed | $8.3\times10^{-5}$ ($1.5\times10^{-5}$) | $1.2\times10^{-4}$ ($5\times10^{-6}$) | $1.25\times10^{-4}$ ($7\times10^{-6}$) |
| Seed | $5.6\times10^{-4}$ ($4.6\times10^{-4}$) | $5.2\times10^{-4}$ ($4.1\times10^{-4}$) | $5.15\times10^{-4}$ ($4\times10^{-4}$) |
| ADWIN | $1.24\times10^{-3}$ ($2.6\times10^{-5}$) | $6.85\times10^{-4}$ ($8\times10^{-6}$) | $7.27\times10^{-4}$ ($8\times10^{-6}$) |
| | $\delta = 0.25$, abrupt changes | | |
| Patterns | 3 patterns | 5 patterns | 10 patterns |
| PRESS | $2.98\times10^{-4}$ ($5.5\times10^{-5}$) | $3.13\times10^{-4}$ ($2.7\times10^{-5}$) | $2.84\times10^{-4}$ ($1.7\times10^{-5}$) |
| ProSeed | $7.12\times10^{-4}$ ($2.1\times10^{-5}$) | $6.53\times10^{-4}$ ($2.3\times10^{-5}$) | $4.57\times10^{-4}$ ($2.5\times10^{-5}$) |
| Seed | $1.55\times10^{-3}$ ($8.57\times10^{-4}$) | $1.42\times10^{-3}$ ($7.86\times10^{-4}$) | $1.08\times10^{-3}$ ($6.41\times10^{-4}$) |
| ADWIN | $2.59\times10^{-3}$ ($2.4\times10^{-5}$) | $2.3\times10^{-3}$ ($2.3\times10^{-5}$) | $2.26\times10^{-3}$ ($1.8\times10^{-5}$) |
| | $\delta = 0.05$, gradual changes | | |
| Patterns | 3 patterns | 5 patterns | 10 patterns |
| PRESS | $3.73\times10^{-3}$ ($1.6\times10^{-5}$) | $2.61\times10^{-3}$ ($2.7\times10^{-5}$) | $2.92\times10^{-3}$ ($4.5\times10^{-5}$) |
| ProSeed | $3.67\times10^{-3}$ ($1.6\times10^{-5}$) | $2.95\times10^{-3}$ ($2.8\times10^{-5}$) | $3.33\times10^{-3}$ ($3.1\times10^{-5}$) |
| Seed | $4.17\times10^{-3}$ ($5.14\times10^{-4}$) | $3.38\times10^{-3}$ ($4.37\times10^{-4}$) | $3.79\times10^{-3}$ ($4.76\times10^{-4}$) |
| ADWIN | $1.09\times10^{-2}$ ($1.08\times10^{-4}$) | $6.69\times10^{-3}$ ($7\times10^{-5}$) | $6.21\times10^{-3}$ ($6.4\times10^{-5}$) |
| | $\delta = 0.25$, gradual changes | | |
| Patterns | 3 patterns | 5 patterns | 10 patterns |
| PRESS | $3.76\times10^{-3}$ ($3.6\times10^{-5}$) | $3.07\times10^{-3}$ ($3.9\times10^{-5}$) | $3.44\times10^{-3}$ ($6.8\times10^{-5}$) |
| ProSeed | $5.02\times10^{-3}$ ($2.8\times10^{-5}$) | $4.64\times10^{-3}$ ($3.1\times10^{-5}$) | $4.98\times10^{-3}$ ($5.5\times10^{-5}$) |
| Seed | $5.87\times10^{-3}$ ($8.72\times10^{-4}$) | $5.35\times10^{-3}$ ($7.32\times10^{-4}$) | $5.64\times10^{-3}$ ($6.79\times10^{-4}$) |
| ADWIN | $1.13\times10^{-2}$ ($1.2\times10^{-4}$) | $7.83\times10^{-3}$ ($1.04\times10^{-4}$) | $7.42\times10^{-3}$ ($1.2\times10^{-4}$) |

### 6.3 Evaluating False Positive Rate (FP)

In the experiments, we show that PRESS reduces the false positive. We compare FP for four detectors in both abrupt and gradual changing different streams with 3, 5, and 10 patterns. We use two confidence levels for each detectors as 0.05 and 0.25. FP is calculated as a percentage out of falsely detected concept drifts out of number of instances in the stream (in our case, the stream has 1,000,000 instances)

The results in Table 5 shows the false positive rate of the detector. The number in the () represents the standard deviation of the results. The performance of PRESS is promising as it has lower FP than Seed and ADWIN in both abrupt and gradual changing streams with 3, 5, and 10 patterns. In addition, PRESS has lower FP than ProSeed in any testing data stream when $\delta$ is greater than 0.10. These differences are statistically significant test by unpaired T-Test. We can conclude that PRESS achieves our research objectives: to reduce FP in data streams satisfying our assumptions.

### 6.4 Influences of $\beta$

In this section we show the effect of different values for parameter $\beta$. Table 6 shows FP, TP and Delay of PRESS on 10 patterns abrupt change streams. In abrupt changing streams, experiments show that an increasing FP will result in decreasing FP. However, it also comes with an increase in delay. Thus, $\beta$ controls the trade-off between FP and Delay in PRESS. The TP does not vary significantly with changing $\beta$.

We observe similar trends in gradually changing streams, except that TP has an anti-correlation with $\beta$. It is because a larger $\beta$ may result in overly augmented thresholds such that some true drifts cannot be detected. Choosing values for $\beta$ depends on user scenario. If one considers that a low false positive rate is important, one should use a higher $\beta$ value. However, a decreasing true positive rate may occur due to the trade-off. Whereas if the data stream is monitoring a critical task and TP is necessary then a lower $\beta$ value should be used.

Table 6: Effects of different $\beta$ values

| Abrupt Changes | | | |
|---|---|---|---|
| $\beta$ | Avg. FP (SD) | Avg. TP (SD) | Delay (SD) |
| 0.10 | $4.63\times10^{-4}$ ($2.8\times10^{-5}$) | 99.98 (0.006) | 24.14(0.13) |
| 0.20 | $3.56\times10^{-4}$ ($4.2\times10^{-5}$) | 99.99 (0.0061) | 24.61(0.17) |
| 0.30 | $3.15\times10^{-4}$ ($5.6\times10^{-5}$) | 99.99 (0.0051) | 24.92(0.2) |
| 0.40 | $2.98\times10^{-4}$ ($5.5\times10^{-5}$) | 99.99 (0.0042) | 25.26(0.26) |
| 0.50 | $2.74\times10^{-4}$ ($6.1\times10^{-5}$) | 99.99 (0.0066) | 25.89(0.2) |
| 0.60 | $2.55\times10^{-4}$ ($4.5\times10^{-5}$) | 100.00 (0.0056) | 26.22(0.16) |
| Gradual Changes | | | |
| $\beta$ | Avg. FP (SD) | Avg. TP (SD) | Delay (SD) |
| 0.10 | $4.273\times10^{-3}$ ($3.5\times10^{-5}$) | 99.77 (0.1) | 37.09(0.41) |
| 0.20 | $3.853\times10^{-3}$ ($5.4\times10^{-5}$) | 99.76 (0.06) | 37.78(0.37) |
| 0.30 | $3.607\times10^{-3}$ ($6.1\times10^{-5}$) | 99.75 (0.12) | 38.22(0.45) |
| 0.40 | $3.44\times10^{-3}$ ($6.8\times10^{-5}$) | 99.58 (0.18) | 38.83(0.27) |
| 0.50 | $3.314\times10^{-3}$ ($7.7\times10^{-5}$) | 99.42 (0.12) | 38.25(0.35) |
| 0.60 | $3.21\times10^{-3}$ ($5.3\times10^{-5}$) | 99.24 (0.1) | 38.88(0.25) |

### 6.5 Real-World Data

We evaluated the four drift detectors on MOA real-world data streams: Electricity, Forest Covertype and Poker Hand datasets. We use Naïve Bayes Classifier to generate error streams for three datasets. Error streams are preprocessed. A classification error normally produces a value of "1" and a correct classification produces a value of '0'. The error streams were pre-processed, and transformed in to an aggregated error stream based on a specific window size. The window size was set to 100. This meant that every 100 instances of the error stream was aggregated and transformed into an aggregated value. Finally, we obtained three integer streams to test four detectors. These datasets do not contain the actual positions of concept drifts so we only show the number of detected concept drifts in Table 7. PRESS produces the least number of drifts in most confidence levels. The reason we choose Naïve Bayes classifier is that it has been widely used in testing concept drift detectors. It has a reasonable prediction accuracy to test the effectiveness of most concept drift detectors. The real world data used the online training mode.

Table 7: Number of drifts detected in Real-World datasets

| Forest Covertype | | | | | |
|---|---|---|---|---|---|
| $\delta$ | 0.05 | 0.1 | 0.15 | 0.2 | 0.25 |
| PRESS | 864 | 864 | 864 | 864 | 864 |
| ProSeed | 843 | 889 | 922 | 957 | 969 |
| Seed | 824 | 872 | 932 | 961 | 989 |
| ADWIN | 1428 | 1472 | 1516 | 1561 | 1573 |
| Electricity | | | | | |
| $\delta$ | 0.05 | 0.1 | 0.15 | 0.2 | 0.25 |
| PRESS | 18 | 18 | 18 | 18 | 18 |
| ProSeed | 18 | 20 | 26 | 27 | 29 |
| Seed | 68 | 70 | 72 | 77 | 79 |
| ADWIN | 113 | 116 | 122 | 122 | 122 |
| Poker-Hand | | | | | |
| $\delta$ | 0.05 | 0.1 | 0.15 | 0.2 | 0.25 |
| PRESS | 1290 | 1290 | 1290 | 1290 | 1290 |
| ProSeed | 1382 | 1458 | 1502 | 1540 | 1575 |
| Seed | 1402 | 1478 | 1546 | 1573 | 1607 |
| ADWIN | 2227 | 2291 | 2322 | 2356 | 2386 |

## 7 Conclusion and Future Work

We investigated how concept drift severity will affect the performance of drift detectors. Based on the findings, we proposed the PRESS detector. PRESS uses a probability network and reservoir sampling to capture volatility and severity trends of a stream. Then, it can predict the severity of future drifts to adjust its detection sensitivity to reduce false positive rates. We compare PRESS with three other drift detectors in both synthetic and real-world datasets. Most experiments showed that PRESS generates the lowest false positive rate while maintaining similar true positive rates and delays to other drift detectors. We proposed a new method for severity detection for both Seed and ADWIN drift detectors and we applied this method in PRESS.

Our current limitation is that PRESS experiences slightly lower true positive rates with gradual concept drifts. In our future work, we want to make improvements on our severity detection method so that it can also work well in the case of gradual concept drift. In addition, we hope to further explore potential ways that drift severity may improve drift detection techniques besides adapting the drift detection threshold.

# Predicting Risk Level of Executables: an Application of Online Learning

**Huynh Ngoc Anh, Wee Keong Ng**
Nanyang Technological University, Singapore
hu0001nh@e.ntu.edu.sg, wkn@pmail.ntu.edu.sg

**Kanishka Ariyapala**
University of Padua, Italy
kanishka.ariyapala@math.uni.it

## Abstract

Nowadays, the number of new malware samples discovered every day is in millions, which undermines the effectiveness of the traditional signature-based approach towards malware detection. To address this problem, machine learning methods have become an attractive and almost imperative solution. In most of the previous work, the application of machine learning to this problem is batch learning. Due to its fixed setting during the learning phase, batch learning often results in low detection accuracy when encountered zero-day samples with obfuscated appearance or unseen behavior. Therefore, in this paper, we propose a new online algorithm, called FTRL-DP, to address the problem of malware detection under concept drift when the behavior of malware changes over time. We argue that online learning is more appropriate in this case than batch learning since online algorithms are inherently resilient to the shift in the underlying distribution. The experimental results show that online learning outperforms batch learning in all settings, either with or without retrainings.

## 1 Introduction

VirusTotal.com is an online service which analyzes files and urls for malicious content such as virus, worm and trojan by leveraging on an array of 52 commercial antivirus solutions for the detection of malicious signatures. On record, VirusTotal receives and analyzes nearly 2 million files every day. However, only a fraction of this amount (15%) can be identified as malicious by at least one antivirus solution. Given the fact that it is fairly easy nowadays to obfuscate a malware executable [You and Yim, 2010], it is rather reasonable to believe that a sheer number of the unknown files are actually obfuscated malware samples. In principle, the rest of the unknown cases should be manually reverse engineered to invent new signatures, but this is infeasible due to the large number of files to be analyzed. Therefore, looking for an automated way to address this problem is imperative and has attracted a lot of research effort, especially in the direction of using machine learning which has gained a lot of successes in various domains of pattern recognition such as face analysis [Valenti *et al.*, 2008] and sentiment analysis [Feldman, 2013].

In a recent paper, Saxe et al. [Saxe and Berlin, 2015] train a 3-layer neural network to distinguish between malicious and benign executables. In the first experiment, the author randomly splits the whole malware collection into train set and test set. The trained network can achieve a relatively high detection accuracy of 95.2% at 0.1% false positive rate. It is noted that the first experiment disregards the release time of the executables, which is an important dimension due to the adversarial nature of malware detection practice since malware authors are known to regularly change their tactics in order to stay ahead of the game [Iliopoulos *et al.*, 2011]. In the second experiment, the author uses a timestamp to divide the whole malware collection into train set and test set. The results obtained show that the detection accuracy drops to 67.7% at the same false positive rate. We hypothesize that the reasons for this result are two-fold: the change in behavior of malware over time and the poor adaptation of neural network trained under batch mode to the behavioral changes of malware. In addition, another recent study also reports similar findings when the release time of the executables is taken into account [Bekerman *et al.*, 2015].

The working mechanism of batch learning is the assumption that, the samples are independently and identically drawn from the same distribution (iid assumption). This assumption may be true in domains such as face recognition and sentiment analysis where the underlying concept of interest hardly changes over time. However, in various domains of computer security such as spam detection and malware detection, this assumption may not hold [Allix *et al.*, 2015] due to the inherently adversarial nature of cyber attackers, who may constantly change their strategy so as to maximize the gains. To address this problem of concept drift, we believe online learning is a more appropriate solution than batch learning. The reason is that, online algorithms are derived from a theoretical framework [McMahan, 2015] which does not impose the iid assumption on the data, and hence can work well under concept drift or adversarial settings [GAMA, 2013]. Motivated by this knowledge, we propose the Follow-the-Regularized-Leader with Decaying Proximal (FTRL-DP) algorithm – a variant of the proximal Follow-the-Regularized-Leader (FTRL-Proximal) algorithm [McMahan *et al.*, 2013] – to address the problem of malware detection.

To be specific, the contributions of this paper are as follow:

- A new online algorithm (FTRL-DP) to address the problem of concept drift in Windows malware detection. Our main claim is that online learning is superior to batch learning in the task of malware detection. This claim is substantiated in Section 7 by analyzing the accuracy as well as the running time of FTRL-DP, FTRL-Proximal and Logistic Regression (LR). The choices of the algorithms are clarified in Section 4.

- An extensive data collection of more than 100k malware samples using the state-of-the-art open source malware analyzer, Cuckoo sandbox [Guarnieri *et al.*, 2012], for the evaluation. The collected data comprises many types such as system calls, file operations, and others, from which we were able to extract 482 features. The experiment setup for data collection and the feature extraction process are described in Section 5.

For LR, the samples in one month constitutes the test set and the samples in a number of preceding months are used to form the train set; for FTRL-DP, a sample is used for training right after it is tested. The detailed procedure is presented in Section 6. In Section 2, we review the previous works related to the problem of malware detection using machine learning. We formulate the problem of malware detection as a regression problem in Section 3. Lastly, in Section 8, we conclude the paper and discuss the future work.

## 2 Related Work

### 2.1 Batch Learning in Malware Detection

Traditional antivirus solutions often rely on a signature database to perform detection. The weakness of this design is that the signatures may become out-dated resulting in the failure to detect new malware samples [Adesegun Oreoluwa, 2015]. Inspite of this weakness, they are still widely in use today because they are the only practical approach that generates a manageable amount of false positive rate. Therefore, the idea of a completely automated detection solution that can detect novel samples by modeling the behavior of previous malware samples has been intensively researched. The modeling techniques previously used vary from graph analytics [Anderson *et al.*, 2011], Bayesian network [Zhang *et al.*, 2016] to machine learning methods [Saxe and Berlin, 2015]. Specifically, in the machine learning approach, previous works in the literature are most notably different in the set of features used to perform classification.

The analysis of suspicious executables for extracting the features used for automated classification can be broadly divided into two types: static analysis and dynamic analysis. In static analysis, the executables are not required to be executed and only features extracted directly from the executables are used for classification such as file size, readable strings, binary images, n-gram words, etc. [Makandar and Patrot, 2015; Saini *et al.*, 2014].

On the other hand, dynamic analysis requires the execution of the executables to collect generated artifacts for feature extraction. Dynamic features can be extracted from host-based artifacts such as system call traces, dropped files and modified registries [Norouzi *et al.*, 2016]. Dynamic features can also be extracted from network traffic such as the frequency of TCP packets or UDP packets [Nari and Ghorbani, 2013; Rafique *et al.*, 2014]. While static analysis is highly vulnerable to obfuscation attacks, dynamic analysis is more robust to binary obscuration techniques.

To improve detection accuracy, the consideration of a variety of dynamic features of different types is recently gaining attention due to the emergence of highly effective automated malware analysis sandboxes. In their work, Mohaisen et al. [Mohaisen *et al.*, 2015] studied the classification of malware samples into malware families by leveraging on a wide set of features (network, file system, registry, etc.) provided by the proprietary AutoMal sandbox. In a similar spirit, Korkmaz et al. [Korkmaz, 2015] used the open-source Cuckoo sandbox to obtain a bigger set of features aiming to classify between traditional malware and non-traditional (Advanced Persistent Threat) malware. The general conclusion in these papers is that combining dynamic features of different types tends to improve the detection accuracy.

In light of these considerations, we decided to use the Cuchoo sandbox to execute and extract the behavioral data generated by a collection of more than 100k suspicious executables. Our feature set (Section 5.3) is similar to that of Korkmaz et al. although we address a different problem: regressing the risk levels of the executables. Furthermore, our work differs from previous work in the aspect that we additionally approach this problem as an online learning problem rather than just a batch learning one.

### 2.2 Online Learning in Malware Detection

Although previous studies [Saxe and Berlin, 2015; Mohaisen *et al.*, 2015; Rafique *et al.*, 2014] on the application of machine learning to the problem of malware detection have been reporting promising high accuracy, these results are not directly relevant to the real practice of malware detection. The reason for this claim is that, temporal information of the executables is often ignored in most of these works, resulting in the fact that the algorithms are only able to recognize unknown executables, but not new ones [Saxe and Berlin, 2015]. However, in practice, we are more interested in detecting new malware samples, not just unknown ones. We argue that the right way to evaluate a machine learning algorithm in this practice of malware detection is to construct the train set and the test set by splitting the dataset using some dividing time stamp. The reason for this approach is the evolutionary behavior of malware samples, which is rather different from other disciplines such as object detection or sentiment analysis, in which the objects to be detected are effectively fixed.

The importance of malware's release time has been extensively studied in the domain of Android malware detection. In their paper [Allix *et al.*, 2015], Allix et al. studied the effect of history on the biased results of existing works on the application of machine learning to the problem of malware detection. The author notes that most existing works evaluate their methodology by randomly picking the samples for the train set and the test set. The conclusion is that, this procedure usually leads to much higher accuracy than the cases when the train set and the test set are historically coherent. The author argues that this result is misleading as it is not useful for a detection approach to be able to identify randomly picked samples but fail to identify zero-day or new ones.

To address this problem of history relevance, Narayanan et al. [Narayanan *et al.*, 2016] have developed an online detection system, called DroidOL, which is based on the online Passive Aggressive (PA) algorithm. The novelty of this work is the use of an online algorithm with the ability to adapt to the change in behavior of malware in order to improve detection accuracy. The obtained result shows that, the online PA algorithm results in a much higher accuracy (20%) compared to the typical setting of batch learning and 3% improvement in the settings when the batch model is frequently retrained.

# 3 Problem Statement

Given 52 antivirus solutions, we address the problem of predicting the percentage of solutions that would flag an executable file as malicious. Formally, this is a regression problem of predicting the output $y \in [0, 1]$ based on the input $x \in R^n$ which is the set of 482 hand-crafted features extracted from the reports provided by the Cuckoo sandbox (Section 5.3). The semantic of the output defined in this way can be thought of as the risk level of an executable. We augment the input with a constant feature which always has the value 1 to simulate the effect of a bias. In total, we have 483 features for each malware sample.

In this case, we have framed the problem of malware detection as the regression problem of predicting the risk level of an executable. We rely on the labels provided by all 52 antivirus solutions and do not follow the labels provided by any single one as different antivirus solutions are known to report inconsistent labels [Kantchelian *et al.*, 2015]. In addition, we also do not use two diferent thresholds to separate the executables into two classes, malicious and benign, as in [Saxe and Berlin, 2015] since it would discard the hard cases where it is difficult to determine the nature of the executables, which may be of high value in practice.

# 4 Methodology

The basic idea of the application of machine learning to malware detection is the learning of recurrent patterns in malware behavior exhibited in the past so as to make predictions about the samples at present. This methodology is usually referred to as supervised learning, in which the purpose is to learn a function which can correctly map an input $x$ to an output $y$. We are particularly interested in two ways that this mapping function can be learned: batch learning and online learning. In batch learning, the mapping function is constructed once using the bulk of training data, which is assumed to be available all at once. On the other hand, in online learning, the mapping function is updated frequently based on the most recent training samples.

To allow a fair comparison between batch learning and online learning, we use the models of the same linear form, represented by a weight vector $w$, in both cases. The sigmoid function ($\frac{1}{1+e^{-z}}$) is then used to map the dot product (biased by the introduction of a constant feature) between the weight vector and the input, $w^T x$, to the [0,1] interval of possible risk levels. Additionally, in both cases, we optimize the same objective function, which is the sum of logistic loss (log loss – the summation term in Equation 1). In the batch learning setting, the sum of log loss is optimized in a batch manner in which each training example is visited multiple times in minimizing the objective function. The resultant algorithm is usually referred to as Logistic Regression with log loss (Section 4.1). On the other hand, in the online setting, we optimize the sum of log loss in an online manner, in which each training sample is only seen once. The resultant algorithm is the proposed FTRL-DP algorithm (Section 4.2).

## 4.1 Batch Learning – LR

Given a set of $n$ training examples $\{(x_i, y_i)\}_{i=1}^n$, Logistic Regression with log loss corresponds to the following optimization problem:

$$\operatorname*{argmin}_{w} \left\{ -\sum_{i=1}^{n} \left( y_i \log(p_i) + (1-y_i) \log(1-p_i) \right) + \lambda_1 \|w\|_1 + \frac{1}{2}\lambda_2 \|w\|_2^2 \right\} \text{ in which } p_i = \text{sigmoid}(w^T x_i) \quad (1)$$

The objective function of Logistic Regression (Equation 1) is a convex function with respect to $w$ as it is the sum of three convex terms. The first term is the sum of log losses associated with all training samples (within a time window). The last two terms are the L1–norm regularizer and the L2–norm regularizer. The L1 regularizer is a non–smoothed function used to introduce sparsity into the solution weight $w$. On the other hand, the L2 regularizer is a smooth function used to favor low variance models that have small weight.

The fundamental assumption of batch learning is that, the distribution of the train set must be similar to the distribution of the test set (iid assumption). This is to ensure that the weight learned from the train set is applicable to the test set. As a consequence, this method is quite successful in stationary domains such as object recognition or sentiment analysis, but may fail in malware detection due to the change in behavior of malware over time [Allix *et al.*, 2015]. This is the main motivation of this paper to explore the framework of adaptive online learning [McMahan, 2015] for deriving better algorithms to address this problem of concept drift.

## 4.2 Online Learning – FTRL-DP

**Online Convex Optimization.**
The general framework of online convex optimization can be formulated as follows [Shalev-Shwartz, 2011]. We need to design an algorithm that can make a series of optimal predictions, each at one time step. At time step $t$, the algorithm makes a prediction, which is a weight vector $w_t$. A convex loss function $l_t(w)$ is then exposed to the algorithm after the prediction. Finally, the algorithm suffers a loss of $l_t(w_t)$ at the end of time step $t$ (Algorithm 1). The algorithm should be able to learn from the losses in the past so as to make better and better decisions over time.

The objective of online convex optimization is to minimize the regret with respect to the best classifier in hindsight

**Algorithm 1** Online Algorithm
1: **for** t = 1,2,... **do**
2:     Make a prediction $w_t$
3:     Receive the lost function $l_t(w)$
4:     Suffer the lost $l_t(w_t)$

(Equation 2). The meaning of Equation 2 is that we would like to minimize the total loss incurred up to time $t$ with respect to the supposed loss incurred by the best possible prediction in hindsight, $w^*$.

$$\textbf{Regret}_t = \sum_{i=1}^{t} l_i(w_t) - \sum_{i=1}^{t} l_i(w^*) \quad (2)$$

Since the future loss functions are unknown, the best guess or the greedy approach to achieve the objective of minimizing the regret is to use the prediction that incurs the least total loss on all past rounds. This approach is called Follow-the-Leader (FTL), in which the leader is the best prediction that incurs the least total loss with respect to all the past loss functions. In some cases, this simple formulation may result in algorithms with undesirable properties such as rapid change in the prediction [Shalev-Shwartz, 2011], which lead to overall high regret. To fix this problem, some regularization function is usually added to regularize the prediction. The second approach is called Follow-the-Regularized-Leader (FTRL), which is formalized in Equation 3.

$$w_{t+1} = \underset{w}{\mathrm{argmin}} \left\{ \sum_{i=1}^{t} l_i(w) + r(w) \right\} \quad (3)$$

It is notable to see that the FTRL framework is formulated in a rather general sense and performs learning without relying on the iid assumption. This property makes it more suitable to adversarial settings or settings in which the concept drift problem is present.

**The Proposed FTRL-DP Algorithm.**
In the context of FTRL-DP, an online classification or an online regression problem can be cast as an online convex optimization problem as follows. At time $t$, the algorithm receives input $x_t$ and makes prediction $w_t$. The true value $y_t$ is then revealed to the algorithm after the prediction. The loss function $l_t(w)$ associated with time $t$ is defined in terms of $x_t$ and $y_t$ (Equation 4). Finally, the cost incurred at the end of time $t$ is $l_t(w_t)$. The underlying optimization problem of FTRL-DP is shown in Equation 5.

$$l_t(w) = -y_t \log(p) - (1 - y_t) \log(1 - p)$$
$$\text{in which } p = \text{sigmoid}(w^T x_t) \quad (4)$$

Compared with Equation 3, Equation 5 has the actual loss function $l_t(w)$ replaced by its linear approximation at $w_t$, which is $l_t(w_t) + \nabla l_t(w_t)^T (w - w_t) = g_t^T w + l_t(w_t) - g_t^T w_t$ (in which $g_t = \nabla l_t$). The constant term $\left(l_t(w_t) - g_t^T w_t\right)$ is omitted in the final equation without affecting the optimization problem. This approximation is to allow the derivation of a closed–form solution to the optimization problem at each time step, which is not possible with the original problem in Equation 3.

$$w_{t+1} = \underset{w}{\mathrm{argmin}} \left\{ g_{1:t}^T w + \lambda_1 \|w\|_1 + \frac{1}{2}\lambda_2 \|w\|_2^2 + \frac{1}{2}\lambda_3 \sum_{s=1}^{t} \sigma_{t,s} \|w - w_s\|_2^2 \right\} \text{ in which } g_{1:t}^T = \sum_{i=1}^{t} g_t^T \quad (5)$$

FTRL-DP utilizes 3 different regularizers to serve 3 different purposes. The first two regularizers of L1–norm and L2–norm serve the same purpose as in the case of Logistic Regression introduced in Section 4.1. The third regularization function is the proximal term used to ensure that the current solution does not deviate too much from past solutions with more influence given to most recent ones by using an exponential decaying function $\left(\sigma_{t,s} = \exp(-\gamma|t - s|)\right)$ with $\gamma > 0$. This is our main difference from the original FTRL-Proximal algorithm [McMahan *et al.*, 2013] and its recently proposed variant [Tan *et al.*, 2016]. The replacement of the per coordinate learning rate schedule by the decaying function proves to improve the prediction accuracy in the face of concept drift (discussed in Section 7).

The optimization problem in Equation 5 can be efficiently solved in closed form (Equation 6). The derivation is omitted here due to space constraint. The implementation is available at https://git.io/vDyra.

$$w_{t+1,i} = \begin{cases} 0 & \text{if } |z_{t,i}| \leq \lambda_1 \\ -\frac{z_{t,i} - \lambda_1 \text{sign}(z_{t,i})}{\lambda_2 + \lambda_3 \frac{\exp(-\gamma t) - 1}{\exp(-\gamma) - 1}} & \text{otherwise.} \end{cases} \quad (6)$$

$$\text{in which } z_t = g_{1:t} - \lambda_3 \sum_{s=1}^{t} \sigma_{t,s} w_s$$

In summary, we aim to compare between the performance of batch learning and online learning on the problem of malware detection. To make all things equal, we use the models of the same linear form and optimize the same log loss function, which lead to the LR algorithm in the batch learning case and the FTRL-DP algorithm in the online learning case. For LR, only the samples within a certain time window contribute to the objective function (Equation 1). On the other hand, the losses associated with all previous samples equally contribute to the objective function of FTRL-DP (Equation 5). This difference is critical as it leads to the gains in the performance of FTRL-DP over LR, which is discussed in Section 7.

## 5 Data Collection

### 5.1 Malware Collection

We used more than 1 million files collected in the duration from March 2016 to Apr 2016 by VirusShare.com for the experiments. VirusShare is an online malware analyzing service that allows Internet users to scan arbitrary files against an array of 52 antivirus solutions (the scan results are actually provided by VirusTotal). In this study, we are only interested in executable files and able to separate out more than 100k executables from the 1 million files downloaded.

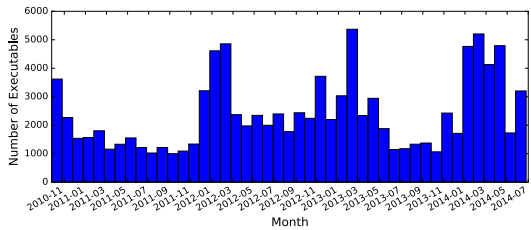

Figure 1: Distribution of Executables through Time.

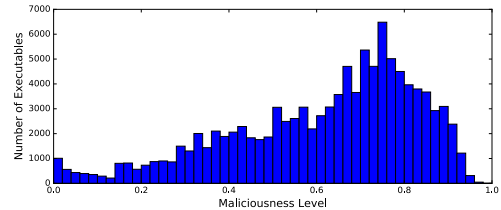

Figure 2: Distribution of Risk Level.

Figure 1 shows the distribution of the executables with respect to executables' compile time. The horizontal axis of Figure 1 shows the months during the 4 years from Nov/2010 until Jul/2014, which is the period of most concentration of executables and chosen for the study. The vertical axis of Figure 1 indicates the number of executables compiled during the corresponding month. Figure 2, instead, shows the maliciousness distribution of the executables. The horizontal axis indicates the maliciousness measure and the vertical axis the number of corresponding executables.

### 5.2 Malware Execution

Due to the large number of samples to be executed, we have to make use of some kind of the parallelism to cut down the time required to conduct the experiments. For this requirement, we use the Cuckoo sandbox, which is an open source malware analysis sandbox. Cuckoo sandbox supports both physical and virtual sandbox machines such as VirtualBox, VMWare and KVM. Since it is infeasible to execute this large number of samples on physical environments, we decided to execute them on an array of VirtualBox virtual machines.

We make use of the facility provided by DeterLab [Mirkovic and Benzel, 2013] as the testbed for the execution of the executables. DeterLab is a flexible online experimental lab for computer security, which provides researchers with a host of physical machines to carry out experiments. In our setup, we use 25 physical machines with each physical machine running 5 virtual machines for executing the executables. Each executable is allowed to run for 1 minute. The experiment ran for more than 20 days and collected the behavioral data of roughly 100k executables.

### 5.3 Feature Extraction

There are basically two different ways that we can extract the features that characterize an executable depending on whether the executable is executed or not. In the first method, the executable is not executed and features are extracted directly from the file. This method is called static analysis. The advantage of this approach is the relatively short amount of time that it requires to extract the features but its disadvantage being that it is vulnerable to obfuscation techniques [You and Yim, 2010].

On the other hand, dynamic analysis is usually robust to obfuscation techniques. In this method, the executables are actually executed in a protected environment to collect generated artifacts such as network traffic, dropped files, and created registries. Dynamic features can then be extracted from the generated artifacts for further study such as classification and regression. The downside of dynamic analysis in the tedious task of executing the malware and the timespan required to do so. However, this disadvantage can be effectively overcome by paralleling the experimental process, and analysis sandboxes such as Cuckoo supports the required features.

In this paper, we mostly consider dynamic features of the following 4 categories for regression: file system category, registry category, system call category, and the category of other miscellaneous features.

We do not focus on engineering novel features to increase the detection accuracy but perform the comparative analysis between batch learning and online learning. The features that we used were mostly influenced by the recent work of Korkmaz et al [Korkmaz, 2015] who also used Cuckoo sandbox as the framework for data collection. All these features can be easily derived from the reports by the Cuckoo sandbox and will be described in details as follows.

**API Call Category**

API (Application Programming Interface) calls are the functions provided by the operating system to grant application programs the access to basic functionality such as disk read and process control. Although these calls may ease the process of manipulating the resources of the machine, it also provides hackers with a lot of opportunities to obtain confidential information. For this category, we consider the invoking frequencies of the API calls as a set of features. In addition, we also extract as features the frequencies that the API files are linked. The total number of features in this category is 353 and the complete set of API calls as well as the set of API files are available at https://git.io/vDywd.

**Registry Category.**

In Windows environments, the registry is a hierarchal database that holds the global configuration of operating system. Ordinary programs often use the registry to store information such as program location and program settings. Therefore, the registry system is like a gold mine of information for malicious programs, which may refer to it for information such as the location of the local browsers or the version of the host operating system. Malicious program may also add keys to the registry so as to be able to survive multiple system restarts. We extract the following 4 registry related features: the number of registries being written, opened, read and deleted.

**File System Category.**
File system is the organization of the data that an operating system manages. It includes two basic components: file and directory. File system-related features are an important set of features to consider since malware has to deal with the file system in one way or another in order to cause harm to the system or to steal confidential information. We consider the following file-related features: the number of files being opened, written, in existence, moved, read, deleted, failed and copied. In addition, we also consider the following 3 directory related features: the number of directories being enumerated, created and removed. In total, we were able to extract 11 features in this category.

**Miscellaneous Category.**
In addition to out-of-the-box functionalities, Cuckoo sandbox is further enhanced by a collection of signatures contributed by the public community. These signatures can identify certain characteristics of the analyzed binary such as the execution delay time or the ability to detect virtual environment. All these characteristics are good indicators for the high risk level of an executable but may just be false positives. We consider the binary features of whether the community signatures are triggered or not. In addition, we also consider 3 other features that may be relevant to the behavior characterization: the number of mutex created, the number of processes started and the depth of the process tree. The total number of features being extracted in this category is 118.

In summary, we are able to extract 482 features that spans 4 different categories: API calls, registry system, file system and miscellaneous features.

## 6 Evaluation

### 6.1 Experiment with LR

We evaluate LR in four different settings: once, multi-once, monthly, and multi-monthly. In the once setting, the samples appeared in the first month of the whole dataset are used to form the train set and the rest of the samples are used to form the test set. The multi-once setting is similar to the once setting except that the samples in the first 6 months are used to form the train set instead. It should be noted that retraining is not involved in the first two settings.

On the other hand, the other two settings do involve retraining, which is a crude mechanism to address the change in behavior of malware over time. Since it is infeasible to carry out retraining upon the arrival of every new sample, we perform retraining on a monthly basis. Due to the characteristic of our dataset, we find that the monthly basis is a good balance to ensure that we have enough samples for the train set and the training time is not too long (the monthly average number of samples is 2.4k). In the monthly setting, we use the samples released in a month to form the test set and the samples released in the immediately preceding month to form the train set. The multi-monthly setting is similar to the monthly setting except that we use the samples in the preceding 6 months to form the train set instead.

For a quick evaluation, we make use of the LR implementation, provided by the TensorFlow library [Abadi *et al.*, 2016] to train and test the LR regressors. TensorFlow is a framework for training large scale neural network, but in our case, we only utilize a single layer network with sigmoid activation, binary cross-entropy loss and two regularizations of L1-norm and L2-norm. 20% of each train set is dedicated for validation and the maximum number of epochs that we use is 100. We stop the training early if the validation loss does not get improved in 3 consecutive epochs.

### 6.2 Experiment with FTRL algorithms

We use the standard procedure to evaluate FTRL-DP and FTRL-Proximal (jointly referred to as FTRL algorithms). Each new sample is tested on the current model giving rise to an error, which is then used to make modification to the current model right after. This evaluation is usually referred to as the mistake-bound model.

Due to their simplicity, FTRL-DP and FTRL-Proximal can be implemented in not more than 40 lines of python code. The implementation makes heavy use of the numpy library, which is mostly written in C++. As TensorFlow also has C++ code under the hood, we believe that the running time comparison between the two cases is sensible. Evaluated on the same computer, it actually turns out that the running time of FTRL algorithms is much lower than that of LR. We use the same amounts of three regularizations for both FTRL-DP and FTRL-Proximal. For FTRL-DP, we report the best possible setting for hyperparameter $\gamma$.

The computer used for all the experiments has 16GB RAM and operates with a 1.2GHz hexa-core CPU. The running times of all experiments are shown in Table 1. The mean cumulative absolute errors are reported in Figure 3.

## 7 Discussion

### 7.1 Prediction Accuracy

We use the mean cumulative absolute error (MCAE) to compare the performance between FTRL-DP, FTRL-Proximal and different batch settings of LR, which are reported in Figure 3. The MCAE is defined in Equation 7, in which $y_t$ is the actual risk level of an executable and $p_t$ the risk level predicted by the algorithms. In Figure 3, the horizontal line shows the cumulative number of samples and the vertical line the MCAE. There are four notable observations that we can see from Figure 3.

$$\frac{1}{n} \sum_{t=1}^{n} |y_t - p_t| \qquad (7)$$

Firstly, the more data that we train the LR model on, the better performance we can achieve. This observation is evidenced by the fact that, in most of the time, the error line of the multi-once setting stays below the error line of the once setting, and the error line of the multi-monthly setting stays below the error line of the monthly setting. A possible explanation for this observation is that the further we go back in time to obtain more data to train the model on, the less variance the model becomes, which results in the robustness to noise, and consequently, higher prediction accuracy.

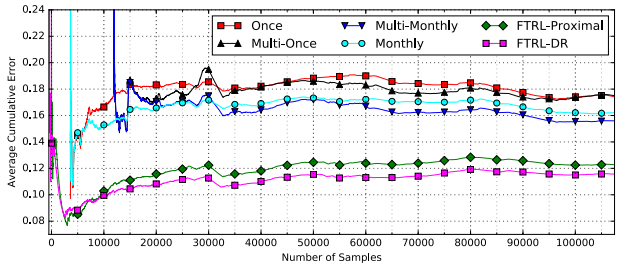

Figure 3: Mean cumulative absolute errors of of FTRL-DP, FTRL-Proximal and different settings of LR.

| Experiment | Running Time |
|---|---|
| LR Multi-monthly | 44m 31s |
| LR Monthly | 14m 14s |
| LR Multi-once | 55s |
| LR Once | 42s |
| FTRL-Proximal | 28s |
| FTRL-DP | 26s |

Table 1: Running time of FTRL-DP, FTRL-Proximal and different settings of LR.

Secondly, the retraining procedure does help to improve prediction accuracy. It is evidenced by the fact that the monthly setting outperforms the once setting, and similarly the multi-monthly setting outperforms the multi-once setting. This observation is a supporting evidence for the phenomenon of evolving malware behavior. As a consequence, the most recent samples would be more relevant to the current samples, and training on most recent samples would result in a more accurate prediction model.

From the first two observations, we can conclude that the further we go back in time to obtain more samples and the more recent the samples are, the better the trained model would perform. This conclusion can be exploited to improve prediction accuracy by going further and further back in time and retraining the model more often. However, this approach would become unpractical at some point when the training time required to frequently update an accurate model via periodic retrainings would become too long to be practical. It turns out that this issue can be elegantly addressed by the FTRL algorithms, which produces much higher prediction accuracy at considerable lower running time.

Thirdly, FTRL algorithms (worse MCAE of 0.123) are shown to outperform the LR algorithm in all settings (best MCAE of 0.156). The error lines corresponding to the performance of FTRL algorithms consistently stays below other error lines. The gain in the prediction accuracy of FTRL algorithms over all settings of LR can be explained by the contribution of all previous samples to its objective function. In different batch settings of LR, only the losses associated with the samples within a certain time window contribute to the respective objective functions.

Finally, the fourth observation is that FTRL-DP (MCAE of 0.116) outperforms FTRL-Proximal (MCAE of 0.123). The gain in performance of FTRL-DP over FTRL-Proximal can be explained by the ability of FTRL-DP to cope with concept drift via the use of a specially designed adaptive mechanism. This mechanism makes use of an exponential decaying function to favor the most recent solutions over older ones. The effective result is that the most recent samples would contribute more to the current solution thereby alleviating the problem of concept drift.

### 7.2 Running Time

In terms of running time (training time and testing time combined), FTRL-DP and FTRL-Proximal are clearly advantageous over LR. From Table 1, we can see that the running times of FTRL algorithms are much lower than that of LR, especially compared to the settings with retraining involved (monthly and multi-monthly). The reason for this result is that FTRL algorithms only needs to see each sample once to update the current weight vector whereas in the case of LR, it requires multiple passes over each sample to ensure convergence to the optimal solution.

## 8 Conclusions and Future Work

The evolving nature of malware over time makes the malware detection problem more difficult. According to previous studies, batch learning based methods often perform poorly when encountered zero-days samples. Our research is motivated to fill in this gap by proposing FTRL-DP – a variant of the FTRL-Proximal algorithm – to address this problem. We evaluated two learning paradigms using an extensive dataset generated by more than 100k malware samples executed on Cuckoo sandbox. The experimental results show that FTRL algorithms (worse MCAE of 0.123) outperforms LR in the typical setting of batch learning as well as the settings with retrainings involved (best MCAE of 0.156). The gain in performance of FTRL algorithms over different batch settings of LR can be accounted for by its objective function taking into account the contribution of all previous samples. Furthermore, the improvement of FTRL-DP over FTRL-Proximal can be explained by the usage of an adaptive mechanism that regularizes the weight by favoring recent samples over older ones. In addition, FTRL algorithms are also more advantageous in terms of running time.

It can be noticed that all above methods are black-box solutions, which do not gain domain experts any insights. An interesting development of this work is to enable the direct intereaction with a domain expert using a visualization. The domain expert could prioritize or discard weight alterations suggested by the learning algorithm via the interactive exploration of malware behavior. This visual analytics approach would lead to a transparent solution where the domain expert can benefit most of his knowledge in colaboration with black-box automated detection solutions.

# Co-training semi-supervised learning for single-target regression in data streams using Random AMRules

Ricardo Sousa
LIAAD/INESC TEC, University of Porto,
Portugal
rtsousa@inesctec.pt

João Gama
LIAAD/INESC TEC, University of Porto,
Portugal
Faculty of Economics, University of Porto,
Portugal
jgama@fep.up.pt

## Abstract

This paper addresses the development of semi-supervised and co-training method for single-target regression based on ensembles and rule learning. Systems based on data streaming often produce considerable number of unlabeled data due to label assignment impossibility, high cost of labeling or labeling long duration tasks.

Co-Training is a semi-supervised approach that was created to benefit the models and predictions accuracy through input information that is present in the unlabeled examples. This approach is essentially employed in classification and batch mode context.

Considering these facts, this work presents a co-training online method for single-target regression. This method explores the advantages of rule and ensembles learning in order to benefit from the input information for regression models improvement.

In the experimental evaluation, the performance of proposed algorithm was compared in a context where all unlabeled examples are not used (real standard scenario) and a context where unlabeled data is used in the model training so that the advantage of using unlabeled data could be verified.

The results reveal that the method utilization of unlabeled data can provide better performance than the standard rejection of unlabeled data. However, the improvements on the predictions still are small.

## 1 Introduction

Prediction has been a fundamental task in online data streams due to the fact that many areas (where data streams are the main source of data) rely on accurate predictions for decision making or planning [1; 2; 3]. In these areas context, large amount of data is not label assigned due to label assignment impossibility, high cost of label assignment or long time tasks. In fact, sensor malfunction or database failure easily produce vast amount of unlabeled data ( without output values). Moreover, sensitive data (e.g., privacy preservation ) often forces to the label omission [4; 5; 6].

Unlabeled data usually appear in many areas such as Engineering Systems ( video object detection ) [7], Physics (weather forecasting and ecological models) [8], Biology (model of cellular processes) [9] and Economy/Finance (stock price forecasting) [1]. In most of these areas, data from streams are obtained and processed in real time [4].

Semi-supervised Learning (SSL) is a methodology that takes advantages of the input information present on unlabeled data for accuracy improvement of the predictions [4; 10]. This methodology is only beneficial in a condition of significant unlabeled data abundance [11]. The drawbacks of these methodologies may lay on the introductions of errors by inaccurate artificial labels propagation [11; 12; 13].

More formally, $\mathbf{X} = \{X_1, ..., X_j, ..., X_J\} \in \mathbb{R}^J$ represents a vector of input random variables and $\mathbf{Y}$ represents a scalar random variable, with a joint probability distribution $P(\mathbf{X}, \mathbf{Y})$. The vectors $\mathbf{x}_i = (x_{i,1}, ..., x_{i,j}, ..., x_{i,J}) \in \mathbb{R}^J$ and $y_i$, where $i \in \{0, 1, 2, ...\}$, represent realizations of $\mathbf{X}$ and $\mathbf{Y}$, respectively. A stream is defined as the sequence of examples $\mathbf{e}_i = (\mathbf{x}_i, y_i)$ represented as $\mathcal{S} = \{(\mathbf{x}_0, y_0), (\mathbf{x}_1, y_1), ..., (\mathbf{x}_i, y_i), ...\}$. Label absence is represented by $y_i = \emptyset$. SSL methods have in view the use of examples $(\mathbf{x}_i, \emptyset)$ to enhance the regression model $y_i \leftarrow f(\mathbf{x}_i)$ and reduce the error of prediction for both labeled and unlabeled examples.

Batch mode is most abundant among SSL methods which are often applied to classification and can not be immediately adapted to regression [3; 11; 14].

Co-training approach explores the synergy of the two perspectives. This method is one prominent SSL approach which consists of creating two ore more different models for the same problem through different inputs, different regression methods or different parametrization [4]. In fact, the Co-training is a natural evolution of the Self-training method since the Self-training is a particular case of one regressor Co-training that teaches to itself [4].

The training stage produces an artificial label for the unlabeled example from the regressors predictions of the same example, according to a criterion (e.g., mean of all predictions) [11]. Posteriorly, this artificially labeled example is used in the training of the regressors. The prediction stage yield a final prediction from the regressors predictions of the example, according a similar criterion as in training.

The purpose of this paper is to present a Co-training method based on ensembles and rule leaning to online, single-target regression. This work may pave the way for the extension to online multi-target regression using the Random Adaptive Model Rules (Random AMRules) algorithm in future works [15; 16].

This document is structured as follows. In Section 2, the fundamentals of SSL and available methods of co-training are briefly revised. Section 3 describes the modifications of the Co-training to online learning and regression using ensembles of rule models. Section 4 describes the evaluation method. The results are discussed in Section 5 and the main conclusions are remarked in Section 6.

## 2 Related Work

The main fundamentals of Co-training and some available methods are briefly described in order to create the technical context. For the best of our knowledge, online versions of Co-training methods were not found in the literature. Moreover, the available batch mode methods are mainly applied to classification [11; 17]. However, these presented methods represent a reasonable initial platform for the development of a Co-training method for online single-target regression.

In a more generalized view, Co-training implies the training of two or more different models in some aspects ( e.g., different inputs, different regressors, different parametrization, different examples ...). To process the unlabeled examples, the regressors are trained with examples (previously unlabeled) that were artificially labeled by other complementary regressors. This fact means that the complementary regressors are assumed to predict reliably which makes the Co-training confidence driven.

Co-training methods consider the following assumptions: consensus, complementary, sufficiency, compatibility and conditional independence.

- **Consensus** assumption states that the more close (high level of agreement according to an error or a dispersion measure) the regressors predictions are, the more reliable is the respective combination of predictions [18].

- **Complementarity** assumption states that each regressor contains information that the complementary regressors do not contain. This assumption ensures that the complementary regressors can learn with each other. Hence, the use of multiple regressor increase the amount of information to construct more accurate models [18].

- **Sufficiency** assumption states that each regressor should be sufficiently consistent (e.g., by enough number of attributes) to build a model.

- **Compatibility** assumption implies that the predictions of different models are very similar for the same example in terms of probabilistic distribution.

- **Conditional independence** assumption considers that the learning process of each regressor is independent and at least one predicts with low errors. The regressor predictions can teach the other regressors with the most probably correct example [19]. Despite being very important for Co-training, the independence assumption is very restrictive. Therefore, related but less restrictive assumptions were considered. **Weak dependence** assumption tolerates a small dependence level between inputs which lead to positive results [20].

  **Large diversity** assumption considers that using different algorithms or the same algorithms but with different parametrization lead to independent models [21].

Concerning the drawbacks, the inaccuracy of the artificially labeled examples introduce error into the models and it is the main cause of model degradation. Moreover, the artificially labeled examples may not carry the information to the regressor leading to unnecessary operations [11]. Different strategies to artificially label or criteria to discard non-beneficial artificially labeled examples may be present in some Co-training variants. The prediction stage generally combines the predictions of the models according to a pre-defined criterion to produce the final prediction [11].

Co-training regression (COREG), Co-training by Committee for Regression (CoBCReg), Co-regularised least squares regression (coRLSR) and Tri-training were the main Co-training studied methods.

COREG method uses two regressors [4]. First, the labeled and unlabeled examples are separated in two sets. For each regressor, a set of labeled examples which are close (using a distance measure) to the input vector of unlabeled example are selected. Each regressor predicts a value to artificially label the example and uses it to re-train the models with all labeled examples. Mean Squared Error (MSE) variation is computed between the scenarios with and without the artificially labeled examples. If MSE is reduced, the artificially labeled example is joined to the labeled examples set. The process stops when none of unlabeled examples is interchanged between labeled and unlabeled sets. The final prediction is obtained by averaging the predictions of the two regressors.

CoBCReg resorts to Radial Basis Functions regressors ensembles (with a Gaussian basis function that uses the Minkowski distance) and Bagging. This method assumes that diversity must exist between the regressor of the ensemble, which is achieved by different input subsets random initialization. Each regressor of the ensemble selects the unlabeled examples that are more relevant for the respective model [22].

coRLSR (Co-regularised least squares regression) formulates the Co-training problem into a regularised risk in Hilbert spaces minimisation problem [23]. This method aims to find the models that minimizes the error of all models and the disagreement on unlabeled examples predictions.

Tri-training is a Co-training algorithm for classification which use three classifiers. The classifier achieve diversity by using different sets of examples. The final prediction is obtained by majority vote. For regression, this operation would be the equivalent of prediction mean [24].

## 3 Online Co-training Regression

This section provides the description of the proposed Co-training method through the presentation of the main adaptations to the online and regression context. A small description of the underlying algorithm regressor Random AMRules (ensemble rules based method) is also presented.

The new method that is being proposed divides the inputs variables of the example into two groups randomly which is defined in the initial stage. Here, weak dependence is assumed since no independence information between pairs of attributes is available. The complementarity assumption is also used since each produced model contains information that other does not contain.

The two groups are forced to share a randomly selected inputs by a pre-defined overlap percentage. Two Random AMRules complementary regressors are used to produce artificial labels through prediction for the unlabeled example. The initial models are obtained previously in a training stage using a dataset portion. The size of dataset portion should be sufficient to produce a consistent model. Here, the inputs overlapping increase the number of attributes in each model and contribute for the sufficiency assumption.

A score that reflects the benefit or confidence of artificially labeled example is calculated for the decision of being accepted for training. The score is the relative difference (RD) compared to the maximum of absolute values of the output found in the stream $y_{max}$. Here, the consensus assumption is used. Equation 1 defines de relative difference.

$$RD = \frac{|\hat{y}_i^1 - \hat{y}_i^2|}{y_{max}} \quad (1)$$

If the score is lower than a pre-defined threshold, the predictions are used to train the complementary regressor. Otherwise, the artificially labeled example is rejected. The consensus assumption is used in this step. If the example is labeled, this example is used to compute the mean error for each regressor. Next, the example is used for all regressors training. Here, the compatibility assumption is used since both models are trained with the same output. Algorithm 1 explains the training procedure of the proposed method.

Prediction is performed by combining the regressor predictions through prediction weighting. The weights are computed by inverting the values of the respective error produced by labeled examples in the training stage since the higher the error is, the less the artificial example benefits the model. In other words, this strategy gives more credit to the regressor that produces less errors. Algorithm 2 shows the steps of label prediction.

The Random AMRules regressor was employed to train the models and to produce the artificial labels for the unlabeled examples [16]. Random AMRules is a multi-target algorithm (predicts several outputs for the same example) that is based on rule learning which can be calibrated to work on single-target mode [3].

In essence, Random Rules is an ensemble based algorithm that uses bagging to create diversity and uses AMRules algorithm as a regressor. AMRules divides the input space in order to train local model in each partition. AMRules partionates the input space and creates local models for each partition. The local models are trained using a single layer perceptron. Its main advantages are models simplicity, low computational cost and low error rates [3].

Modularity is one of the main advantages. In fact, this method allows the train of models for local input sections

---

**Algorithm 1** Training algorithm of the proposed method

1: **Initialization:**
2:   $\alpha - Overlap\ percentage$
3:   $s - Score\ Threshold$
4:
5:   $Random\ input\ allocation\ and\ overlapping$
6:   $into\ the\ two\ groups\ using\ \alpha$
7:
8: **Input:** $Example\ (\mathbf{x}_i, y_i) \in \mathcal{S}$
9:
10: **Output:** $Updated\ Models$
11:
12: **Method:**
13:
14:   $Divide\ \mathbf{x}_i\ into\ \mathbf{x}_i^1\ and\ \mathbf{x}_i^2$
15:
16:   **if** $(y_i = \emptyset)$ **then**
17:
18:     $\hat{y}_i^1 = PredictModel1(\mathbf{x}_i^1)$
19:
20:     $\hat{y}_i^2 = PredictModel2(\mathbf{x}_i^2)$
21:
22:     **if** $(|\hat{y}_i^1 - \hat{y}_i^2|/y_{max} < s)$ **then**
23:
24:       $TrainModel1((\mathbf{x}_i^1, \hat{y}_i^2))$
25:
26:       $TrainModel2((\mathbf{x}_i^2, \hat{y}_i^1))$
27: **else**
28:
29:     $\bar{e}_1 = Update\ the\ mean\ error\ of\ Model1(\hat{y}_i^1, y_i)$
30:
31:     $\bar{e}_2 = Update\ the\ mean\ error\ of\ Model2(\hat{y}_i^2, y_i)$
32:
33:     $TrainModel1((\mathbf{x}_i^1, y_i))$
34:
35:     $TrainModel2((\mathbf{x}_i^2, y_i))$

---

**Algorithm 2** Prediction algorithm of the proposed method

1: **Input:** $Example\ (\mathbf{x}_i, y_i) \in \mathcal{S}$
2:
3: **Output:** $Example\ prediction\ \hat{y}_i$
4:
5: **Method:**
6:
7: $Divide\ \mathbf{x}_i\ into\ \mathbf{x}_i^1\ and\ \mathbf{x}_i^2$
8:
9: $\hat{y}_i^1 = PredictModel1(\mathbf{x}_i^1)$
10:
11: $\hat{y}_i^2 = PredictModel2(\mathbf{x}_i^2)$
12:
13: $w_1 = \bar{e}_2/(\bar{e}_1 + \bar{e}_2))$
14:
15: $w_2 = \bar{e}_1/(\bar{e}_1 + \bar{e}_2)$
16:
17: $\hat{y}_i = w_1 * \hat{y}_i^1 + w_2 * \hat{y}_i^2$

limited by the rule that are more precise. This algorithm also resorts to anomaly detection to avoid data outliers damage. Moreover, change detection on the stream is also employed by this method in order to avoid the influence of old information on the current predictions. The ensembles of AMRules can benefit the prediction by creating multiple and diverse regressor models by pruning the input partitions. The multiple regressor predictions create more possibilities to find a more accurate value. The ensembles also lead to a more stable final prediction and data change resilience.

## 4 The Evaluation Method

The evaluation method and the material used in the experiments are described in this section.

Real-world and artificial datasets were used to evaluate the proposed algorithm through a data stream simulation. A portion of 30% of the first examples of the stream were used for a initial consistent model training and the remaining 70% were used in the testing.

In order to produce an unlabeled examples in the test stage, a binary Bernoulli random process with a probability *p* was used to assign an example as labeled or unlabeled. In case of unlabeled assignment, the true output value is hidden from the algorithm. The *p* probabilities of unlabeled examples occurrence were 50%, 80%, 90%, 95% and 99%.

The score threshold values for algorithm parametrization/calibration were $1 \times 10^{-4}$, $5 \times 10^{-4}$, 0.001, 0.005, 0.01, 0.05, 0.1, 0.5 and 1. These values of score threshold are justified by the possibility of algorithm behaviour observation in multiple scales of this parameter. The overlap percentages assume the following values: 0%, 10%, 30%, 50%, 70% and 90%. The evaluation was performed in Prequential mode where in example arrival, the label prediction is performed first and then the example is used in the training [25]. Each Random AMRules regressor consists of ten regressors ensemble. This value was determined in a validation step were no significant improvement was observed above 10 regressors.

In these experiments, five real world and four artificial datasets were used to simulate the data stream. The real world datasets were House8L (Housing Data Set), House16L (Housing Data Set), CASP ( Physicochemical Properties of Protein Tertiary Structure Data Set), California, blogData-Train and the artificial datasets were 2dplanes, fried, elevators and ailerons. These datasets contain a single-target regression problem and are available at UCI repository [26].

Table 2 an shows the features of the real world and artificial data sets used in the method evaluation.

The performance measures used in these experiments were the mean relative error (MRE) and the mean percentage of accepted unlabeled examples (MPAUE) in the training. The MRE is used as an intermediate measure to quantify the prediction precision of each test scenario for both labeled and unlabeled examples. The MRE Reduction (MRER) was measured by using the relative difference (in percentage) between the reference scenario (no unlabeled examples used) $MRE_0$ and the case with the parametrization that lead to the lowest error $MRE_{lowest}$ (includes the reference

Table 1: Real world datasets description

| Dataset | # Examples | # Inputs |
|---|---|---|
| House8L | 22784 | 8 |
| House16H | 22784 | 16 |
| calHousing | 20640 | 7 |
| CASP | 45730 | 9 |
| blogDataTrain | 52472 | 281 |

Table 2: Artificial datasets description

| Dataset | # Examples | # Inputs |
|---|---|---|
| 2dplanes | 40768 | 10 |
| fried | 40768 | 10 |
| ailerons | 13750 | 41 |
| elevators | 8752 | 18 |

case $MRE_0$). Equation 2 defines the MRER performance measure.

$$MRER = \frac{|MRE_0 - MRE_{lowest}|}{MRE_0}.100 \ (\%) \qquad (2)$$

If the reference case yields the lowest error, then the MRER is zero, which means that the algorithm is not useful for that particular scenario.

MPAUE is used to measure the amount of useful unlabeled examples in the process of training which contributed to model improvement.

Massive Online Analysis (MOA) platform was used to accommodate the proposed algorithm [27]. This platform contains Machine Learning and Data Mining algorithms for data streams processing and was developed in JAVA programming language.

## 5 Results

In this section, the evaluation results are presented and discussed. Some scenarios examples of MRE and MPAUE plots for overlap percentage and score threshold combination are presented. In the MRE plots, the reference curve (Ref) corresponds to the scenario where no unlabeled examples are used in the training, hence this curve is constant. Similarly, in the MPAUE plots, the reference curve is constant and equal to zero (zero unlabeled examples accepted). For each combination of overlap percentage and score threshold, the experiments were performed in 10 runs due to the fact that the inputs are selected randomly. This procedure is important to obtain more consistent values. The results also include the presentation of the MRER for each dataset and the unlabeled examples percentage simulation.

Figure 1 depicts the MRE and MPAUE values for a successful case for 80% of unlabeled examples stream. The curves on the top reveal the existence of some combination of overlap percentage and score threshold values that lead to beneficial(lower error than the reference curve) use of unlabeled examples.

For the particular case of overlap of 50% and score threshold of 0.001, the use of 9.1% of the unlabeled examples in the training lead to reduction of 3.85% of the MRE in average for the House16H dataset. In general, it was also observed that the overlapping decrease the MRE.

Figure 2 presents a case where the method does not present any combination of overlap percentage and score thresholds values that lead to model improvement. In this scenario, the failure is explained by the fact of many unlabeled examples lead to model degradation and the artificial labels were very inaccurate (the curves of unlabeled examples scenarios are above the reference curve). This fact indicates that features of the datasets such as inputs variables distributions may dictate the performance.

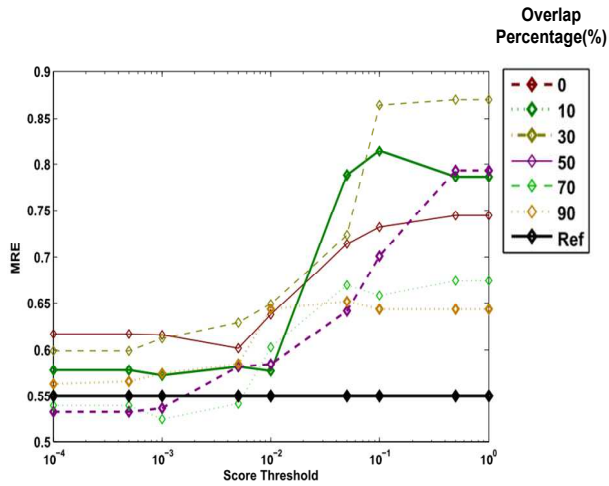

(a)

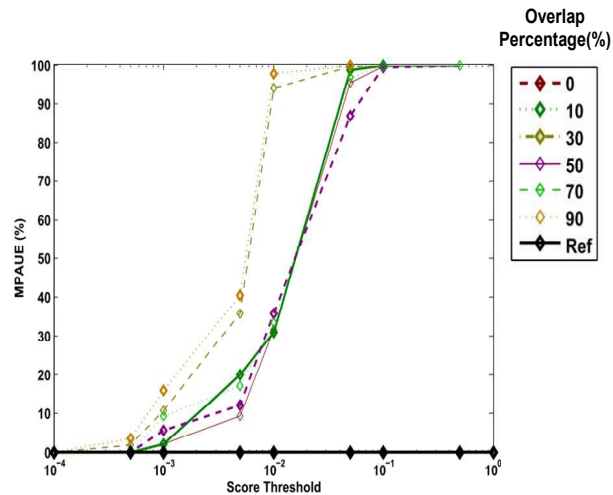

(b)

Figure 1: Mean relative error (a) and mean percentage of accepted unlabeled examples (b) for a data stream with 80% of unlabeled examples. The examples are from the House8L dataset.

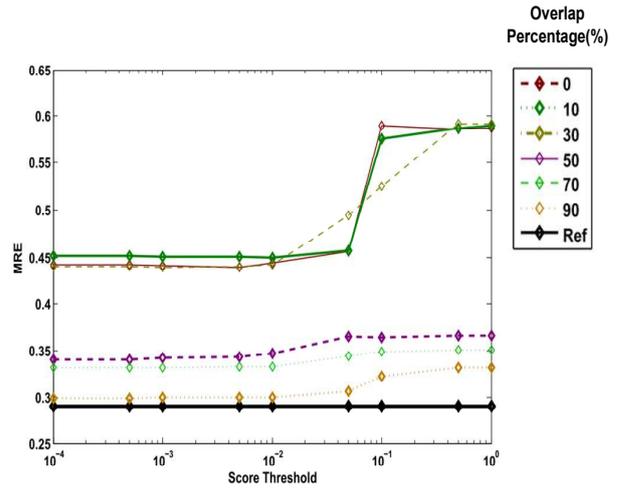

(a)

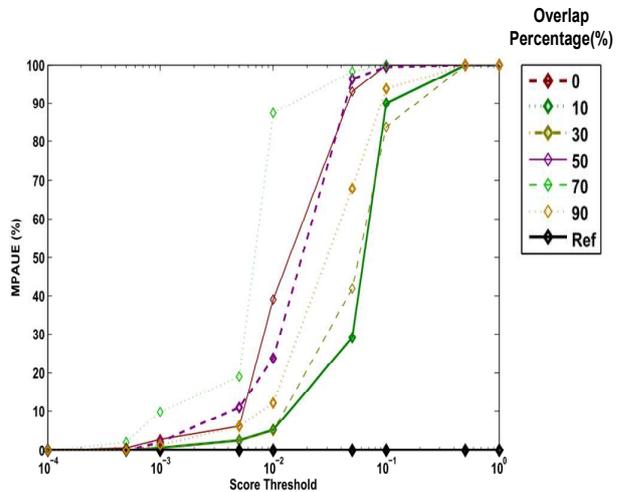

(b)

Figure 2: MRE (a) and MPAUE (b) for a data stream with 90% of unlabeled examples. The examples are from the CASP dataset.

This methods are prone to error propagation through the model. The error propagation through the model lead to worst predictions in the artificial labeling. This effect leads to a cycle that reinforce the error on each unlabeled example processing. In fact, the more unlabeled examples arrive the higher is the error.

Table 3 provides the MRER values of the experiments on real world datasets for each chosen unlabeled examples probabilities. When MRER assumes the zero value, a combination of overlap percentage and score threshold values that improves the model was not found and the reference scenario presents the lower MRE.

Table 3: MRER (%) for real world datasets

| Datasets | Unlabeled examples probabilities | | | | |
|---|---|---|---|---|---|
| | 50% | 80% | 90% | 95% | 99% |
| House8L | 2,23 | 3,21 | 2,77 | 0,00 | 0,00 |
| House16H | 3,85 | 1,93 | 0,32 | 0,00 | 0,00 |
| calHousing | 2,37 | 2,02 | 0,75 | 0,01 | 0,00 |
| CASP | 0,8 | 1,65 | 0,00 | 0,00 | 0,00 |
| blogDataTrain | 1,17 | 0,40 | 0,37 | 0,00 | 0,00 |

Table 3 suggests that the proposed algorithm seems to improve the performance for most part of the scenarios. Despite this fact, the MRER are in general relatively small. As anticipated, the more elevated the probability of unlabeled example is, the less is the MRER. Table 4 provides the MRER value for real artificial datasets in similar way as the real world datasets presented in Table 3.

Table 4: MRER (%) for artificial datasets

| Datasets | Unlabeled examples probabilities | | | | |
|---|---|---|---|---|---|
| | 50% | 80% | 90% | 95% | 99% |
| 2dplanes | 2,39 | 0,90 | 0,75 | 0,00 | 0,00 |
| fried | 3,55 | 3,35 | 1,71 | 0,00 | 0,00 |
| ailerons | 2,67 | 1,79 | 0,01 | 0,95 | 0,00 |
| elevators | 1,35 | 1,11 | 0,71 | 0,00 | 0,00 |

The results on artificial datasets also support the view that the more elevated the unlabeled probability is, the less is the benefit of the unlabeled examples. The MRER is similarly small.

In general, the MRE curves and the error reduction tables support the conclusion that the algorithm leads to an error reduction (despite being small) by using labeled examples in most cases. The results show that for 99% of unlabeled examples probability, the method does not produce beneficial artificial labels. This high level of unlabeled examples in the stream represents an extreme scenario where the model is training almost with artificially labeled examples and the high error propagation can frequently occur.

# 6 Conclusion

This paper addresses an online Co-training and Semi-supervised algorithm for single-target regression based on ensembles of rule models. This work is the base for the development of multi-target regression methodology capable of using unlabeled examples information for model improving.

The results support that this Co-training approach using Random AMRules method reduces the error with the appropriate parameters calibration. The main contribution was the overlapping and the consensus measure strategies that contribute to increase diversity and model consistency in a online co-training scenario. In fact, the MRER is positive when an amount of unlabeled examples are used in the training in most evaluation combinations. Despite this fact, the model benefit is still relatively small and the performance is highly dependent of a good parametrization tuning (score threshold and overlap percentage). In addition, the amount of unlabeled examples is relatively small to obtain some model improvement.

Considering future work, this work will be extended to multi-target regression. The fact that very few unlabeled examples can lead to some improvement may suggest the study of the conditions that lead to this improvement. To increase the method validity, future works will include a higher number of real world datasets with higher amount of examples. Datasets with particular features such drifts presence are also in view.

# 7 Acknowledgements

This work is financed under the project "NORTE-01-0145-FEDER-000020" funded by the North Portugal Regional Operational Programme (NORTE 2020), under the PORTUGAL 2020 Partnership Agreement, and through the European Regional Development Fund (ERDF).